\documentclass[10pt, dvipsnames]{article}
\selectfont

\pdfoutput=1
\usepackage{enumitem}
\usepackage{titlesec}
\usepackage{authblk}
\usepackage[dvipsnames]{xcolor} 
\usepackage{ucs}
\usepackage{longtable}
\usepackage[colorlinks=true,
  urlcolor=ForestGreen,
  linkcolor=Blue,
  citecolor=Violet,
  pagebackref=true,
  pdftitle={Titel},
  pdfsubject={},
  pdfauthor={},
  pdfkeywords={keywords}
]{hyperref}
\usepackage{graphicx}
\usepackage{subfigure}
\usepackage[subfigure]{tocloft}
\usepackage[font=footnotesize]{caption}
\usepackage{tabularx}
\usepackage{nameref}
\usepackage{tikz-cd}

\usepackage{amsmath} 	  
\usepackage{amssymb}	  
\usepackage{amsthm}	    
\usepackage{mathtools}  
\usepackage[normalem]{ulem} 
\usepackage{graphicx,wrapfig,lipsum}
\usepackage{float}
\usepackage{floatflt}
\usepackage{natbib}
\usepackage[capitalize,nameinlink]{cleveref}

\newtheorem{lemma}{Lemma}

\usepackage[most]{tcolorbox}
\tcbuselibrary{breakable} 
\usepackage{wrapfig}
\usepackage[toc,page]{appendix}

\newtcolorbox[auto counter, number within=section
                            ]{blueproposition}[1][]{enhanced jigsaw,
  colback=white, 
  coltext={black},
  coltitle={black},
  boxrule=0pt,
  arc=1mm,
  frame hidden,
  auto outer arc,
  boxsep=5pt,
  left=2pt,
  right=2pt,
  bottom=2pt,
  top=2pt,
  borderline west={1mm}{0mm}{white!60!blue},
  before skip=3mm,
  after skip=3mm,
  breakable,
  title={Proposition~\thetcbcounter.},
  attach title to upper=\quad,
  fonttitle=\bfseries,
  #1}

\newtcolorbox[auto counter, number within=section
  ]{bluecorollary}[1][]{enhanced jigsaw,
colback=white, 
coltext={black},
coltitle={black},
boxrule=0pt,
arc=1mm,
frame hidden,
auto outer arc,
boxsep=5pt,
left=2pt,
right=2pt,
bottom=2pt,
top=2pt,
borderline west={1mm}{0mm}{white!60!blue},
before skip=3mm,
after skip=3mm,
breakable,
title={Corollary~\thetcbcounter.},
attach title to upper=\quad,
fonttitle=\bfseries,
#1}

\newtcolorbox[auto counter, number within=section
                            ]{bluetheorem}[1][]{enhanced jigsaw,
  colback=white, 
  coltext={black},
  coltitle={black},
  boxrule=0pt,
  arc=1mm,
  frame hidden,
  auto outer arc,
  boxsep=5pt,
  left=2pt,
  right=2pt,
  bottom=2pt,
  top=2pt,
  borderline west={1mm}{0mm}{white!60!blue},
  before skip=3mm,
  after skip=3mm,
  breakable,
  title={Theorem~\thetcbcounter.},
  attach title to upper=\quad,
  fonttitle=\bfseries,
  #1}

\newtcolorbox[auto counter, number within=section
                            ]{redlemma}[1][]{enhanced jigsaw,
  colback=white, 
  coltext={black},
  coltitle={black},
  boxrule=0pt,
  arc=1mm,
  frame hidden,
  auto outer arc,
  boxsep=5pt,
  left=2pt,
  right=2pt,
  bottom=2pt,
  top=2pt,
  borderline west={1mm}{0mm}{white!60!red},
  before skip=3mm,
  after skip=3mm,
  breakable,
  title={Lemma~\thetcbcounter.},
  attach title to upper=\quad,
  fonttitle=\bfseries,
  #1}
  
\newtcolorbox[auto counter, number within=section
                          ]{debox}[1][]{enhanced jigsaw,
  colback=green!0!white,
  coltext={black},
  colframe={black},
  coltitle={black},
  boxrule=0pt,
  frame hidden,
  borderline west={1mm}{0mm}{white!50!green},
  arc=0mm,
  auto outer arc,
  boxsep=5pt,
  left=4pt,
  breakable,
  right=4pt,
  bottom=0pt,
  top=0pt,
  before skip=3mm,
  after skip=3mm,
  title={Definition~\thetcbcounter.},
  attach title to upper=\quad,
  fonttitle=\bfseries,
  #1}
\newtcolorbox[auto counter,number within=section
                          ]{yellowRemark}[1][breakable]{enhanced jigsaw,
  colback=green!0!white,
  coltext={black},
  colframe={black},
  coltitle={black},
  boxrule=0pt,
  frame hidden,
  breakable,
  borderline west={1mm}{0mm}{white!50!yellow},
  arc=0mm,
  auto outer arc,
  boxsep=5pt,
  left=4pt,
  right=4pt,
  bottom=0pt,
  top=0pt,
  before skip=3mm,
  after skip=3mm,
  title={Remark~\thetcbcounter.},
  attach title to upper=\quad,
  fonttitle=\bfseries,
  #1}
\newtcolorbox[auto counter,number within=section
                          ]{example}[1][breakable]{enhanced jigsaw,
  colback=green!0!white,
  coltext={black},
  colframe={black},
  coltitle={black},
  boxrule=0pt,
  frame hidden,
  breakable,
  borderline west={1mm}{0mm}{white!50!yellow},
  arc=0mm,
  auto outer arc,
  boxsep=5pt,
  left=4pt,
  right=4pt,
  bottom=0pt,
  top=0pt,
  before skip=3mm,
  after skip=3mm,
  title={Example~\thetcbcounter.},
  attach title to upper=\quad,
  fonttitle=\bfseries,
  #1}

  \newtcolorbox[auto counter,number within=section
                          ]{proofenv}[1][breakable]{enhanced jigsaw,
  colback=black!0!white,
  coltext={black},
  colframe={black},
  coltitle={black},
  boxrule=0pt,
  frame hidden,
  breakable,
  borderline west={1mm}{0mm}{white!50!black},
  arc=0mm,
  auto outer arc,
  boxsep=5pt,
  left=4pt,
  right=4pt,
  bottom=0pt,
  top=0pt,
  before skip=3mm,
  after skip=3mm,
  title={Proof.~},
  attach title to upper=\quad,
  fonttitle=\bfseries,
  #1}

  \newtcolorbox[auto counter,number within=section
                          ]{proofexp}[1][breakable]{enhanced jigsaw,
  colback=black!0!white,
  coltext={black},
  colframe={black},
  coltitle={black},
  boxrule=0pt,
  frame hidden,
  breakable,
  borderline west={1mm}{0mm}{white!50!black},
  arc=0mm,
  auto outer arc,
  boxsep=5pt,
  left=4pt,
  right=4pt,
  bottom=0pt,
  top=0pt,
  before skip=3mm,
  after skip=3mm,
  title={Proof and Explanation.~},
  attach title to upper=\quad,
  fonttitle=\bfseries,
  #1}

\usepackage{pictexwd,dcpic}

\newtcolorbox[
                          ]{shortsummary}[1][breakable]{enhanced jigsaw,
  colback=red!0!white,
  coltext={black},
  colframe={black},
  coltitle={black},
  boxrule=0pt,
  frame hidden,
  breakable,
  borderline west={1mm}{0mm}{white!50!red},
  arc=0mm,
  auto outer arc,
  boxsep=5pt,
  left=4pt,
  right=4pt,
  bottom=0pt,
  top=0pt,
  before skip=3mm,
  after skip=3mm,
  title={Short and important~\thetcbcounter:},
  attach title to upper=\quad,
  fonttitle=\bfseries,
  #1}

\usepackage{geometry}
\def\wider{%
   \advance\leftskip -\parindent
   \advance\rightskip -\parindent}

\newcommand{\te}[1]{\text{#1}}

\newcommand{\br}[1]{\mathrel{ \left(#1\right)  }}

\newcommand{\bt}[1][normal]{\begin{tikzcd}[ampersand replacement = \&, column sep=#1,row sep=#1]}
\newcommand{\et}{\end{tikzcd}}

  \tikzset{
    symbol/.style={%
        draw=none,
        every to/.append style={%
            edge node={node [sloped, allow upside down, auto=false]{$#1$}}}
    }
}

\makeatletter
\newcommand{\tpitchfork}{%
    \raise-0.1ex
    \vbox{
    \baselineskip\z@skip
    \lineskip-.52ex
    \lineskiplimit\maxdimen
    \m@th
    \ialign{##\crcr\hidewidth\smash{$-$}\hidewidth\crcr$\pitchfork$\crcr}
  }%
}
\makeatother

\usepackage{tocloft}

\usepackage{newtxtext}
\usepackage{lipsum} 

\makeatletter
\def\moverlay{\mathpalette\mov@rlay}
\def\mov@rlay#1#2{\leavevmode\vtop{%
   \baselineskip\z@skip \lineskiplimit-\maxdimen
   \ialign{\hfil$\m@th#1##$\hfil\cr#2\crcr}}}
\newcommand{\charfusion}[3][\mathord]{
    #1{\ifx#1\mathop\vphantom{#2}\fi
        \mathpalette\mov@rlay{#2\cr#3}
      }
    \ifx#1\mathop\expandafter\displaylimits\fi}
\makeatother

\usepackage{mathabx}
\usepackage{calc}
\usepackage{amscd}
\usepackage{pictexwd,dcpic}
\usepackage{makeidx}
\usepackage{tikz}
\usetikzlibrary{calc,intersections,through,backgrounds}
\usepackage{pgfplots}
\pgfplotsset{compat=1.18}

\usepackage{array}
\usepackage{latexsym}
\usepackage{epsfig}

\newcommand{\vio}[1]{\textcolor{violet}{#1}}

\newcommand{\R}{\mathbb{R}}

\newcommand{\bel}[1]{\begin{equation}\label{#1}}

\newcommand{\be}{\begin{equation}}
\newcommand{\ee}{\end{equation}}
\newcommand{\ba}{\begin{eqnarray}}
\newcommand{\ea}{\end{eqnarray}}
\newcommand{\rf}[1]{(\ref{#1})}

\newcommand{\qe}{\end{equation}}

\numberwithin{equation}{section}   
\numberwithin{figure}{section}

\newtheorem{theo}{Theorem}[section]
\newtheorem{coro}{Corollary}[section]

\theoremstyle{definition}
\newtheorem{defi}{Definition}[section]
\theoremstyle{remark}
\newtheorem*{pf}{Proof}
\theoremstyle{remark}
\newtheorem*{rem}{Remark}
\theoremstyle{remark}
\newtheorem{ex}{Example}

\title{IsUMap: Manifold Learning and Data Visualization leveraging Vietoris-Rips filtrations}
\author[1]{Lukas Silvester Barth, \thanks{lukas.barth@mis.mpg.de}}
\author[1,2]{Fatemeh (Hannaneh) Fahimi, \thanks{fatemeh.fahimi@mis.mpg.de}}
\author[1,2]{Parvaneh Joharinad, \thanks{parvaneh.joharinad@mis.mpg.de}}
\author[1,2,3]{J\"urgen Jost, \thanks{jjost@mis.mpg.de}}
\author[1,4]{Janis Keck,\thanks{janis.keck@maxplanckschools.de}}
\affil[1]{Max Planck Institute for Mathematics in the Sciences, Leipzig, Germany}
\affil[2]{Center for Scalable Data Analytics and Artificial Intelligence (ScaDS,AI) Dresden/Leipzig, Germany}
\affil[3]{Santa Fe Institute for the Sciences of Complexity, New Mexico, USA}
\affil[4]{Department Psychology (Doeller), Max Planck Institute for Human Cognitive and Brain Sciences, Leipzig, Germany}
\begin{document}

\maketitle
\begin{abstract}
  This work introduces IsUMap, a novel manifold learning technique that enhances data representation by integrating aspects of UMAP and Isomap with Vietoris-Rips filtrations. We
  present a systematic and detailed
  construction of a metric representation for locally distorted metric spaces that 
  captures complex data structures more accurately than the previous schemes.
  Our approach addresses
   limitations in existing methods by accommodating non-uniform data distributions and intricate local geometries. 
We validate its performance through extensive experiments on examples of various geometric objects and benchmark real-world datasets, demonstrating significant improvements in representation quality.    
\end{abstract}
\section*{Introduction}
A good representation of  complex data can enable an intuitive understanding of
their important features, and as such is indispensable for data analysis in a
wide range of scientific disciplines and applications. Furthermore, good representations are often more compact (and hence computationally more efficient) than the original data, while capturing or even pronouncing the relevant features. This makes them useful for downstream tasks, transfer learning and a better understanding of the data.\\ 
Numerous tools have been developed to transform complex data into low-dimensional representations. For instance, feature learning or representation learning techniques in machine learning enable systems to automatically discover the necessary representations for various downstream tasks.  
A standard strategy in representation learning is to train a deep neural network and call one of the latent layers the representation layer, which is then investigated with a variety of techniques (cf. \cite{Bengio13}, \cite{Botteghi22}). 
Though this a fascinating and fruitful approach, it has the disadvantage that those representations are hard to interpret because of the large number of learnable parameters in those networks. 
In this article, we therefore concentrate on a manifold learning algorithm that has only very few predefined parameters and is hence particularly transparent.\\
Another widely used representation is the set of barcodes in persistent homology, a method in Topological Data Analysis (TDA), which captures the persistent topological features of a point cloud.  
TDA employs filtered simplicial complexes
as described in \cite{Zomorodian04, Carlsson09}. In this paper, we develop a conceptually
different approach,  leveraging these filtrations to combine local geometries,  to achieve a unified global geometric representation rather than focusing solely on topological inference.\\
To achieve this, we combine ideas from TDA with the established dimensionality reduction  methods of UMAP \cite{McInnes18} and Isomap \cite{Tenenbaum00}, but our approach improves on them, in particular when the data is not uniformly distributed and concentrated around a low-dimensional manifold. In fact, starting with the assumption that data is initially sampled from a Riemannian manifold, we eventually end up with a triangulated metric space composed of distorted local pieces of that manifold.\\ 
We call our method IsUMap to express that it represents a combination of Isomap and UMAP and that it employs the category UM of uber metric spaces. According to its objective, it can be considered an isometric uniform metric space approximation method.\\  
A specific challenge that our method addresses is the aggregation of pairwise similarities between data points, represented by a distance function, across overlapping neighborhoods. This approach is for instance relevant when applying local distortions to the distance function, initially defined as a global measurement, to facilitate specific downstream tasks. \\
 Traditionally, a dataset is sampled from an ambient space, often a Euclidean space. However, this paradigm is evolving as tasks increasingly require specialized similarity measurements between data points. These measurements might be represented by non-scalar features or by functions that only locally satisfy the properties of a distance function. \\
   We construct a unified intrinsic global distance function from metrics defined on overlapping local neighborhoods. The resulting global distance function is then embedded into a low-dimensional space using Multidimensional Scaling (MDS, cf. \cite{Torgerson52, lee07}), a classical method that represents metric data in a low-dimensional Euclidean space while preserving the global metric structure. \\
 MDS is particularly effective when the input metric is a reliable intrinsic distance function (cf. \cite{lim22}). However, when dealing with data sets, the initial distance function often provides intrinsic similarity only at local scales. A metric $d$ on a set $X$ is called intrinsic if the distance between each pair of points is obtained as the infimum of the length of the paths connecting those points, where the paths traverse through $X$. The choice of  using MDS for the final embedding in some Euclidean space in our algorithm is then justified by the fact that it is applied only after inferring the intrinsic metric.  \\
 Other manifold learning-based representation schemes include the prominent method of Laplacian Eigenmaps (LE) (cf. \cite{Belkin03}), which is often leveraged as an initialization in UMAP. Another notable method is Isomap, which enhances the aforementioned Multidimensional Scaling by incorporating a pre-step to compute intrinsic distances using local metric information. The difference between our method and Isomap is that the global metric is obtained by computing geodesic distances in a graph formed through a combinatorial combination of Vietoris-Rips filtrations derived from the initial local distance functions. 
\\ 
Essentially, a manifold learning scheme is a process of distilling a combinatorial model (usually a graph) from a metric, inferring some information about the geometric structure of the underlying manifold from that structure, and finally providing an embedding using this specific information. \\
We propose using simplicial complexes, in particular the Vietoris-Rips filtration corresponding to the given metric, as the combinatorial structure. Simplicial complex filtrations can encode higher-order relationships, such as those found in social networks, which cannot be captured by binary metric relations. This capability suggests that simplicial complexes may be a natural mathematical object for data analysis, even when the analysis is geometric, and extends beyond inferring topological characteristics as in TDA.

We use Vietoris-Rips filtrations to combine information from a family of metrics obtained by imposing local distortions on an initial metric. This process yields a weighted simplicial complex, from which we then infer a metric on its vertex set through metric realization based on the exposition in \cite{Spivak09}.

We also clarify how the relationships between Vietoris-Rips filtrations and metric spaces can equivalently be expressed in category theoretical language, and how the conciseness of this language can help to motivate definitions and guide the search for theorems. \\

Since we do not visualize abstract (geometric or topological) features, as for instance with barcodes in TDA, but rather want to represent the data directly,
it is crucial to estimate the appropriate dimension based on the specific task at hand. However, for the mere visualization purpose dimension 2 or at most three will suffice.\\
When the  data points are represented by scalar features and thereby embedded in a possibly high-dimensional Cartesian space, our method can be viewed as a dimensionality reduction method. Similar to many such methods, IsUMap mainly relies on pairwise similarities depicted in a distance function.
\medskip
The organization of this article is as follows. In Section
\ref{preliminaries} we elaborate on preliminaries, reiterating pertinent
definitions and theorems from metric and Riemannian geometry,  together with a
discussion of  filtrations of simplicial complexes, notably the Vietoris-Rips
filtration. In Section \ref{sec:metricrealization} we explain how to obtain the metric realization of an (admissible) weighted simplicial complex based on \cite{Spivak09}. In Section \ref{Sec:merging}, we combine metric spaces utilizing their corresponding Vietoris-Rips filtrations. In Section \ref{sec:CatTheory}, we explain how all the concepts that were described up to this point, can also be reformulated and generalized in category theoretical language and cite corresponding theorems from our related article \cite{otherPaper}.
In Section \ref{sec:IsUMap}, we provide a detailed description of the IsUMap algorithm and in
the final Section \ref{sec:illustrations}, we illustrate the performance of
the algorithm on various data sets. 

\section{Preliminaries}\label{preliminaries}
We consider a discrete data set $X=\{x_1,x_2,\dots ,x_N\}$, where distinct points can be quantitatively compared via some proximity measure. This proximity is best formulated as a metric (or distance function). 
\begin{defi}\label{metric-def}
A metric $d$ on a set $X$ is a map $d:X\times X\to \mathbb{R}_{\ge 0}$ which assigns to each pair $(x,y)$ of points in $X$  a non-negative real number, satisfying the following properties:
\begin{enumerate}
\item[1.] $d(x,y)\ge 0$, and  $d(x,y)=0$ iff $x=y$;
\item[2.]  $d(x,y)=d(y,x)$; and
\item[3.]  $d(x,z)\le d(x,y)+d(y,z)$
\end{enumerate}
for all $x,y,z\in X$.
\end{defi}
In practice, the function $d$ that quantifies the similarity between points
may sometimes fail to satisfy one of the properties of a metric. However,
theoretically, there usually exists a semi-metric obtained by relaxing one of
these properties. When  $d(x,y)=0$ may also happen for some points $x\neq y$,
we get a pseudo-metric. We get an  extended metric when distances may be
infinite and where the triangle inequality need not hold when  $d(x,z)=\infty$. Actually, we shall simply write "metric space" below when more precisely, we mean such an extended metric space.\\
The metric $d$ defines a \emph{topology} on $X$ by considering open subsets
obtained from finite intersections of metric open balls 
\be\label{openball}
U(x,r):=\{y\in X: d(x,y)<r\}.
\qe
A \emph{topological space} is a set $X$ along with a family ${\cal O}$ of
subsets, the  \emph{open sets}, satisfying the following properties:
\begin{enumerate}
\item
$\emptyset , X \in {\cal O}$.
\item
If $U_1, U_2 \in {\cal O}$, then also $ U_1 \cap U_2 \in {\cal O}$;
\item
for any index set $I$, if $(U_\iota)_{\iota \in I} \subset {\cal O}$, then also
$\bigcup_{\iota \in I} U_\iota \in {\cal O}$:
\end{enumerate} 
When we have topologies, we can define continuity of maps. Thus, a map $F:X\to Y$, where $X$ and $Y$ are both topological spaces, is \emph{continuous} if and only if the inverse image $F^{-1}({\cal O})$ of every open ${\cal O}\subset Y$ is open in $X$.\\
There is also a method for constructing scale dependent graphs or simplicial
complexes from a dataset $X$ that is equipped with a metric $d$. For every
scale $r\ge 0$, two data points are connected by an edge whenever their
distance is $\le r$. We can then construct the clique complex of the resulting
graph, called  the \emph{Vietoris-Rips complex}, by filling in triangles, tetrahedrons, and so forth whenever all the required edges exist in the graph. \\
In combinatorics, an abstract \emph{simplicial complex} with a vertex set $X$ is a family of finite subsets in $\mathcal{P}(X)$, the power set of $X$, which is closed under taking subsets. This combinatorial structure  models relations with a hereditary property, where the relation extends to all subsets of a subset $\sigma$ that satisfy the relation. 
Here is the formal definition of the  Vietoris-Rips complex: 
\begin{defi}\label{vrcomplex-def}  
For a metric space $(X,d)$ and $r\ge 0$, the Vietoris-Rips complex $VR(X,r)$ is the simplicial complex with vertex set $X$, where every finite subset $\{x_0,x_1,\dots,x_n\}$ of $X$ spans a simplex if its diameter is $\leq r$, i.e the distance of any pair of points in the set is not larger than $r$. 
\end{defi} 
As we vary $r$, a family of simplicial complexes emerges, evolving as the scale increases, i.e. 
\[
VR(X,r)\subseteq VR(X,r'), \: \text{for}\; r\leq r'.
\]
To visualize an abstract simplicial complex $K$, we turn it into a topological space $|K|$, its geometric realization.  While there is not a unique geometric realization for every simplicial complex, the standard geometric realization provides a canonical way to construct it, especially when the vertex set $X$ is finite.\\
 Let assign a total order on the vertex set $X$ of $K$ which is assumed to be finite, that is we assume that $X=\{x_1,\dots, x_N\}$. We then simply employ the standard $(N-1)$-simplex in $\R^N$
\bel{N-1sim}
\Delta_{N-1}:=\left\{(t_0,\dots,t_{N-1})\in\R^{N}\bigg|\sum_{i=0}^{N-1}t_i=1: t_i\geq 0\right\}
\qe 
to construct the standard geometric realization of $K$, i.e. $|K|$, as we explain below.\\ 
$\Delta_{N-1}$ is a simplex with vertex set consisting of elements in the standard basis $\{e_1,e_2,\dots,e_N\}$ of $\R^N$. Every face spanned by $\{e_{i_0},e_{i_1},\dots,e_{i_k}\}$ is the convex hull of the corresponding vertices. $\Delta_{N-1}$ is equipped with a canonical metric, i.e.\ the restriction of Euclidean distance to $\Delta_{N-1}$. \\
$|K|$ is a subcomplex of $\Delta_{N-1}$ defined as the collection of faces spanned by $\{e_{i_0},e_{i_1},\dots,e_{i_k}\}$ for which $\{x_{i_0},x_{i_1},\dots,x_{i_k}\}$ form a $k$-simplex in $K$. $|K|$ inherits the canonical metric of $\Delta_{N-1}$, making it a metric space. \\
 Although this process yields a metric realization for each $VR(X,r)$, this metric assigns a distance between vertices of $|VR(X,r)|$ which differs from that given by $d$. Specifically,  the Euclidean distance between every pair $(e_{i},e_{j})$ is equal to $\sqrt{2}$ while the distance between corresponding vertices in $X$ (that is $d(x_i,x_j)$) is not necessarily constant. \\
 While the geometric realization (specifically, the standard geometric realization) of Vietoris-Rips complexes does not accurately reflect the geometry of $X$, it can recover its topology. This is supported by the following \emph{Hausmann theorem}, c.f \cite{Hausmann95} (see also \cite{Latschev01}):
\begin{theo}\label{Hausmann-theo}
let $(M,g)$ be a Riemannian manifold that is compact (or more generally,
  satisfies some technical condition that essentially amounts to a positive
  lower bound on the injectivity radius), and $VR(M,r)$ be the Vietoris-Rips
complex at scale $r$, corresponding to the metric $d$ defined by the Riemannian metric tensor $g$. Then there exists $r_0>0$ such that for every $r\leq r_0$, the geometric realization of $VR(M,r)$ is homotopy equivalent to $M$. 
\end{theo}  
Homotopy equivalence is important because in algebraic topology many concepts
are homotopy invariant, that is, their values are invariant under  homotopy equivalence. \\
More generally, the widely-used method of persistent homology in topological
data analysis records the homology of the scale-dependent combinatorial model
$\left(VR(X,r)\right)_{r\geq 0}$ to recover topological features of a dataset
$(X,d)$. Here, however, one is interested both in invariances and in
  changes of homology, the appearance and disappearance of homology
  generators, to identify scales of particular interest.\\
Indeed, for every scale $r$, the object $VR(X,r)$ forms a simplicial complex, enabling the computation of corresponding simplicial homology groups $H_*(VR(X,r))$ and their generators, i.e., non-trivial cycles which represent topological features. \\
The inclusion map $VR(X,r)\subseteq VR(X,r')$ for $r\leq r'$ induces morphisms $H_*(VR(X,r))\to H_*(VR(X,r'))$ between homology groups, facilitating the tracking of topological features across the scaling parameter. \\
Persistent features, enduring over substantial intervals, indicate 
intrinsic topological features of $X$, whereas those that vanish quickly are
indicative of noise. This method is justified and supported by the equivalence
of \v{C}ech homology and singular homology of a topological space under
certain conditions, c.f. \cite{Wallace07}, that is more general than the Hausmann theorem.\\
$1$-parameter filtrations $\left(S_r\right)_{r\geq 0}$ of simplicial complexes,
where $r$ is not necessarily a distance radius, often arise in practice, such as from a complex network via parameter-dependent relationships like the clique complex of a social network. Persistent homology has also been used for such a filtrations if the parameter-dependent correlation is defined in a way that the generated simplicial complexes grow as the parameter increases, c.f. 
\cite{Horak09, Myers19, Battiston20}. \\
The question then arises whether there exists a distance relation on the set
of nodes of a complex network that leads to such a filtration. This is the inquiry we aim to address utilizing the theory of metric realization introduced in \cite{Spivak09}.\\
More precisely, we seek a metric space $(Y,d)$ that encompasses the set of nodes $X$ such that $VR(Y,r)$ reflects the geometry of our $1$-parameter filtration. \\
For this purpose, we require a combinatorial model, a weighted simplicial complex, to represent the correlation between data points up to a certain scale. This model will be utilized to construct the metric realization of the corresponding complex.
\begin{defi}
Let $\left(VR(X,r)\right)_{0\leq r\leq R}$ be a portion of the Vietoris-Rips
filtration of the metric space $(X,d)$. Then we define the corresponding
weighted simplicial complex as a full simplex whose vertices are the elements
of  $X$, with the weights assigned by 
\be \label{fuzzyvietoris}
\begin{split}
W(\sigma)&:=\begin{cases}
\min\{0\leq r\leq R ; \sigma\in VR(X,r)\} , & \text{if} \; \sigma \in VR(X,R) \ \\
\infty \ &  \text{otherwise} 
\end{cases}
\end{split}
\qe
This assignment ensures that  the weight of each simplex is equal to the maximum of the weights of its $1-$dimensional faces, i.e. edges.
\end{defi}
Let $\Phi:[0,\infty]\to [0,1]$ be a strictly decreasing function such that
$\Phi(0)=1$ and $\Phi(\infty)=0$. When combined with the weight function $W$,
this results in a fuzzy simplicial complex. This is a classical fuzzy set,
with the particular property that the strength function  does not decrease as one traverses from a simplex to its faces.  \\
We now want to identify the conditions  to ensure that the inverse of $\Phi:[0,\infty]\to [0,1]$ (strictly decreasing with $\Phi(0)=1$ and $\Phi(\infty)=0$) converts a fuzzy simplicial complex with  strength function $w$ into a weighted simplicial complex with  weight function $W$, as described in \ref{fuzzyvietoris}, corresponding to some extended metric $d$ on the vertex set $X$. 
\begin{itemize}
\item[1-]The value of $w$ on vertices must be $1$ and $<1$ for other simplices. This ensures that the first property of a metric (being reflexive) is satisfied. 
\item[2-]The symmetry of the metric is guaranteed as there is no direction on $1$-simplices (edges).
\item [3-] The triangle inequality for the extended metric is  satisfied if and only if for any pair of vertices $(x,z)$, either $w([x,z])=0$ 
  or
\[ 
w([x,z]) \geq \Phi\left(\Phi^{-1}(w[x,y]) + \Phi^{-1}(w[y,z])\right), \forall y\in X
\] 
\item[4-] The weight $w(\sigma)$ of each simplex should be equal to the minimum of the weights on its $1$-dimensional faces.
\end{itemize}
We here set the strength of simplices, that are not present, to $0$.\\ 
The last condition could be relaxed to only imply that $w(\sigma)$ is at most equal to the minimum of the weights on its $1$-dimensional faces. This, for instance, corresponds to the \v{C}ech complexes of open balls centered at points in $X$ and with varying radii.\\
Therefore, if assuming that $(S, w)$ is a fuzzy simplicial complex on the vertex set $X$ with the strength function $w$ satisfying conditions 1,2,4 above, then one can search for a proper function $\Psi:[0,1]\to [0,\infty]$ (strictly decreasing with $\Psi(1)=0$ and $\Psi(0)=\infty $) such that the triangle inequality is satisfied. That is, for $(x,z)$ with $w([x,z])\neq 0$, one has $\Psi(w([x,z]))\leq \Psi(w([x,y]))+\Psi(w([y,z]))$.\\
\begin{ex}\label{expweight}
Let the fuzzy simplicial complex $(S,w)$ with vertex set $X$ satisfy conditions 1,2,4 above. If for every $(x,z)$ with $w([x,z])\neq 0$
\[
w([x,z])\geq w([x,y])w([y,z]), \;  \forall y\in X,
\]
 Then the map $\Psi=-log$ converts $S$ to a Vietoris-Rips filtration.
 \end{ex}

This fuzzy simplicial complex model proves especially advantageous when  combining two or more such complexes as we describe in Section \ref{Sec:merging}. 

\section{Metric realizations}\label{sec:metricrealization}
As outlined in Section \ref{preliminaries}, for each scale $r$, there exists a geometric realization of the simplicial complex $VR(X,r)$, denoted by $|VR(X,r)|$, where each $k$-simplex is represented by the standard geometric simplex $\Delta^k$ as defined in \ref{N-1sim}. \\
Inspired by this geometric approach, Spivak \cite{Spivak09} constructed the
metric realization of a fuzzy simplicial set. We use this to obtain a
realization for the weighted simplicial complex of Definition
\ref{fuzzyvietoris}. Since we are interested in finite data sets equipped with
discrete metrics, we will not  discuss non-finite metric spaces. \\
Here, we present a version of metric realization for a weighted simplicial
complex  slightly modified from  that of \cite{Spivak09}. We specifically tailor this method to accommodate the weighted Vietoris-Rips complex obtained from $\left(VR(X,r)\right)_{0\leq r\leq R}$, which represents a subset of the entire spectrum $\left(VR(X,r)\right)_{r\geq 0}$. \\
In the sequel, we assume that $(S,W)$ represents an admissible weighted simplicial complex on the vertex set $X$:
\begin{defi}\label{admissimp}
 A weight function $W$ on the simplicial complex $S$ on the  vertex set $X$, assigning to each simplex $\sigma \in S$ a non-negative finite weight $W_{\sigma}:=W(\sigma)$, is called \emph{admissible} if it satisfies the following conditions:
\begin{enumerate}
\item $W_{x}=0$ for every vertex $x\in X$, and for any other simplex $\sigma \in S$, $W_\sigma$ is strictly positive. 
\item The weight is non-increasing
when moving from a simplex to its facets.  
More restrictively, we can assume that the weight of each simplex is the maximum of the weights on its $1$-skeleton.
\end{enumerate} 
\end{defi}
\begin{rem}\label{violationoftriangle}
The weight function \rf{fuzzyvietoris} corresponding to the Vietoris-Rips filtration has the following additional property 
\begin{enumerate}
\item[3.] If $[x,z]\in S$, then for every $y\in X$ we have 
\bel{triangle}
W_{[x,z]}\leq W_{[x,y]}+W_{[y,z]},
\qe
where we assume an infinite weight for $[x,y]$ or $[y,z]$ when not appearing in $S$.
\end{enumerate}
However, this assumption is not necessary for metric realization. The process of metric realization resolves any violations of the triangle inequality by the weight function.\\
In fact, even if starting with a weighted simplicial complex $(X,W)$ on the vertex set $X$ that fail to satisfy \eqref{triangle}, the metric realization will ultimately define a metric on $X$ whose Vietoris-Rips filtration is identical with $S$ while the weight function $W$ is being improved to satisfy the triangle inequality \eqref{triangle}.    
\end{rem}
The process of metric realization consists of two main steps. First, we construct the metric realization of each simplex $(\sigma, W_{\sigma})$. Second, we glue different simplices together to obtain the realization of the entire simplicial complex $(S, W)$.\\
For each $k-$simplex $\sigma\in S$ with weight $W_{\sigma}$, we map $\sigma$ to $\Delta^k_{W_{\sigma}}$, the simplex whose vertices lie on the coordinates axes at a distance  of 
$W_{\sigma}$ from the origin, that is, 
\bel{simreal}
\Delta^k_{W_{\sigma}}=\{(t_0,\dots,t_k)\in\R^{k+1}|\sum_{i=0}^nt_i=W_{\sigma}: t_i\geq 0\}.
\qe
It is equipped with the induced Euclidean metric.\\ 
Thus, $\Delta^k_{W_{\sigma}}$ is isometric to $\Delta^k$ equipped with the Euclidean metric scaled by 
$W_{\sigma}$ 
and we denote this scaled metric by $d_{W_\sigma}$. \\
For the second step, i.e. gluing different simplices, we use the general procedure for gluing different metric spaces.  
In metric geometry, when gluing two or more metric spaces, the first step involves fixing the gluing parts, c.f. \cite{Burago01}. For simplicity, let's consider two metric spaces $(X, d_X)$ and $(Y, d_Y)$, although this procedure can be generalized to more than two spaces.
\begin{defi}\label{gluingmet}
Let $(X,d_X)$ and $(Y,d_Y)$ be two metric spaces. If there is a bijection $\phi: X'\subset X\to Y'\subset Y$, the \emph{gluing} of $X$ and $Y$ along $X'$ and $Y'$ is obtained as follows.\\
First, one takes the disjoint union $Z:=X\bigsqcup Y$ of $X$ and $Y$ as sets and defines an extended metric $d_Z$ on $Z$ as follows.
\be \label{eq:productMetric}\begin{split}
d_Z(z,z') &=\begin{cases}
d_X(z,z') , &z,z' \in  X \ \\
d_Y(z,z') , &z,z' \in  Y \ \\
\infty \ & \text{else.}
\end{cases}
\end{split}\end{equation}
Next, gluing is performed by identifying points $x \in X$ and $y \in Y$ whenever $(x, y)$ is on the graph of $\phi$. In other words, the bijection $\phi$ defines an equivalence relation, $x \sim y$ if $y = \phi(x)$. The gluing metric space is $({Z}/{d_\sim}, d_\sim)$ with the quotient metric $d_\sim$, which is defined by the following general procedure.\\
Let $(Z,d_Z)$ be a metric space with an equivalence relation $\sim$. We then let
\begin{equation}
\label{merge1}
d_\sim(p,q):=\inf \sum_{i=1}^\ell d_Z(p_{i-1},q_i)
\end{equation}
where the infimum is taken over all sequences of the  form $(p_0,q_1,\cdots,p_{l-1},q_l)$ with  $ p_0=p, p_\ell=q, p_i\sim q_i\text{ for }i=1,\dots \ell -1$. $d_{\sim}$ defines a pseudo metric on $Z$ and the quotient metric space $({Z}/{d_\sim}, d_{\sim})$ is obtained by identifying points with zero $d_\sim$ distance.  \\
\end{defi}

In the simplicial complex $(S, W)$, each simplex $\sigma$ carries a weight $W(\sigma)$, which may differ from the weights assigned to its corresponding boundary faces $\partial \sigma$. For example, if $\sigma = {x_{i_0}, \dots, x_{i_k}}$ is a $k$-simplex, then for each $0 \leq j \leq k$, the simplex $\partial_j \sigma := {x_{i_0}, \dots, \widehat{x_{i_j}}, \dots, x_{i_k}}$ represents one of its $(k-1)$-dimensional facets. These facets are mapped to their respective metric simplices $\Delta^{k-1}_{W_{\partial_j\sigma}}$ (or equivalently $(\Delta^{k-1}, d_{W_{\partial_j\sigma}})$) during the realization process. The following lemma then specifies the metric on the entire simplex obtained by attaching all the corresponding faces. 
\begin{lemma}
The mapping $\phi_j: \Delta^{k-1}_{W_{\partial_j\sigma}}\to \Delta^{k}_{W_{\sigma}}$, defined as
\be \label{metboundmap}
(t_0,\dots, t_{k-1})\mapsto \dfrac{W_{\sigma}}{W_{\partial_j\sigma}}(t_0,t_1,\dots,t_{j-1},0,t_{j+1},\dots, t_{k-1}),
\qe
 is a bijection onto a face of $\Delta^k_{W_{\sigma}}$.
 The resulting metric realization is the gluing metric of Definition
 \ref{gluingmet} induced by the gluing maps $\phi_j$ (which maps facets to their corresponding co-faces) and the identity map on shared faces (which attaches two simplices that share a face). 
\end{lemma}
\begin{pf}
Multiplying by $\dfrac{1}{W_{\partial_j\sigma}}$ maps
$\Delta^{k-1}_{W_{\partial_j\sigma}}$ to the standard simplex $\Delta^{k-1}$,
which is then mapped to the $j-$th face of $\Delta^{k}$ (obtained by excluding the vertex $e_{j+1}$) that finally is being mapped to the $j-$th face of $\Delta^{k}_{W_\sigma}$ by multiplying by $W_{\sigma}$. \\
This process effectively shrinks the $j$-th face of $\Delta^{k}_{W_\sigma}$ to become isometric to $\Delta^{k-1}_{W_{\partial_j\sigma}}$.
\end{pf}
\begin{rem}
The gluing of the corresponding face induced by $\phi_j$  results in the shrinking of that face proportional to its weight.\\
This process is iteratively extended to all the faces of a simplex. Therefore,
one can start from the $1$-skeleton and develop the gluing of $(S,W)$ by
iteratively attaching simplices. At the end, the distance between each pair of points is obtained by \rf{merge1}.  
\end{rem}
\begin{coro}
Since the weight decreases as traversing from simplex to each of its facets,
the distance between points in the interior of each simplex is obtained by
possibly traversing through its faces. Thus, the shortest path connecting two
interior points of a simplex never travels through its higher dimensional
co-faces. \\
For  pairs of vertices, as the exceptional case, the distance is realized by a shortest path consisting of edges (i.e. the graph distance on the $1-$skeleton). Therefore, in application, one could only restrict the realization to the realization of the $1-$skeleton of $(S,W)$ if, at the end,  the aim is to only obtain the restriction of this realization to the vertex set.  
\end{coro}

\begin{ex}
Starting from a fuzzy simplicial complex $(S,w)$ that satisfies the conditions
outlined in  Example \ref{expweight}, applying the function $\Psi=-log$
converts the strength function $w$ into a weight function $W$, thereby
yielding an admissible simplicial complex $(S,W)$. The metric realization of
$(S,W)$ is the same as that of $(S,w)$ as outlined in \cite{Spivak09}. 
\end{ex}

\section{Combining metric spaces}\label{Sec:merging}
In this section we will use the Vietoris-Rips filtration and the resulting
admissible simplicial complex model to combine two (or more) different metrics on the same set $X$. \\
 Suppose we have two metric spaces $X_1 := (X, d_1)$ and $X_2 := (X, d_2)$, both defined over the finite point set $X$. Additionally, let $\left(VR(X_i, r)\right)_{0\leq r\leq R_i}$ represent a portion of the Vietoris-Rips filtration for each metric space $X_i$, $i = 1, 2$. Correspondingly, we denote their resulting admissible simplicial complexes by $(S^i, W^i)$ for $i = 1, 2$.\\
To combine these two metric spaces, we need to define a process that combines their respective Vietoris-Rips filtrations and admissible simplicial complexes. The goal is to create a unified representation that incorporates the information from both metrics.
While we have a canonical way to directly combine $(S^1, W^1)$ and $(S^2, W^2)$,
we develop a probabilistic merging procedure for the corresponding  fuzzy simplicial complexes.\\
Such merging employs \emph{triangular conorms} (briefly \emph{t-conorms}) which are dual to \emph{triangular norms}
(briefly \emph{t-norms}), under the order-reversing operation that assigns $1-a$ to $a$ in $\left[0,1 \right]$. 
t-norms are binary operations introduced by Menger \cite{Menger42} for
generalizing transitivity criteria in the context of probabilistic metric
spaces. Later, Zadeh \cite{Zadeh65} used t-norms to generalize logical
operations in fuzzy logic, a particular class of  multi-valued logics. The definitions are as follows   
\begin{defi}\label{t-norm}
A t-norm is a function $T:\left[0,1 \right] \times \left[ 0,1 \right] \to \left[ 0,1 \right]
$, satisfying the following properties :
\begin{itemize}
\item Commutativity: $T(a,b)=T(b,a),$
\item Monotonicity: $T(a,b)$ is non-decreasing in either variable. i.e., $T(a,b)\leq T(c,d)$ if $a\leq c$ and $b\leq d,$
\item Associativity: $T(a,T(b,c))=T(T(a,b),c),$
\item Identity element $1$: $T(a,1)=a$
\end{itemize}
\end{defi}
The name triangular norm refers to the fact that, in the context of probabilistic metric spaces, t-norms generalize the triangle inequality of ordinary metric spaces, c.f. \cite{Menger42}. 
\begin{defi}\label{t-conorm}
A t-conorm is a binary operation
$T^{co}:\left[0,1 \right] \times \left[ 0,1 \right] \rightarrow \left[ 0,1 \right]$ 
that has the same properties as a t-norm, but its identity element is $0$ (i.e., $T^{co}(a,0)=a$). 
\end{defi} 
We can apply the t-(co)-norms iteratively and obtain something akin to a product
\begin{defi}
For $T$ either a t-norm or a t-conorm, and $\alpha_1,...,\alpha_n \in [0,1]$ we define their $T$-product  as follows
\be \label{Tproduct}
\begin{split}
\overset{T}{\prod}:&\underset{\text{n times}}{[0,1]\times [0,1]\times \dots \times [0,1]}\to [0,1], \ \\
\left(\alpha_1,\dots, \alpha_n\right)&\mapsto T(\alpha_1(T(\alpha_2,( \cdots T(\alpha_{n-1},\alpha_n))))). \
\end{split}
\qe
\end{defi}
Note that by associativity, a $T$-product does not depend on the order. We also note that the monotonicity carries over to this $T$-product:
\begin{lemma}\label{lemma:monotoneproduct}
If $\alpha^i \leq \beta^i$ for $1\leq i\leq n$, then $\overset{T}{\prod} \alpha_i  \leq \overset{T}{\prod} \beta_i.$
\end{lemma} 
When merging fuzzy simplicial complexes, this property ensures that the
resulting object remains a fuzzy simplicial complex. In other words, the
combination of probabilistic weights on each simplex does not exceed the
combination of probabilistic weights on any of its faces. One can define
  isomorphic counterparts of t-norms and t-conorms for combining
  metric-related weights. 
  For instance, The t-conorm $T(a,b):=\max\{a,b\}$, called the canonical t-conorm, has an isomorphic counterpart $\hat{T}:[0,\infty]\times[0,\infty]\to [0,\infty]$, with properties of commutativity, associativity, monotonicity and identity element $0$, defined by $\hat{T}(A,B):=\min\{A,B\}$. This t-conorm is the one that we will use in the IsUMap pipeline, as its metric counterpart facilitates the gluing of every pair of metric graphs $(G,W^1)$ and $(G,W^2)$ as outlined in Definition \ref{gluingmet}. \\
With a function $\Phi:[0,\infty]\to [0,1]$ that is strictly decreasing with $\Phi(0)=1$ and $\Phi(\infty)=0$, we convert the metric weights into probabilistic weights. We assign a probability weight of zero to simplices that do not appear in $(S,W)$. This means that simplices which are not present in $S$ are assumed to only appear at infinity. This way, the fuzzy simplicial complex corresponding to every admissible simplicial complex $(S,W)$ is the complete simplex on $X$ with the strength function 
\bel{eq:strengthfun}
\begin{split}
w(\sigma) &=\begin{cases}
\Phi(W(\sigma)) , &\text{if} \: \sigma \in S \ \\
0 \ & else
\end{cases}
\end{split}
\qe
Following this procedure, $(S^1,W^1)$ and $(S^2,W^2)$ are converted to fuzzy simplicial complexes $(S,w^1)$ and $(S,w^2)$. 
 Here, $S$ is the $(N-1)$-simplex with vertex set $X$. 
By selecting a t-conorm to merge $w^1$ and $w^2$, we can equip $S$ with a strength function $w$ to create a fuzzy simplicial complex as follows.\\
Let $T$ be a t-conorm, and denote $S_k$ as the set of $k$-simplices of $S$ for $k=0,\dots, N-1$. Then, defining $w(\sigma):=T(w^1(\sigma),w^2(\sigma))$ assigns a strength  to each simplex $\sigma$. This operation, in line with Lemma \ref{lemma:monotoneproduct}, ensures that $(S,w)$ forms a fuzzy simplicial complex. In fact, $w^i(\sigma)$ (for $i=1,2$) for each $\sigma\in S_k$ is at most the minimum of the strengths of its corresponding facets, i.e. $w^i(\sigma)\leq w^i(\partial_j\sigma)$ for $j=0,\dots, k-1$. Consequently,  Lemma \ref{lemma:monotoneproduct} implies that $w(\sigma)=T(w^1(\sigma),w^2(\sigma)) \leq w(\partial_j\sigma)=T(w^1(\partial_j\sigma),w^2(\partial_j\sigma))$. \\
To convert the resulting fuzzy simplicial set into an admissible simplicial
complex $(S^1\bigcup S^2)$,  the strength function $w$ has to be converted
into a distance-related weight function $W$ on $S^1\bigcup S^2$, as outlined
in Definition \ref{admissimp}, for example with the map $\Phi^{-1}$. The strict monotonicity of the map $\Phi$ and the fact that $\Phi(0) = 1$ ensure that conditions 
1 and 2
in Definition \ref{admissimp} are satisfied, respectively.  
The 3rd condition in eq. (\ref{triangle}), however, may fail to be satisfied after merging the fuzzy simplicial complexes and converting the resulting probabilistic weight function $w$ back with $\Phi^{-1}$.  For
instance, even if both $w^1$ and $w^2$ satisfy the criteria outlined in
Example \ref{expweight}, it is not necessarily true that the combined strength
function $w:=T(w^1,w^2)$ for a t-conorm $T$ will still meet the same criteria
to make it suitable for conversion through the $-\log$ map that we shall use
below. \\
This, however, does not pose any problem, as the metric realization will address this issue, as referenced in Remark \ref{violationoftriangle}.

\section{Category theoretical formulations}
\label{sec:CatTheory}

In this section, we briefly draw a connection to category theoretical formulations of previously elaborated concepts. However, this is only intended to be a short and rather informal outline and we refer the reader to \cite{otherPaper} for a more detailed exposition. In general, the category theoretical approach can guide the search for appropriate mathematical constructions and can provide a structurally rich language to relate objects from different mathematical areas.

\bigskip

In the context of the present article, the two important categories involved are the category of uber-metric spaces\footnote{An uber metric space is a metric space, where distances can be infinite.} $\mathbf{UM}$ or extended-pseudo metric spaces\footnote{An extended-pseudo metric space is an uber metric space, where the triangle inequality does not need to hold for infinite distances, cf. Section
\ref{preliminaries}.} $\mathbf{EPMet}$, and the category of fuzzy simplicial complexes $\mathbf{compFuz}$ or fuzzy simplicial sets\footnote{Roughly speaking, a fuzzy simplicial set is a fuzzy simplicial complex, in which simplices have 
\vio{orientation} and which contains all possible degeneracies of each simplex.} $\mathbf{sFuz}$. Using those categories, a couple of operations, that were described above, find a very natural category theoretical description.\\
First of all, the gluing of generalized metric spaces, as described in equation \eqref{merge1}, can be shown (cf. \cite{otherPaper}) to be equal to the colimit of uber metric spaces\footnote{Note that this colimit is not defined for metric spaces without infinite distances because there are not even coproducts (as defined in \eqref{eq:productMetric}) for those. That is one motivation to introduce infinite distances from a category theoretical point of view: It closes the category under colimits.}. Even more importantly, there is an adjunction\footnote{An adjunction can roughly speaking be thought of as a weak version of an equivalence between two categories and can be very useful for transferring mathematical constructions between different categories. Formally, an adjunction between the categories $\mathbf{A}$ and $\mathbf{B}$ is described by two functors, $F:\mathbf{A}\to\mathbf{B}$ and $G:\mathbf{B}\to\mathbf{A}$ that fulfill certain conditions, which allow those functors to be restricted to an equivalence on appropriate subcategories of $\mathbf{A}$ and $\mathbf{B}$.} between $\mathbf{UM}$ and $\textbf{sFuz}$, as shown in \cite{Spivak09}. 
In fact, one can show that there are infinitely many adjunctions between those two categories (namely, for each strictly decreasing $\Phi$ as defined below Definition \ref{fuzzyvietoris}, we obtain one such adjunction) and that there are similar adjunctions between $\mathbf{EPMet}$ and $\mathbf{sFuz}$ (cf. \cite{otherPaper}). One direction of this adjunction is called Sing-functor\footnote{in analogy to the singular homology functor in the context of algebraic topology} and one writes, for example, $\te{Sing}:\mathbf{UM}\to \mathbf{sFuz}$. It turns out that such Sing-functors extract (oriented versions of) the Vietoris-Rips filtration from a (uber or extended-pseudo) metric space and therefore entail what was denoted by $\left(VR(X,r)\right)_{r\geq 0}$ above. Even better, the other direction of the adjunctions corresponds to metric realizations of fuzzy simplicial sets, and is therefore referred to as metric-realization-functor, and we write, for example, $\text{Re}:\mathbf{sFuz}\to\mathbf{UM}$. This metric realization functor employs the colimit in the category $\mathbf{UM}$, which, as mentioned above, corresponds to the gluing operation \eqref{merge1}, and applies it to the geometric realization of the simplices that compose the simplicial set, which results in the geometric realization of the entire set (or complex), just as it was described at the end of Section \ref{sec:metricrealization}. 
During this process, the geometric realization of the simplices can be chosen comparatively freely, which is ultimately the reason why there are so many adjunctions between the geometric and the fuzzy simplicial categories. One specific example is the realization described in \eqref{simreal} but other choices that interpolate the datapoints differently, are possible as well. In fact, IsUMap described below uses this freedom to choose an adjunction that has favourable computational properties.\\
An important aspect of the adjunctions mentioned above is that they can be used to show that $\mathbf{UM}$ and $\mathbf{EPMet}$ can be embedded as subcategories into $\mathbf{sFuz}$. On the one hand, this shows that $\mathbf{sFuz}$ provides a general setting for data analysis that encompasses different metric definitions, and on the other hand, it establishes the fact that one does not lose any information when going through the adjunction in the direction \begin{tikzcd} \mathbf{UM}\ar{r}{\text{Sing}}&\mathbf{sFuz}\ar{r}{\text{Re}}&\mathbf{UM}\end{tikzcd}. The latter fact is useful because it ensures that one can perform certain combinatoric or probabilistic operations $O$ on the fuzzy simplicial set $\text{Sing}(X,d)$ corresponding to a metric space $(X,d)$, and then map the result back to a metric space $\text{Re}(O(\text{Sing}(X,d)))$, without losing information that one would not have lost when performing the operation inside $\mathbf{UM}$. An important example of such an operation $O$ is the t-conorm, that can naturally be used to define a merge operation $\text{merge}_{\mathbf{sFuz}}$ of two fuzzy simplicial sets or complexes as explained in Section \ref{Sec:merging}. The category theoretical description of the adjunction then allows to translate this merge operation to the category $\mathbf{UM}$ without loss of information (because $\mathbf{UM}$ is a subcategory of $\mathbf{sFuz}$), simply by defining the combination operation in $\mathbf{UM}$ as the composition of three functors, namely $\text{merge}_{\mathbf{UM}}:=\te{Re}\circ\te{merge}_{\mathbf{sFuz}}\circ \te{Sing}^2:\mathbf{UM}\times_{\mathbf{Set}}\mathbf{UM}\to\mathbf{UM}$. This composition then exactly results in (a generalized version of) what was described in Section \ref{Sec:merging}. In particular, the gluing operations in the metric space categories that we are employing then is achieved by applying the realization functor to fuzzy simplicial sets that were merged with t-conorms.\footnote{One can see here that infinite distances in the generalized metric spaces play two roles: On the one hand, they ensure that coproducts and hence colimits exist and on the other hand, they correspond to vanishing probabilities in the setting of t-conorms, that make certain formulations simpler.} It is remarkable that the categorical constructions have such a concise description, while reproducing well-known notions from metric geometry and persistent homology. The interested reader is again referred to \cite{otherPaper} for further details.
\section{IsUMap}\label{sec:IsUMap}
We consider a discrete data set $X=\{x_1,x_2,\dots ,x_N\}$, equipped with some
distance function $d_X(.,.)$ between the data points. \\
A key assumption is that the data exhibits certain dependencies and regularities, suggesting that it has fewer intrinsic degrees of freedom than initially apparent. Additionally, continuity plays a critical role, implying that the data can be interpolated, which is useful for inferring positions of points that have not yet been sampled.  
When data is initially presented as a subset of $\R^n$, increasing the number of dimensions $n$ results in an exponential growth for the amount of data required to efficiently sample the space, a phenomenon known as "curse of dimensionality". This effect leads to data sparsity, which, in turn, complicates the recognition of meaningful patterns and structures.
 The goal is to represent the original metric information, at least at a local scale, in a low-dimensional Cartesian space (typically equipped with the Euclidean metric) to facilitate specific tasks such as clustering or exploratory data analysis.\\ 
One approach is to assume that the data points lie on, or near, a smooth submanifold $M$ of some low dimension $m$ (which is assumed to be much less than $n$, if $X$ is initially given as $X\subset \R^n$). When $M$ is close to an affine linear subspace, linear dimension reduction techniques like Principal Component Analysis (PCA) can effectively project the data into a lower-dimensional space, such as $R^m$. \\
However,  $M$ may be nonlinear, that is, a curved submanifold of $\R^n$ and
stretch into many, if not all ambient dimensions which may already occur in the case where $m=1$, i.e.\ where $M$ is a space curve. This issue is  handled by manifold
learning schemes, which seek to recover the intrinsic metric geometry of 
$X$ incorporating local geometric information (e.g. Isomap \cite{Tenenbaum00}), or Laplacian Eigenmaps \cite{Belkin03}). 
While these manifold learning methods primarily focus on local structure, they
may struggle when the data is distributed non-uniformly across the manifold. This can cause a major limitation in many manifold learning approaches, as they are grounded on the assumption of uniform distribution. When this assumption does not hold, it can result in inaccurate geometric recovery.\\
One way to address this issue from a geometric perspective is by applying conformal transformations to the metric. This involves adjusting the metric by a factor that is large in regions of high density, where many data points are close together, and smaller in regions of low density, making distances to nearest neighbors more uniform. However, implementing such a transformation for discrete data is generally impossible.\\
To work around this limitation, we attempt to mimic these conformal changes in the construction of a combinatorial model for $(X,d)$. This approach aims to indirectly incorporate the effects of a conformal transformation, even in discrete settings, to better reflect the underlying geometric structure of the data.\\
The combinatorial object we use for modeling $(X,d)$ is the $k$-neighborhood
graph, a common structure in manifold learning techniques. This graph
represents relationships between data points by connecting each point to its
$k$ nearest neighbors, determined by  the given metric $d$. It helps to capture local structure and proximity in the data, providing a foundation for subsequent analysis and manifold learning tasks.\\
In different manifold learning techniques, this graph may be weighted in various ways. In the context of IsUMap, the weights are tied to the Vietoris-Rips filtration, derived from a metric that results from combining
the  metric components $d_i$. This process incorporates local distortions of the original metric $d$ around each $x_i\in X=\{x_1,\dots, x_N\}$.\\
Our algorithm (IsUMap) shares similarities with Isomap, introduced in \cite{Tenenbaum00}. 
In fact, when we forgo local modifications of the initial distance function, our approach essentially becomes an implementation of the Isomap algorithm. 
However, our method offers additional flexibility by allowing the use of arbitrary t-conorms, 
and facilitating data uniformization such as that used in UMAP
\cite{McInnes18}. Moreover, our theoretical analysis demonstrates that this
approach leads to an embedding that is theoretically equivalent to the
framework proposed in \cite{McInnes18}, establishing a close connection
between the two methods that was not apparent previously.  This connection
inspired the name IsUMap  of this algorithm, reflecting its blending of Isomap and UMAP, and the fact that it employs the category $\mathbf{UM}$ of uber metric spaces as explained in Section \ref{sec:CatTheory}.\\
In the following, we first present a simplified outline of the IsUMap
pipeline, focusing on intuitive descriptions while avoiding complex
theoretical explanations.

\subsection{The IsUMap algorithm}\label{sec:concreteAlgo}
\begin{enumerate}
\item  Given a metric space $(X,d)$, we construct (extended pseudo-) metric spaces $d_i$, 
\begin{eqnarray}
  \label{localmet}
  d_i(x_i,x_{i_j})=d_i(x_{i_j},x_i)&=&\frac{d(x_i,x_{i_j})-\rho_i}{\sigma_i}\quad
                               \text{ for }j=1,\dots k\\
  \nonumber
  d_i(x,x)&=& 0\quad \text{ for all }x\in X_i\\
  \nonumber
  d_i(x_j,x)&=&\infty \quad \text{ in all other cases.}
\end{eqnarray}
Here, $\sigma_i$ and $\rho_i$ are two hyperparameters. \\
The weighted simplicial complex corresponding to Vietoris-Rips of $(X,d_i)$ is the star graph, which we denote by $\Gamma_i$, with $x_i$ as the center and its $k$ nearest neighbors $x_{i_j},\; j=1,\dots k$ as the outer vertices and $W_{ij}:=d_i(x_i,x_{i_j})$  the corresponding weight on the edge between $x_i$ and $x_{i_j}$. 
\item Bringing together all the existing distances in an $N\times N$ matrix $A$ with entries $d_{ij}=d_i(x_i,x_j)$ yields a non-symmetric sparse matrix.  In each row $i$ all elements are $\infty$ except for $k$ elements corresponding to the $k$ nearest neighbors of $x_i$. \\
This matrix corresponds to the weight matrix of the union of the star graphs $\Gamma_i$, $i=1,2,\dots,N$, which could be considered either as a multi-graph (allowing for multiple edges between pairs of nodes) or a directed graph. 
 \item In the next step, we apply the canonical t-conorm described in Section \ref{Sec:merging}. Practically, this means we symmetrize this matrix according to  the following procedure.
    If either $d_i(x_i,x_j)$ or $d_j(x_i,x_j)$ is finite, we simply take the smaller value for both of them. Thus,  we put
  \[
  d_{ij}=d_{ji}:=\min\{d_i(x_i,x_j),d_j(x_i,x_j)\}
\]
as the new values.
    This operation  establishes symmetric connections between points $x_i$ and
    $x_j$ whenever at least one weighted edge exists between them in the union
    of the star graphs.  If multiple edges are present, we combine them into one by selecting the smallest weight as the weight on this edge. This process reduces the sparsity of the distance matrix, as some $\infty$ entries are replaced by a finite value.
    \item  To complete the distance matrix, replacing the remaining entries with $\infty$ with finite values, we utilize Dijkstra's algorithm, which corresponds to the gluing operation described in equation \eqref{merge1}, in particular to the computation of the infimum over all paths. During this computation, a path of infinite length is substituted by the length of a path composed of a sequence of edges with finite weights. Dijkstra's algorithm computes these shortest paths, thereby completing the distance matrix and enabling accurate distance calculations between all pairs of points in the dataset.\\
\item  After completing the distance matrix, we utilize the multidimensional
  scaling (MDS) algorithm to find the desired embedding in the Euclidean space
  of dimension $m$
  . We have implemented both classical and metric MDS variants in our algorithm.
\end{enumerate}

 \subsubsection*{Complexity of the algorithm}
The step where we compute the missing entries of the weight matrix $A$ by
replacing the remaining $\infty$ values with finite ones using Dijkstra's
algorithm (Step $4$) turns out to be the most time consuming one in the pipeline.
 It has a complexity of
\begin{equation*}
           \mathcal{O}(N(N\log(N)+E)),
   \end{equation*}
where $N$ is the number of vertices of the Dijkstra graph, corresponding to
the number of points in the metric space $(X,d)$, and $E$ is the number of
edges, corresponding to the number of finite entries in $A$ after symmetrization. \\
Since, even after symmetrization,  our matrix is very sparse because most entries are infinite, Dijkstra is still quite efficient for our application. 
In fact, since we use a $k$-nearest neighbourhood algorithm to split the metric space into $N$ copies, each of which has at most $k$ non-infinite distances, the number of edges $E$ is bounded by $kN$ (or $2kN$ after symmetrization with the t-conorm), where $k$ is typically very small and chosen independently of $N$. Even if we assume a weak dependence of $k$ on $N$, say $k ~\propto \log(N)$, we end up with $E~\propto N\log(N)$ and hence with an overall complexity of approximately
\begin{equation}
    \begin{split}
        \mathcal{O}( N^2 \log (N) ).
    \end{split}
    \label{eq:overallComplexity}
\end{equation}
This quadratic complexity  admittedly  slows the algorithm down
compared with other algorithms like UMAP 
that has a
complexity that is only slightly higher than $\mathcal{O}(N^{1.14})$.
However, we implemented a \textit{parallel Dijkstra} algorithm, which can reduce the time approximately by the number of processors used. With this implementation, one can still tackle somewhat larger datasets in a reasonable amount of time.\\
For classical MDS, which can be solved by eigenvalue decomposition, a high
efficiency can be reached when using the Lanczos algorithm because it can
compute the decomposition element-wise and hence, for a visualization in $2$
or $3$ dimensions, only has to compute $2$ or $3$ such eigenvectors. When using metric MDS, we optimize via stochastic gradient descent (implemented with pytorch) to find a local minimum, that often delivers good results when initialized with  LE, or classical MDS, and hence generally also does not have a computational complexity that exceeds the one of Dijkstra for larger datasets.
The overall complexity of the algorithm is thus bounded by the complexity of the
parallel Dijkstra implementation.\\
\vio{The open-source implementation of the IsUMap algorithm is available on GitHub. The repository can be accessed at \cite{github_repository}.}

\subsection{The IsUMap pipeline: A structural overview}\label{sec:IsUMapPipeline}
This subsection explores the IsUMap pipeline, providing a comprehensive elaboration of the  steps from various perspectives, including geometry and combinatorics. It closely aligns with the theoretical framework of merging metric spaces and the metric realization of admissible simplicial complexes, as described in Section \ref{Sec:merging} and Subsection \ref{sec:metricrealization}.\\
\begin{enumerate}
\item \textbf{Local metrics.} The first step involves constructing local
  metrics $d_i$, for $i=1,\dots,N$,  as presented in \ref{localmet}.  This
  approach, as highlighted in the section's introduction, aims to address the
  issue of non-uniform data distribution on the manifold through a conformal
  transformation of the metric and to mitigate the curse of dimensionality. \\
When we have  a finite metric space $(X,d)$, derived from a sample drawn
from a Riemannian manifold, one can approximate the radial distances of that
manifold in the local neighborhood around each point $x_i\in X$s. This is
achieved by identifying the $k$ closest points $\{x_{i_1},\dots, x_{i_k}\}$,
and measuring their distances in the metric $d$ to the given point $x_i$. A
suitable choice of $k$ ensures that these distances approximate the geodesic
distances on the manifold, i.e.\ the distances induced by a Riemannian metric
tensor. For the relevant geometric theory, see \cite{Joharinad23, Jost17a}. \\
These geodesic distances represent the first coordinate in a special
coordinate system,  the normal coordinate system introduced by Riemann, which we describe in the following. \\
Let $(M,g)$ be a (finite-dimensional) Riemannian manifold. For every $p\in M$ and every $ V\in T_pM$, the tangent space at $p$, there exists some
$\delta>0$ and a shortest geodesic 
\begin{equation}
\label{rie8}
c_V:[0,\delta]\to M, c_V(0)=p, \frac{d}{dt}c_V(0)=V.
\end{equation} 
 This geodesic segment, as it extends from $p$ to an endpoint, represents the distance between these two points. \\
 When $c_V$ is defined on $[0,\delta]$, then $c_{\delta V}$ is defined on
$[0,1]$.
These considerations imply that $U_p :=  \{ V \in T_p M : c_V \mbox{ is
defined on }[0,1]\}$ contains some neighborhood of the origin $0\in T_pM$.
\begin{defi}\label{def:exp}
Let  $p \in M$, where $M$ is a Riemannian manifold.
\begin{eqnarray}\nonumber
\exp_p :& U_p \to M\\
\label{rie11}
&V \mapsto c_V(1)
\end{eqnarray}
is called the \emph{exponential map} of $M$ at $p$.
\end{defi}
The exponential map thus maps every $V\in U_p\subset T_p(M)$ to the endpoint of a geodesic of
length $\|V\|$ that starts at $p$  in the direction $V$.\\
In particular, the derivative of the exponential map  at $0\in T_pM$ is the
identity, as the image of $V\in T_pM$ is the tangent vector of the geodesic
starting in the direction $V$, that is, $V$ itself. \\
It therefore maps some neighborhood of the origin $0\in T_pM$ diffeomorphically onto a neighborhood of $p\in M$. The coordinates resulting from the inverse of the exponential map are known as \emph{normal coordinates} (or \emph{Riemann normal coordinates}) at $p$.  \\   
If we convert from Cartesian (Euclidean) (i.e.\ $(x^i)$) to polar coordinates (i.e.\ $(r,\varphi^j)$), where
$r=\sqrt{\sum (x^i)^2}$ represents the distance from the center $p$ and
$\varphi$  are the spherical angles that define points on a distance sphere, we create Riemann polar coordinates. These coordinates provide a way to visualize the manifold from the perspective of a single point.\\
When using a $k$-neighborhood graph to model $(X,d)$, with edges weighted according to the distances between the corresponding points, we are effectively capturing the first coordinate in the normal coordinate system at each point $x_i$. 
This coordinate $r$ scales when we apply a conformal change to the Riemannian
metric tensor $g$, by scaling the inner-product $g_p$ on the tangent space $T_pM$ with  a positive scalar $\lambda(p)$.  A conformal transformation involves replacing  $g$ with
$\lambda g$, where $\lambda$ is a positive smooth function on $M$. This transformation scales the norm of tangent vectors by a positive factor, as the norm is determined by the inner product defined on each tangent space by the Riemannian metric tensor. Importantly, while the magnitudes of vectors are scaled, the angles between any pair of vectors in the tangent space remain unchanged by the conformal transformation. Again we refer to \cite{Jost17a} for more on this.\\
The polar coordinate system around $p$ yields a star graph on $T_pM$, capturing the configuration of the neighboring points in the sample. This configuration magnifies or shrinks by scaling the inner product on $T_pM$. After applying this pointwise modification, one 
could map the resulting star graph back to the manifold using the initial geometry, specifically the exponential map $\exp_p$  corresponding to the metric $g$ (i.e. \rf{rie11}). This process, however, moves points along the geodesics connecting them to $p$, either toward or away from $p$. This step, which incorporates the geometric configuration around each point on a local scale, suggests that to obtain geodesics under the new metric, a correction considering the global structure is necessary. \\ 
By choosing a point-dependent scaling factor $\lambda(x_i)=:\frac{1}{\sigma_i}$ in \ref{localmet}, the metric can be adjusted such that it increases in regions with high data density, where many data points are closely packed, and decreases in regions with lower density. This approach aims to make distances between nearest neighbors more uniform.\\ 
Given this scaling factor, the choice of $\sigma_i$ becomes crucial for
effective normalization that aligns with density estimation. One common
approach, as used in UMAP \cite{McInnes18}, sets $\sigma_i$ such that it
equalizes a certain property across all data points. This method involves
choosing $\sigma_i$ with a binary search algorithm so that 
\begin{equation}
  \begin{split}
    \sum_{j=1}^k \exp\br{-d_i(x_i,x_{i_j})}=\log_2(k).
  \end{split}
  \label{eq:normalizationStarGraphs}
\end{equation}
This normalization is very similar to, and perhaps inspired by, the one used in t-SNE, cf. \cite{vandermaaten08a}.\\
However, in our case, we point out that our normalization of $\sigma$ adheres more strictly to the original theoretical idea in UMAP of uniformizing the distribution of the data. We simply set $\sigma_i$ to be the distance to the $k$th neighbor, i.e.\
\bel{sigma_i}
           \sigma_i := d_i(x_i,x_{i_k}).
  \qe
Since this conformal factor $\sigma_i$ varies as we change the point $x_i\in
X$, the corresponding weight $W_{ij}=d_i(x_i,x_j)$ on the $k$-neighborhood
graph could be different from $W_{ji}=d_j(x_i,x_j)$ even if each of them is
among the $k$ nearest neighbors of the other.\\  
This difference could become even larger when  subtracting  $\rho_i$, chosen to be the distance to the first nearest neighbor  $x_{i_1}$.\\
Therefore, we consider this multi-graph (or directed graph) as the union of star graphs $\Gamma_i$, for $i=1,\dots,N$. Each star graph represents the weighted simplicial complex associated with the Vietoris-Rips filtration of the metric space $(X,d_i)$. 
\item\textbf{Combining local distances.} From a geometric point of view, our aim is to merge the Vietoris-Rips filtrations of metric spaces $\{(X,d_i), \; i=1,\dots N\}$ , to create a final filtration that corresponds to the Vietoris-Rips filtration of a distance function on $X$. This final distance function is achieved by gluing the individual metrics $d_i$ along their common parts, which is the original set $X$. This process involves steps 2-4 in the IsUMap pipeline.\\
To this end, the algorithm employs the canonical merging process utilized by
the metric t-conorm, where instead of applying t-conorm to the weights of simplices in the fuzzy setting one directly applies the dual operator to the metric weights, c.f. \cite{Simas21}. This method is particularly useful because it can be applied directly to the set of star graphs $\{\Gamma_i: \; i=1,\dots N\}$,  where each star graph represents the Vietoris-Rips filtration for a local metric $d_i$. The merging process does not require converting the metric-based weights $d_{ij}=d_i(x_i,x_{i_j})$ to probabilistic ones, allowing for a more straightforward and geometrically meaningful combination of the different metric spaces. However, alternative t-conorms can be applied if desired by transforming the weights, such as by  $d_{ij}=:W_{ij}\mapsto w_{ij}:=\exp\br{-W_{ij}}$. We will present some results of applying IsUMap with different t-conorms in Section \ref{sec:illustrations}.  \\
The canonical t-conorm, also known as the Maximum t-conorm, assigns to each pair $(a,b)\in [0,1]\times [0,1]$ their maximum,  $(a,b)\mapsto \max\{a,b\}$.  Its metric counterpart, used in IsUMap, assigns the minimum to each pair. This can be interpreted in terms of metric gluing, where a simplex is merged across multiple Vietoris-Rips filtrations by choosing the smallest scale at which the simplex appears in at least one of the filtrations. This smallest scale becomes the emerging scale in the final filtration, ensuring that the scale at which a simplex emerges in the final filtration corresponds to its earliest appearance in any of the individual filtrations.\\
The final metric is derived through the metric realization of the admissible simplicial complex that arises from the combination of these star graphs, as detailed in Subsection \ref{sec:metricrealization}. \\
The merging process via t-conorm yields the symmetrization of the incomplete
distance matrix $A$ in step 3 of the IsUMap pipeline.  The metric realization
yields the completion of $A$ in step 4 through the Dijkstra algorithm to determine the shortest paths between all pairs of vertices. 
\item\textbf{Low dimensional embedding.} After constructing the metric
  realization of the admissible simplicial complex (in our case an admissible
  graph), and restricting the distance function to the sample $X$ (which
  results in the completed metric matrix $A$), the next step would be to
  visualize $X$ in a low dimensional Euclidean space while preserving the
  mutual distances to the extent possible.\\
At this stage, we have two  options: One is to follow a methodology akin to UMAP, utilizing techniques like a force-directed graph layout or stochastic gradient descent to construct the embedding based on the information recorded in the fuzzy counterpart graph.  In this approach, it is not necessarily required to complete the metric matrix $A$ after symmetrization, as the fuzzy graph can serve as a sufficient basis for embedding.\\
 An alternative approach that is geometrically more resilient can employ multidimensional scaling (MDS) techniques.
 Both approaches have their advantages. The UMAP-like methodology offers
 flexibility and is popular for visualizing complex data with minimal
 computational overhead. MDS, on the other hand, provides a more
 mathematically grounded way to maintain distance relationships, making it
 particularly useful when we want to preserve distances. 
\end{enumerate}

\section{Illustrations}\label{sec:illustrations}
In our empirical experiments, we have consistently observed that our algorithm performs precisely as intended according to its theoretical design. Specifically, it effectively generates a low dimensional representation of a uniformized dataset.\\
We first apply IsUMap to several standard examples, and compare it  with various well known techniques such as Isomap, LE, UMAP and t-SNE.

\subsection{Low-dimensional geometric examples}
We first test it on some toy examples, illustrated in \cref{tab:IsUMap}, samples of size $3000$ from  \emph{Swiss Roll}, \emph{Swiss Roll with a hole} and \emph{Torus},  along with a sample drawn from \emph{Mammoth} with  $20000$ points in $3$-dimensional space. \\
Incorporating  UMAP's emphasis on local relationships and Isomap's geodesic distance computation, IsUMap effectively unfolds the Swiss Roll dataset ((b.1) in \ref{tab:IsUMap}) while preserving both local and global structures and maintaining cluster integrity. In comparison, Isomap relies on geodesic distances, and the discontinuous visualization by UMAP highlights its emphasis on preserving local structures at the expense of overlooking some global patterns.\\
The result of IsUMap with metric MDS on Swiss Roll with a hole (image (b.2) in \ref{tab:IsUMap}) reveals how it successfully unfolds the data similarly to the Swiss Roll while preserving the topological feature (the hole) of the dataset and also maintaining the integrity of clusters within the data. \\
Image (b.3) in \ref{tab:IsUMap} shows the visualization of the Torus dataset using IsUMap, showcasing its ability to preserve both the shape and topological features. This projection is akin to Isomap's result (see image (c.3) in \ref{tab:IsUMap}). Notably IsUMap offers a more accurate representation compared to UMAP (see image (d.3) in \ref{tab:IsUMap}).\\
The IsUMap outcome on the Mammoth dataset (image (b.4) in \ref{tab:IsUMap})
well preserves   the clusters and features like legs, tusks, body, and bones
by preserving geodesic distance. The different body protrusions are clearly
spread out, and the results are  significantly better than
those of the other two schemes, UMAP (image (d.4)) and Isomap (image (c.4)).
\begin{table}[H]
    \centering
    \tiny
    \begin{tabular}{>{\centering\arraybackslash}m{1cm}|*{4}{>{\centering\arraybackslash}m{2cm}}}
        & \textbf{Data set} & \textbf{IsUMap} & \textbf{Isomap}  & \textbf{UMAP} \\
        & (a) & (b) & (c) & (d)  \\
        \hline \\
        \textbf{(1) Swiss Roll} & \includegraphics[width=0.17\textwidth]{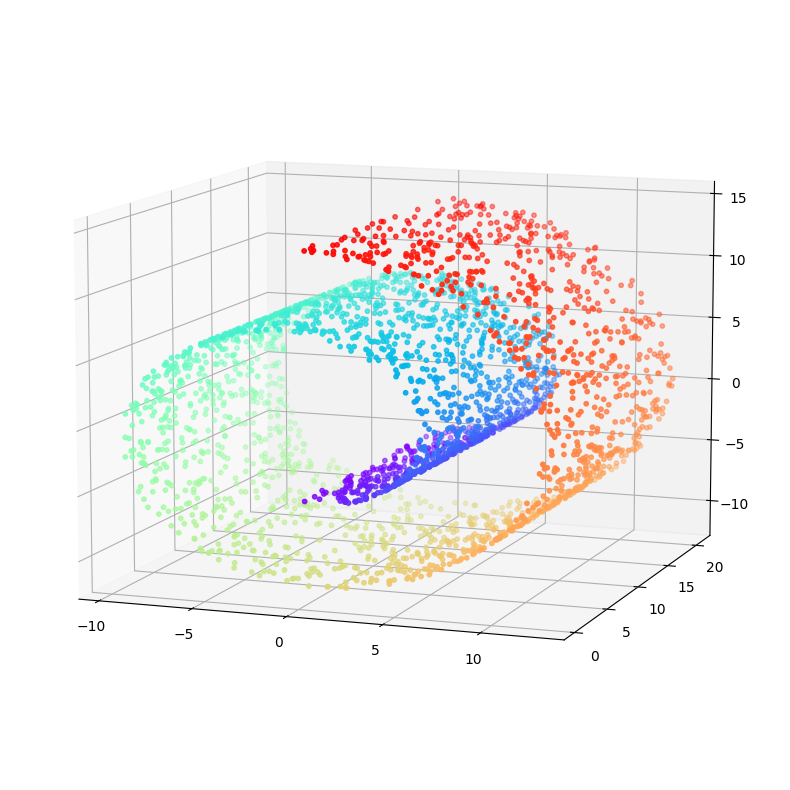} & \includegraphics[width=0.17\textwidth]{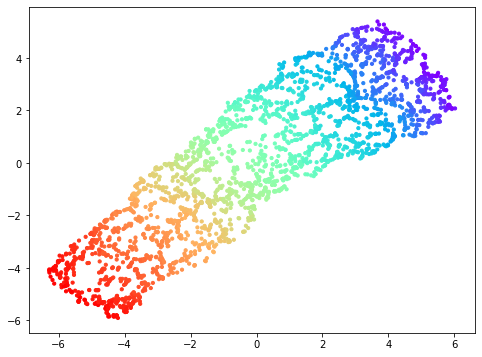} & \includegraphics[width=0.17\textwidth]{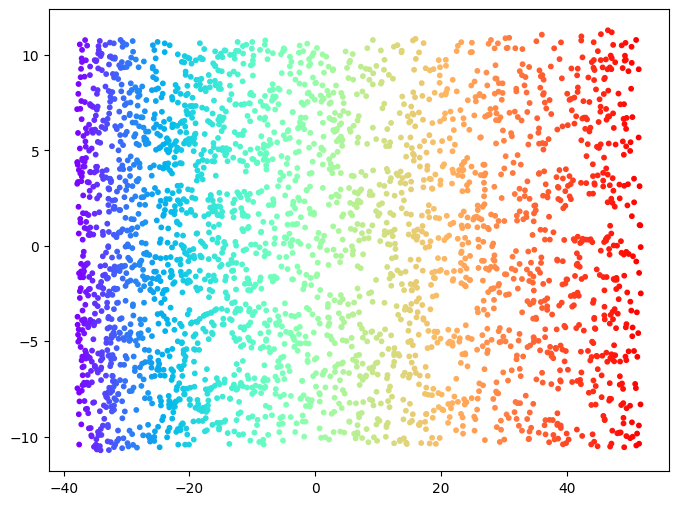} & \includegraphics[width=0.17\textwidth]{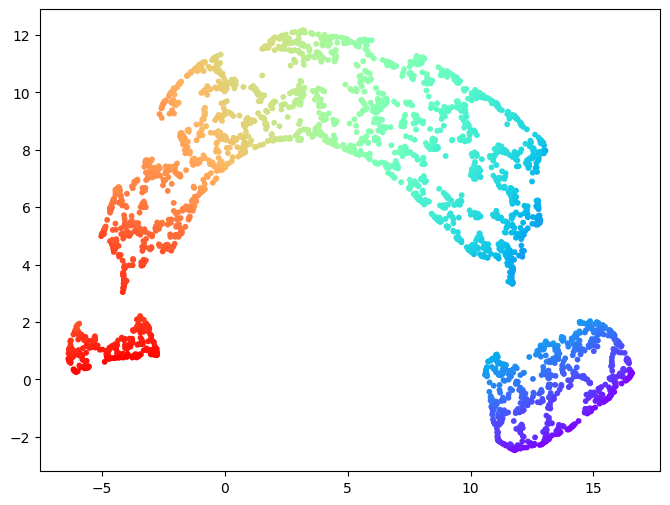} \\
        \textbf{(2) Swiss Roll with a hole} & \includegraphics[width=0.18\textwidth]{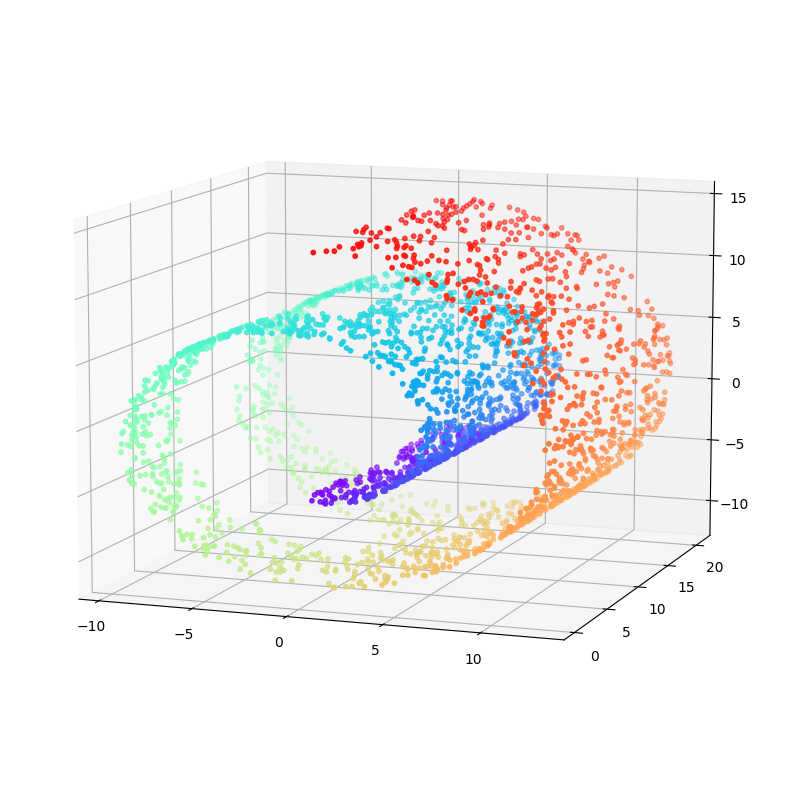} & \includegraphics[width=0.17\textwidth]{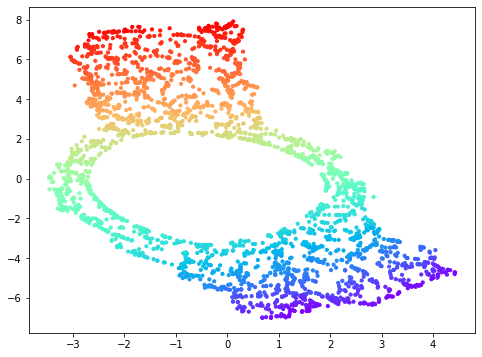} & \includegraphics[width=0.17\textwidth]{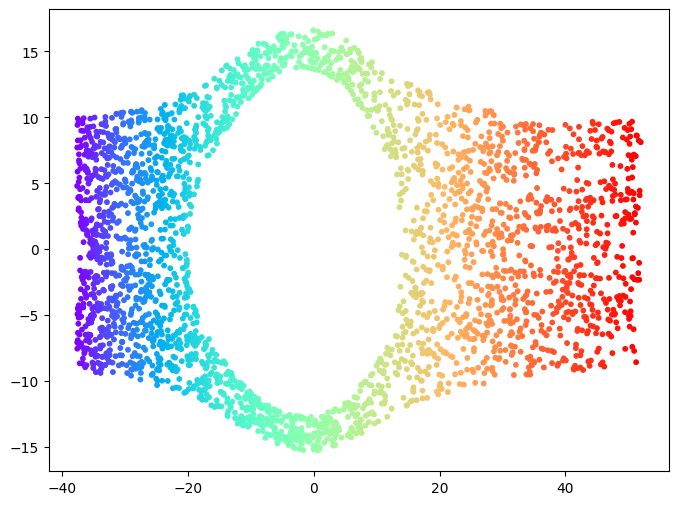} & \includegraphics[width=0.17\textwidth]{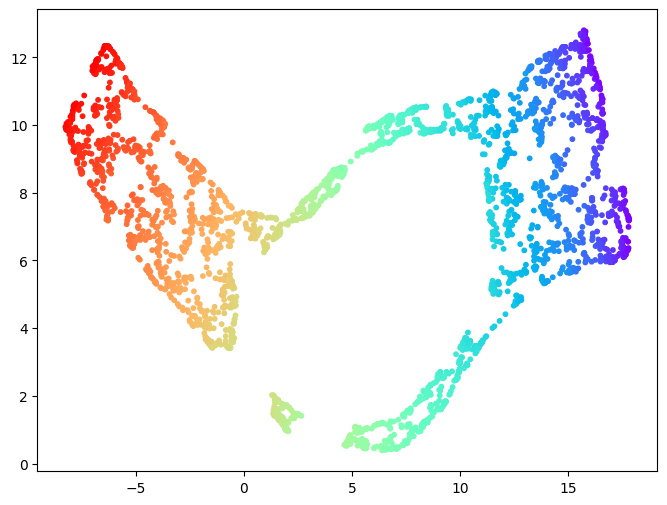} \\
        \textbf{(3) Torus} & \includegraphics[width=0.17\textwidth]{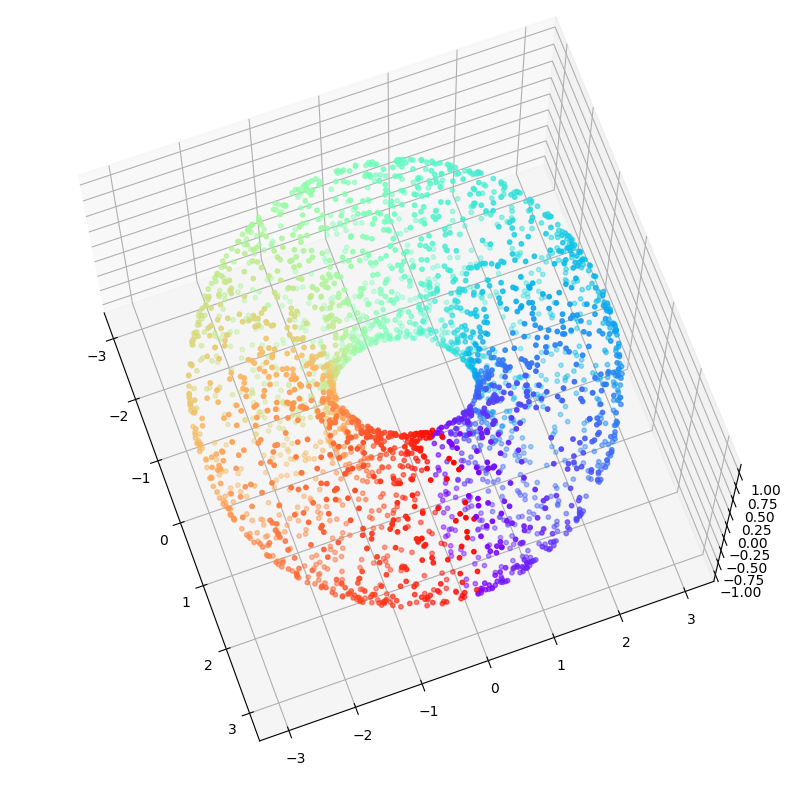} & \includegraphics[width=0.17\textwidth]{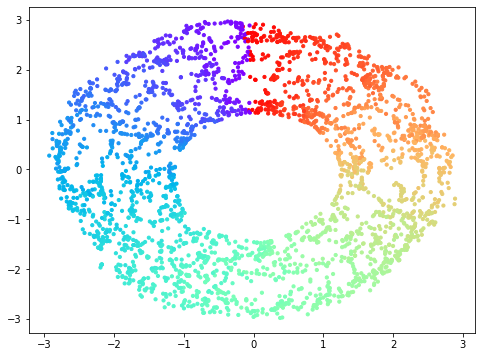} & \includegraphics[width=0.17\textwidth]{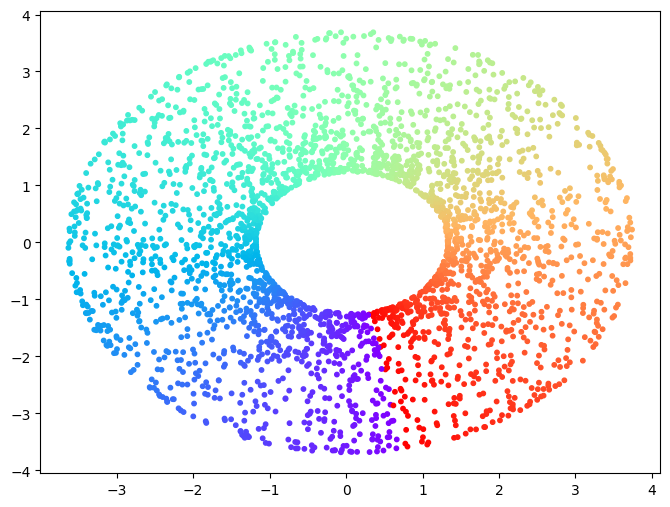} & \includegraphics[width=0.17\textwidth]{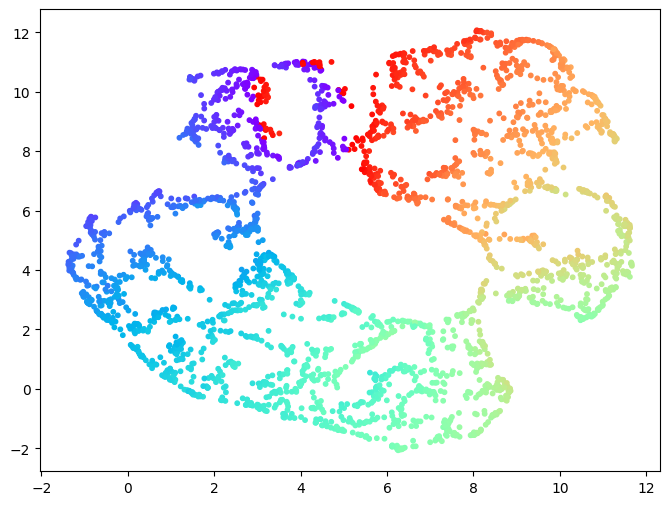} \\
        \textbf{(4) Mammoth} & 
        \includegraphics[width=0.17\textwidth]{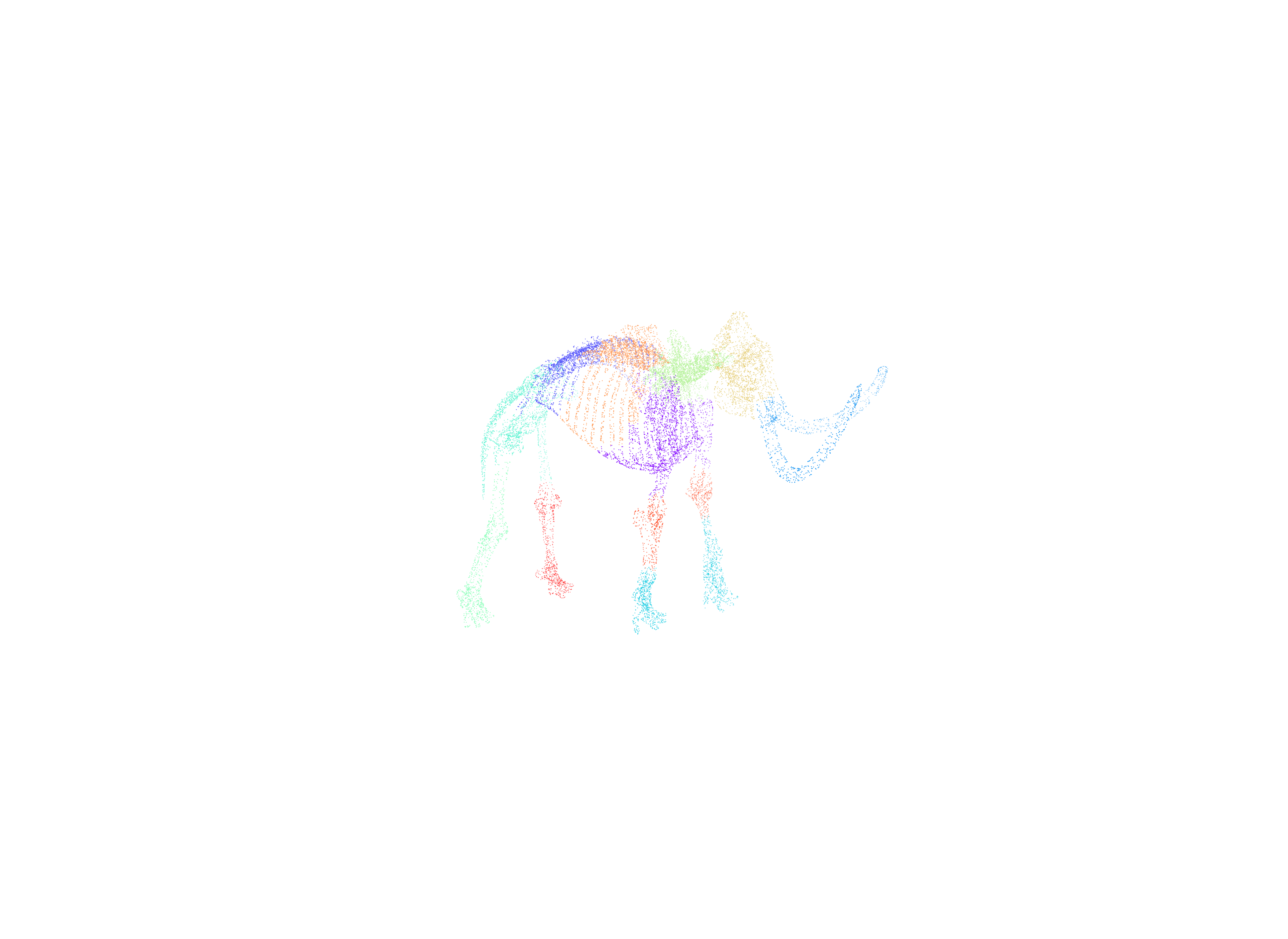}
         & 
        \includegraphics[width=0.17\textwidth]{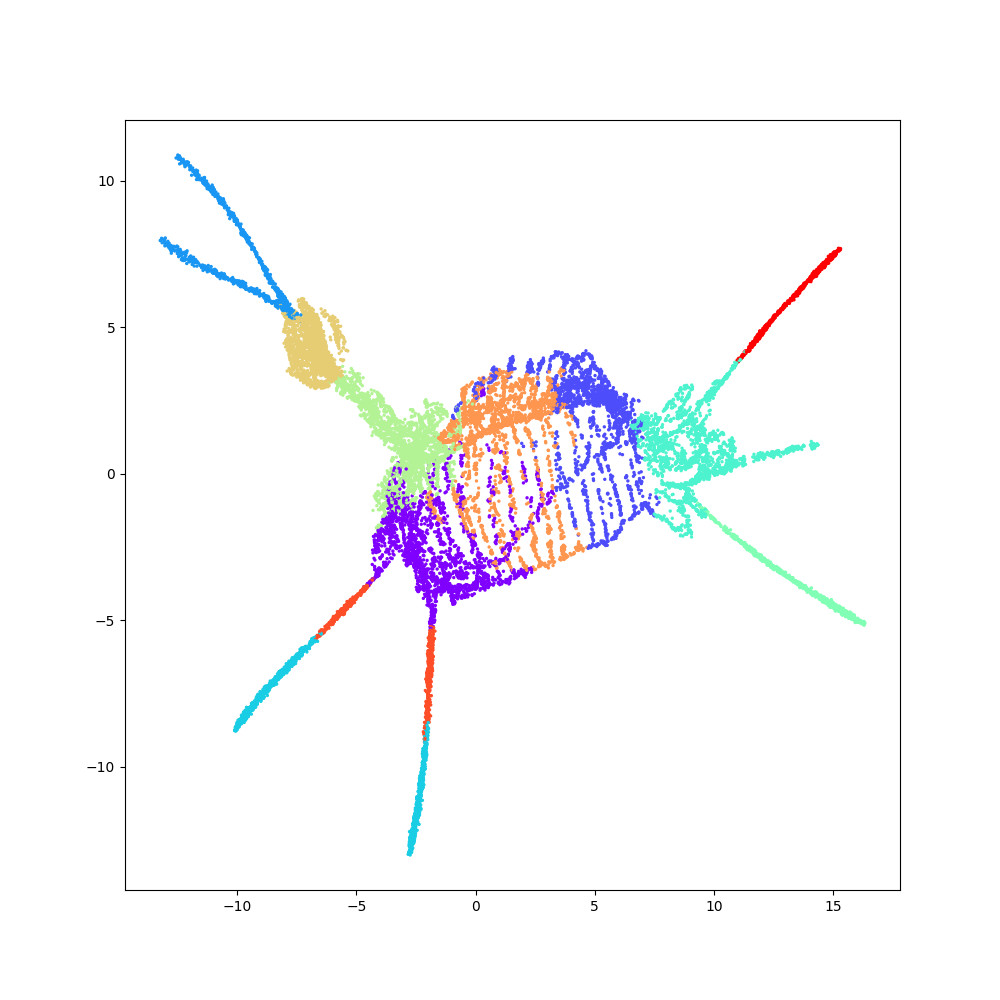}
          & \includegraphics[width=0.17\textwidth]{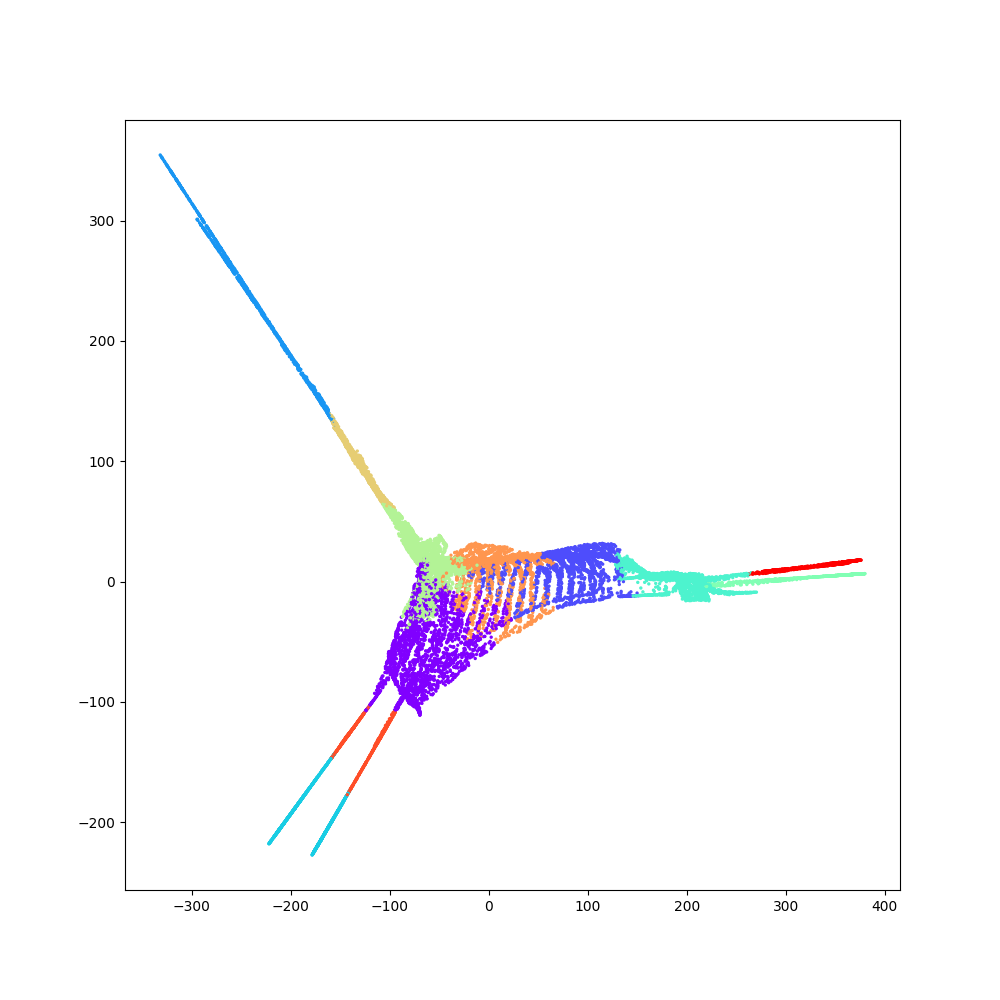} 
          & \includegraphics[width=0.17\textwidth]{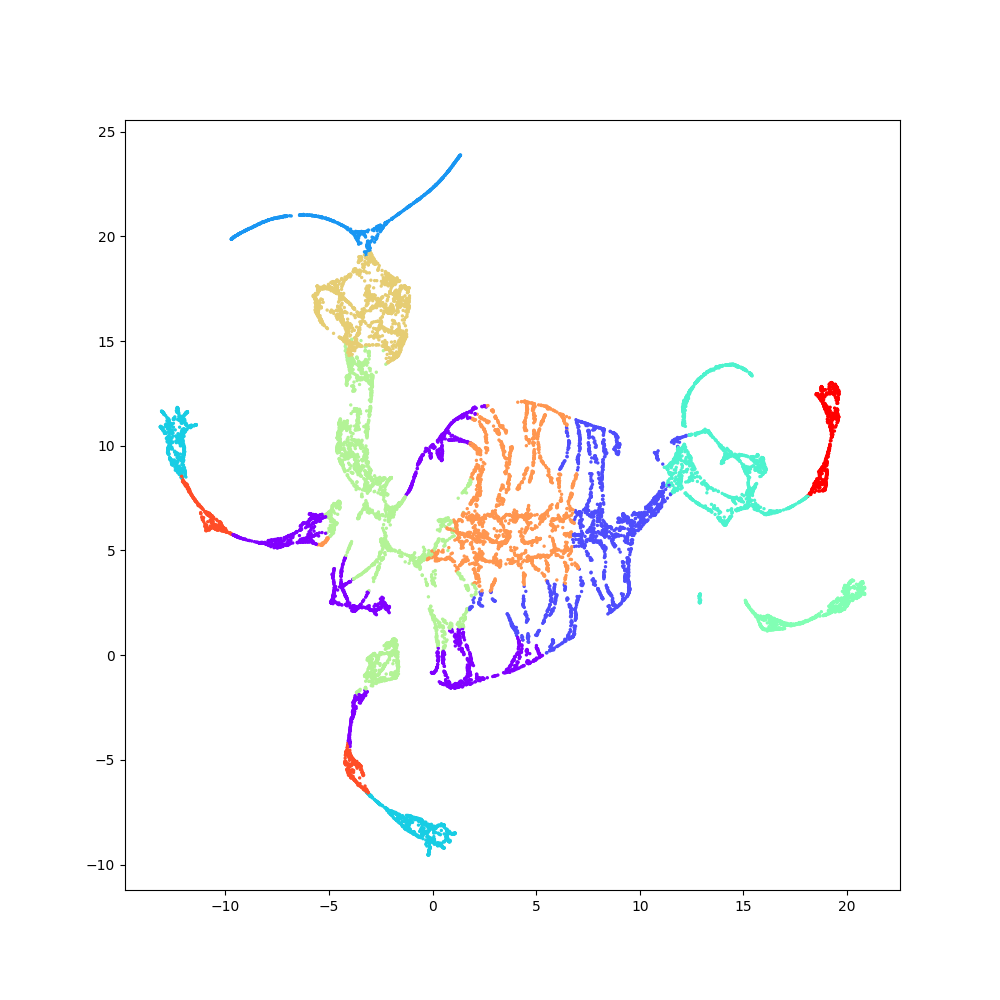} 
    \end{tabular}
    \caption{IsUMap, Isomap and UMAP.}
    \label{tab:IsUMap}
  \end{table}
  
\subsection{Benchmark data sets and clustering}
\begin{figure}[H]
    \centering
    \graphicspath{{image/MNIST/}}
    \subfigure[]{\includegraphics[width=0.30\textwidth]{MNIST10000.IsUMapk20.png}}
    \subfigure[]{\includegraphics[width=0.30\textwidth]{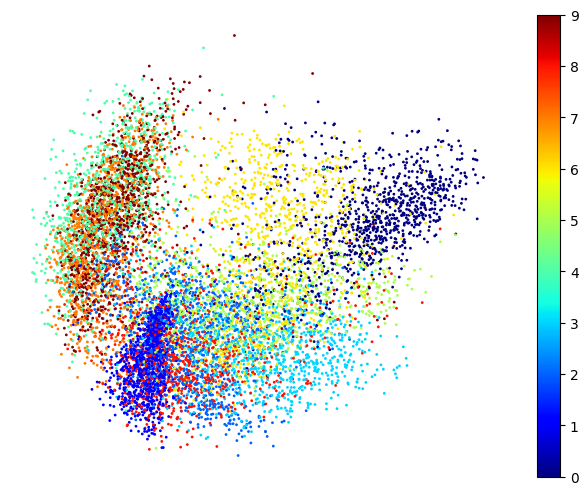}}
    \subfigure[]{\includegraphics[width=0.30\textwidth]{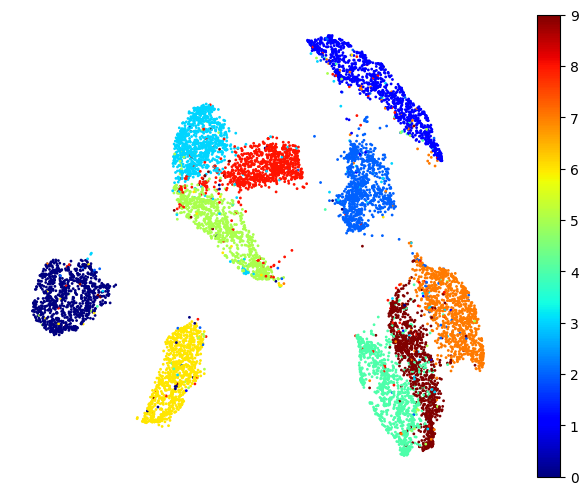}}
    
    \graphicspath{{image/BreastCancer/}}
    \subfigure[]{\includegraphics[width=0.30\textwidth]{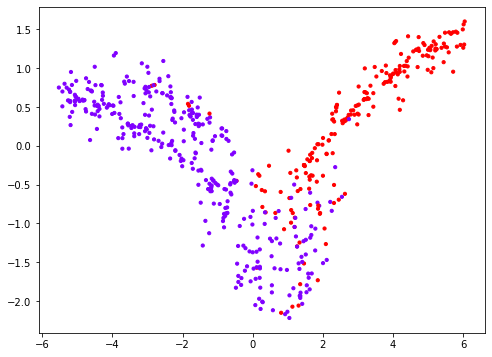}}
    \subfigure[]{\includegraphics[width=0.30\textwidth]{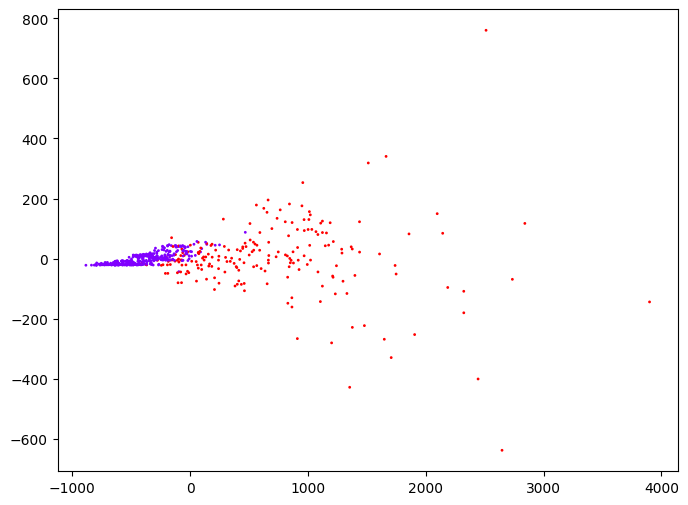}}
    \subfigure[]{\includegraphics[width=0.30\textwidth]{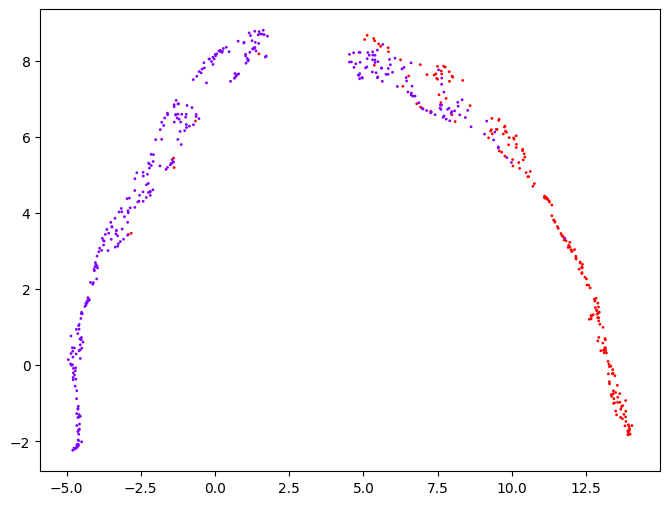}}
\caption{MNIST dataset (data size$=10000$, dim$=28\times 28$,  $k=20$) (a) IsUMap,  (b) Isomap, (c) UMAP. \\
Wisconsin breast cancer datasets (data size $=570$, dim$=32$, $k=20$) (d) IsUMap,  (e) Isomap, (f) UMAP.}
    \label{fig:2.8}
\end{figure}

\begin{figure}[H]
    \centering
    \graphicspath{{image/PSI/}}
     \subfigure[]{\includegraphics[width=0.30\textwidth]{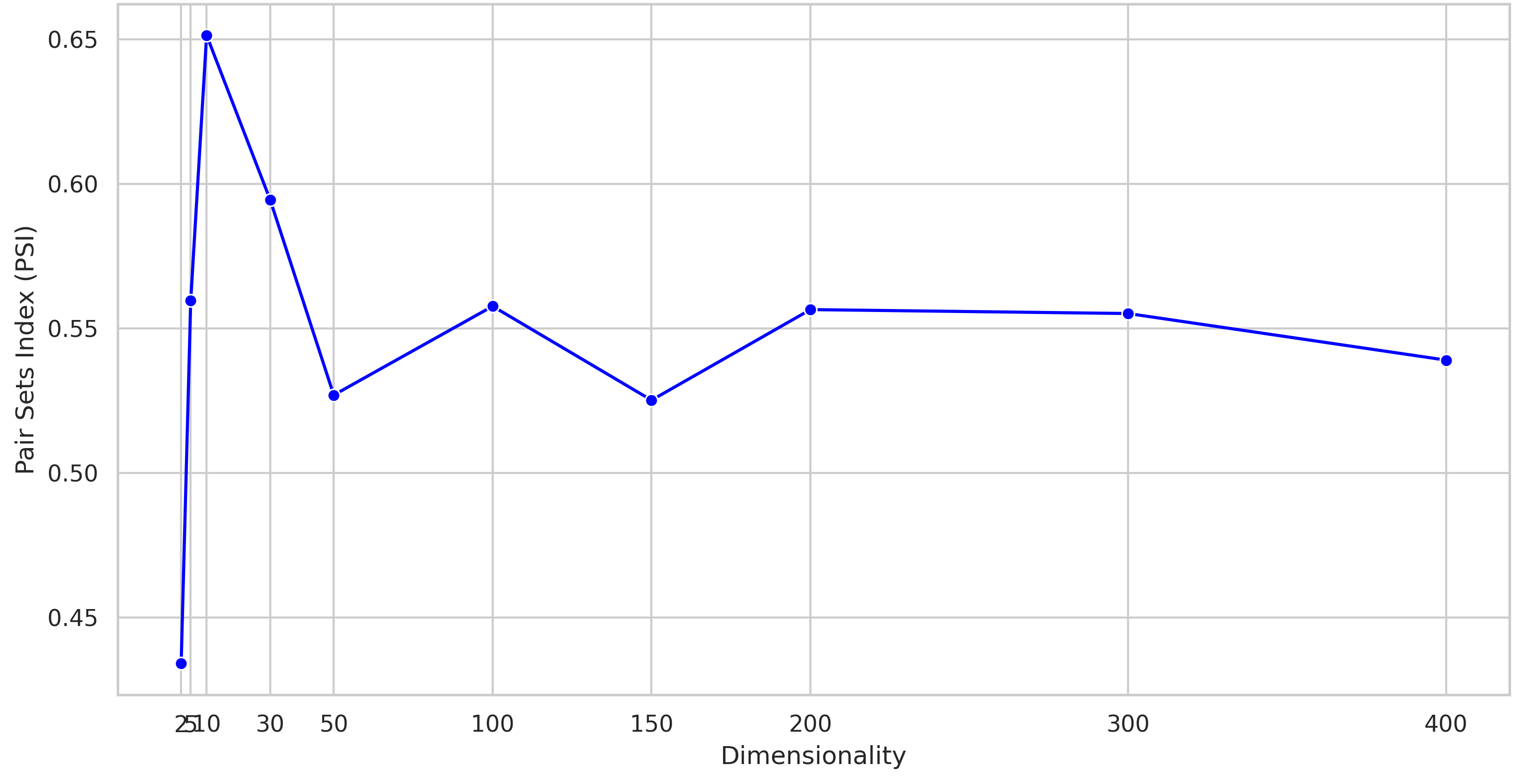}}
    \subfigure[]{\includegraphics[width=0.30\textwidth]{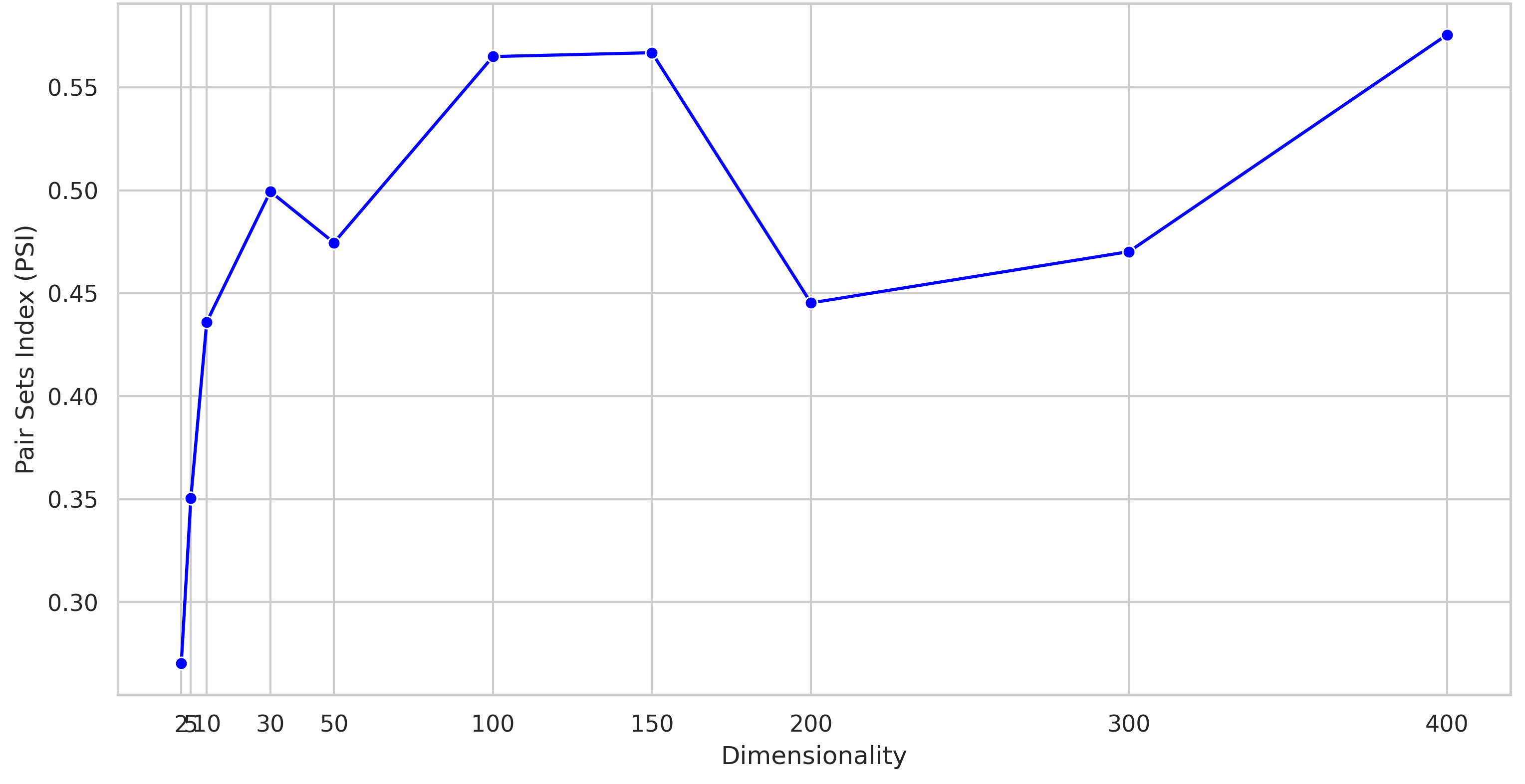}}
    \subfigure[]{\includegraphics[width=0.30\textwidth]{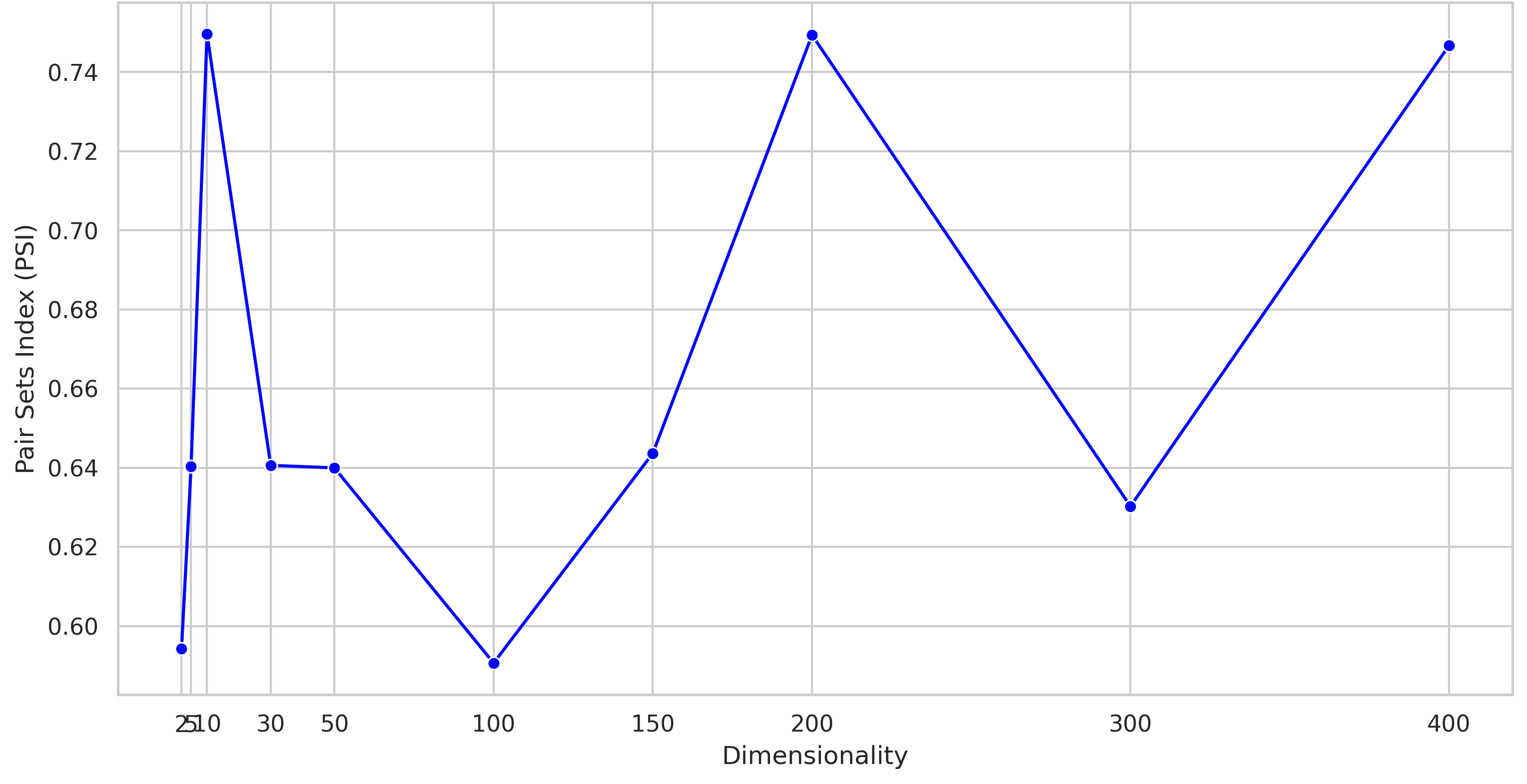}}
    \caption{Clustering performance in MNIST dataset after dimensionality reduction by (a)IsUMap, (b)Isomap and (c) UMAP.}
     \label{fig:2.9}
    \end{figure}

We next apply IsUMap to  some high dimensional real-world datasets which
  are often used as benchmarks in the literature,  MNIST and the Wisconsin breast cancer datasets. The outcomes are presented in Figure \cref{fig:2.8}.\\
In Image (a), we see the result of IsUMap on the MNIST dataset. IsUMap shows clusters more distinctly than Isomap, which tends to mix many clusters. However, IsUMap is not as effective as UMAP at separating different clusters. To systematically explore this finding rather than relying on visual observation,  we applied the Pair Sets Index (PSI) introduced in \cite{Rezaei16} to evaluate the effect of these three dimensionality reduction methods  (Isomap, UMAP and IsUMap) in clustering of the MNIST dataset. PSI is a cluster validation metric based on pair-set matching and adjusted for chance, ranging from $0$ to $1$, where $0$ indicates random partitioning and $1$ represents perfect labeling with respect to ground truth annotations.\\
For all cases, we used  $k-$means as the clustering method with $k=10$, and PSI assessed how the resulting clusters aligned with the initial labeling of the data (e.g., labeling by 10 digits). \\
The graph of the PSI evaluated across multiple dimensions and linearly interpolated in between is presented in Figure \ref{fig:2.9}. This analysis aims to determine whether clustering performance improves as the dimensionality reduces and to identify the point at which further dimension reduction negatively impacts the task. Among these methods, UMAP demonstrates the best clustering performance, followed by IsUMap, with Isomap showing the worst performance. However, all three methods exhibit their optimal clustering performance at dimension 10, after which there is a sharp decline.\\ 
It seems that the clustering capabilities of UMAP mainly come from the way that the embedding is performed, in particular from the negative undersampling (for more details on this, cf. \cite{Damrich2021}). We therefore conjecture that choosing a different embedding method than pure MDS could give rise to better clustering capabilities of IsUMap as well. In particular, one could combine MDS with clustering functors, as described in \cite{shiebler2020} and \cite{shiebler2020clustering}.
 In image (d) of Figure \ref{fig:2.8}, we observe the performance of IsUMap on the Wisconsin breast cancer dataset. Unlike Isomap (Image (e)), IsUMap aims to differentiate between two clusters: benign and malignant. It effectively highlights these two clusters and tends to separate them well. Moreover, since IsUMap aims to preserve the geodesic distances between data points based on the local distances \ref{localmet}, its underlying shape may provide a clearer representation of the structure of the dataset than UMAP (image (f)).\\
 While we do not have the original shapes of these datasets for direct comparison, the efficiency of methods can be assessed based on the objectives or tasks for which these representations are intended. This allows for selecting the most suitable method as well as representation dimension for the given purpose. \\

 \subsection{Non-uniformly distributed data}
\begin{figure}[H]
    \graphicspath{{image/Hemisphere/}}
    \centering
    \subfigure[]{\includegraphics[width=0.24\textwidth]{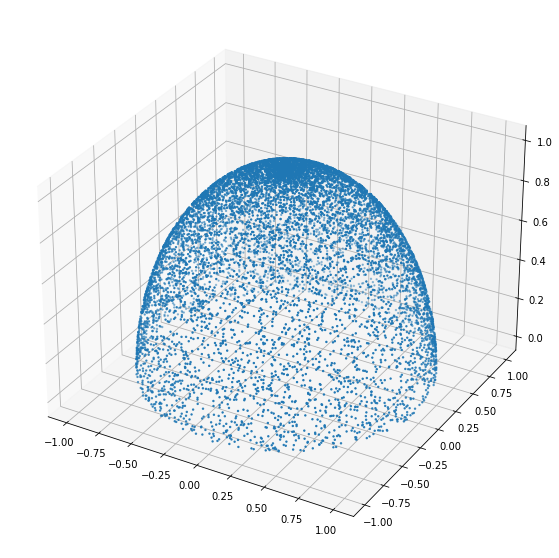}}
     \subfigure[]{\includegraphics[width=0.24\textwidth]{reduced_data_isumapk30_canonical.png}}
     \subfigure[]{\includegraphics[width=0.24\textwidth]{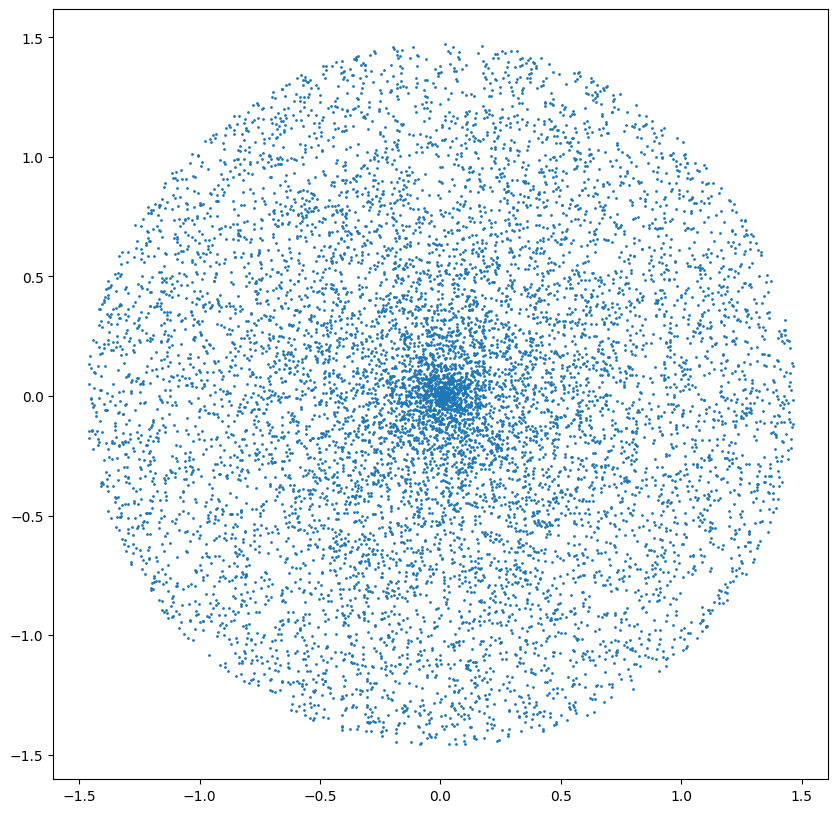}}
    \subfigure[]{\includegraphics[width=0.24\textwidth]{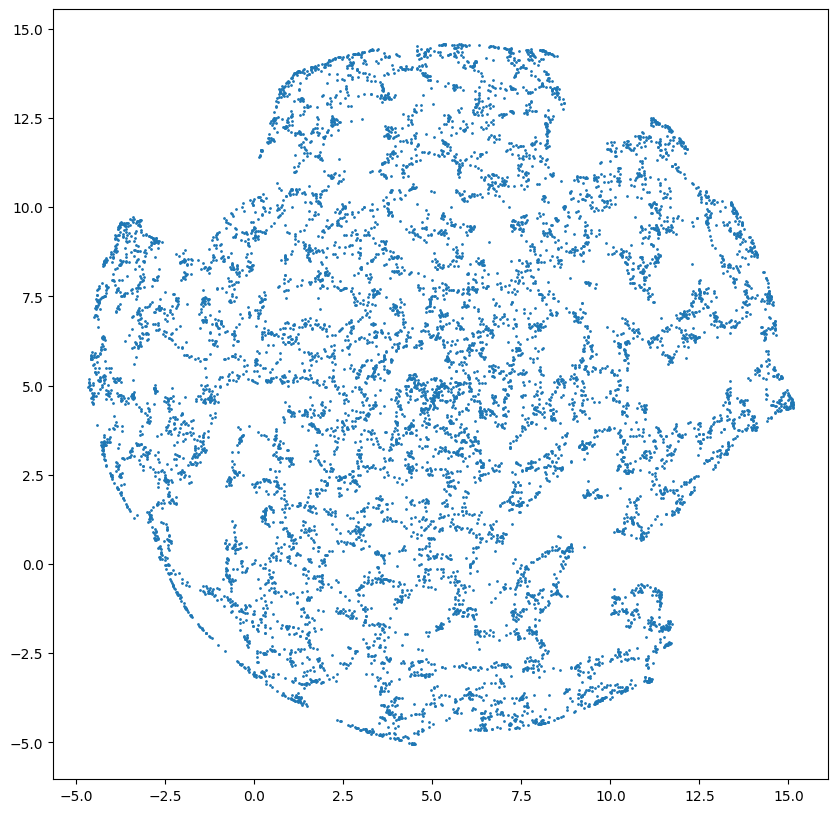}}  
    \caption{visualization of a sample of size $10000$ generated on Hemisphere with non-uniform distribution in dimension $2$ with $k=30$, (a) Data set, (b) IsUMap,  (c)Isomap, (d)UMAP.}
    \label{fig:3.8}
\end{figure}

\begin{figure}[H]
    \graphicspath{{image/Hemisphere/UMAP_IsUMap_diff_Tnorm}}
    \centering
    \subfigure[]{\includegraphics[width=0.24\textwidth]{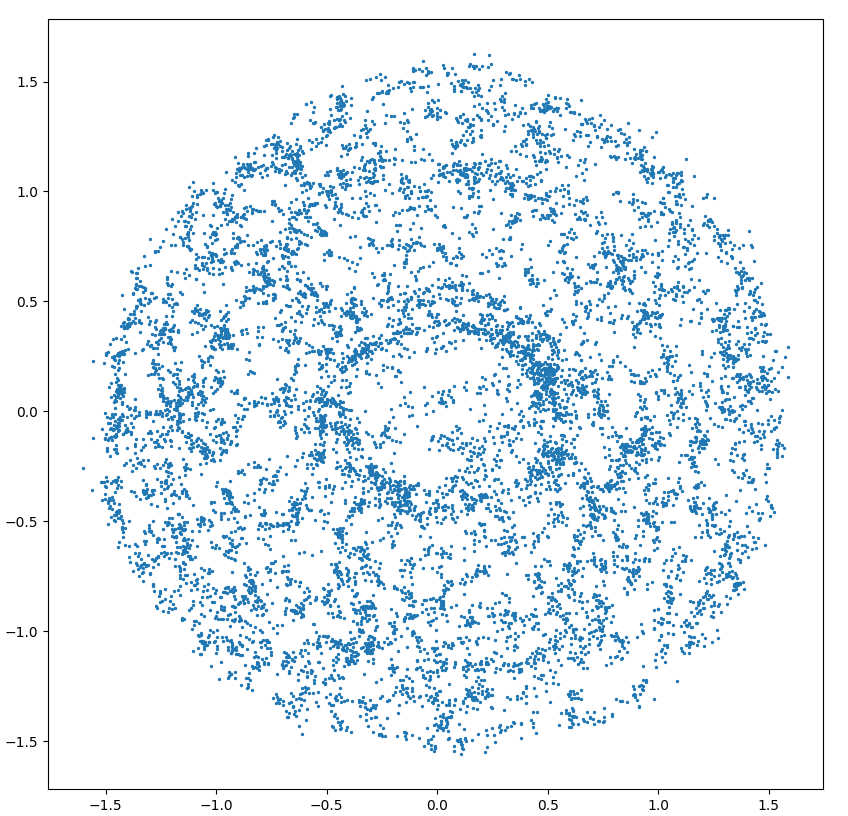}}
     \subfigure[]{\includegraphics[width=0.24\textwidth]{reduced_data_isumapk30_canonical.png}}
     \subfigure[]{\includegraphics[width=0.24\textwidth]{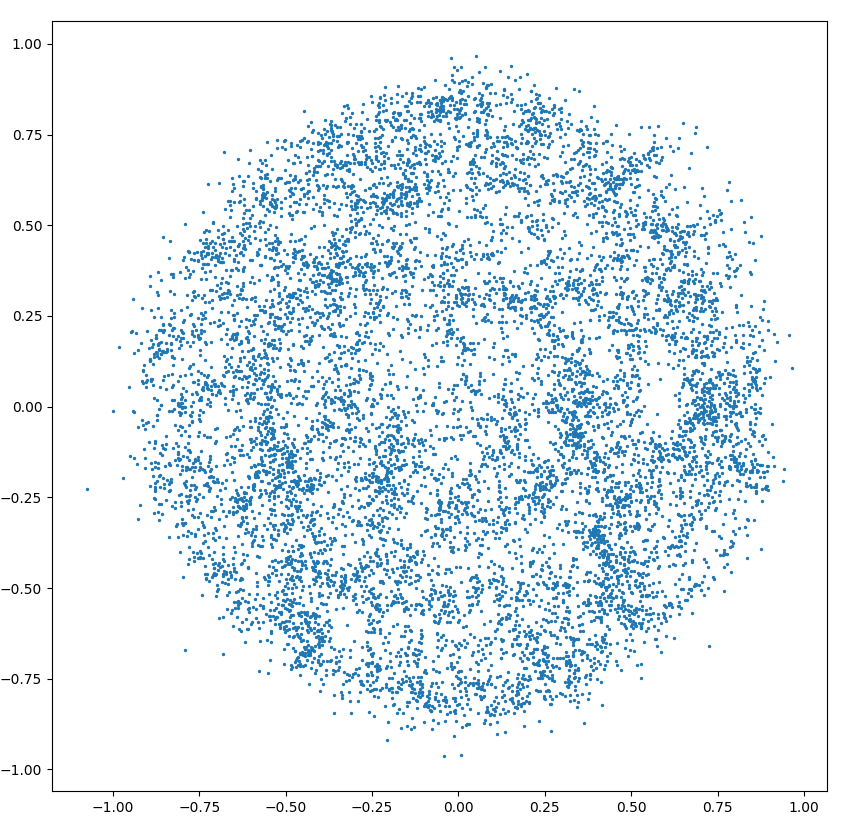}}
      \subfigure[]{\includegraphics[width=0.24\textwidth]{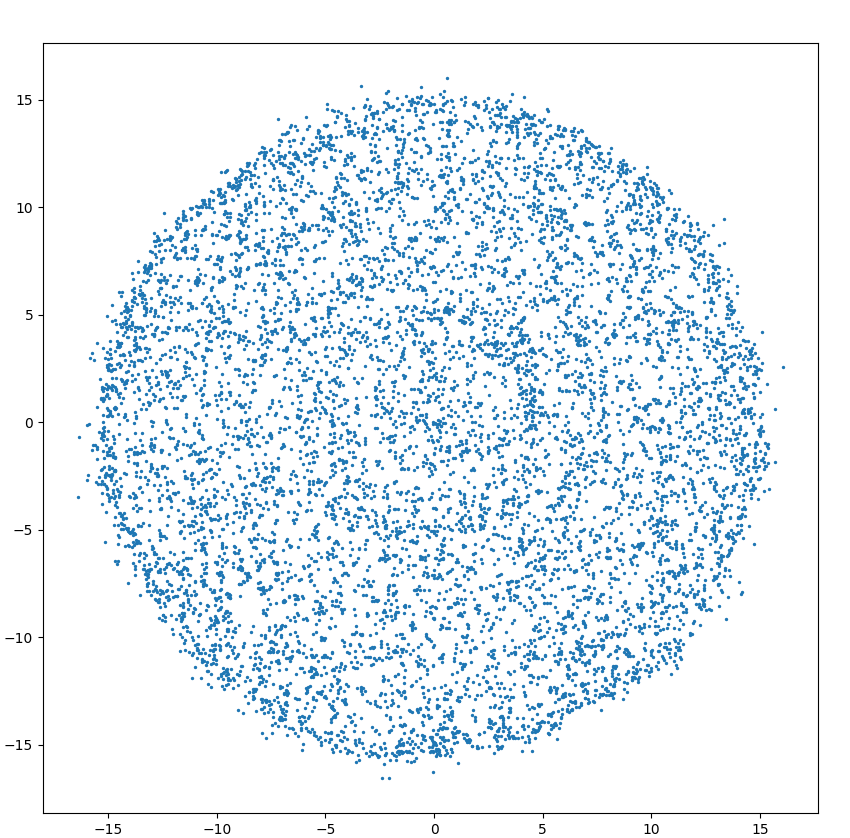}}

    \subfigure[]{\includegraphics[width=0.24\textwidth]{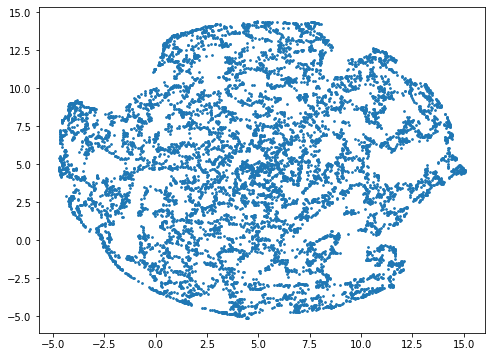}}
     \subfigure[]{\includegraphics[width=0.24\textwidth]{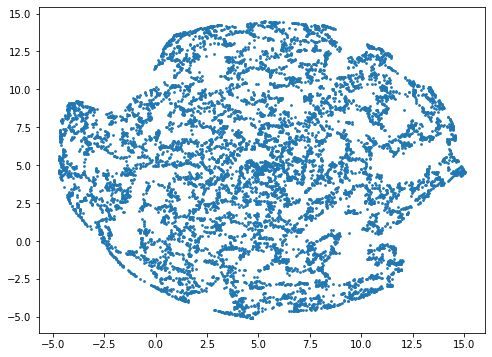}}
      \subfigure[]{\includegraphics[width=0.24\textwidth]{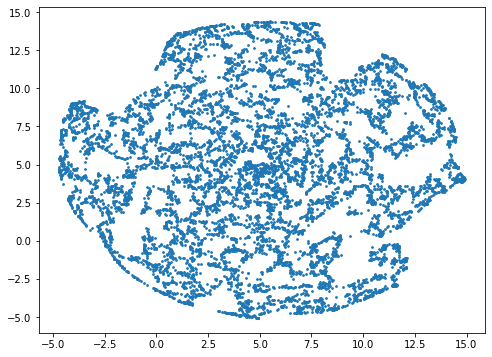}}
      \subfigure[]{\includegraphics[width=0.24\textwidth]{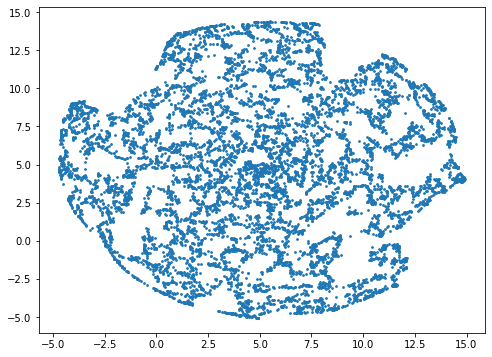}}
    \caption{visualization of a sample of size $10000$ generated on Hemisphere with non-uniform distribution in dimension $2$ by IsUMap (first row) and UMAP (second row)  with $k=30$ various t-conorms: Algebraic sum (a,e), Canonical (b,f), Bounded sum (c,g), Drastic sum (d,h).}
    \label{fig:4.8}
\end{figure}

\begin{figure}[H]
    \graphicspath{{image/Hemisphere/UMAP_IsUMap_diff_Tnorm}}
    \centering
    \subfigure[]{\includegraphics[width=0.24\textwidth]{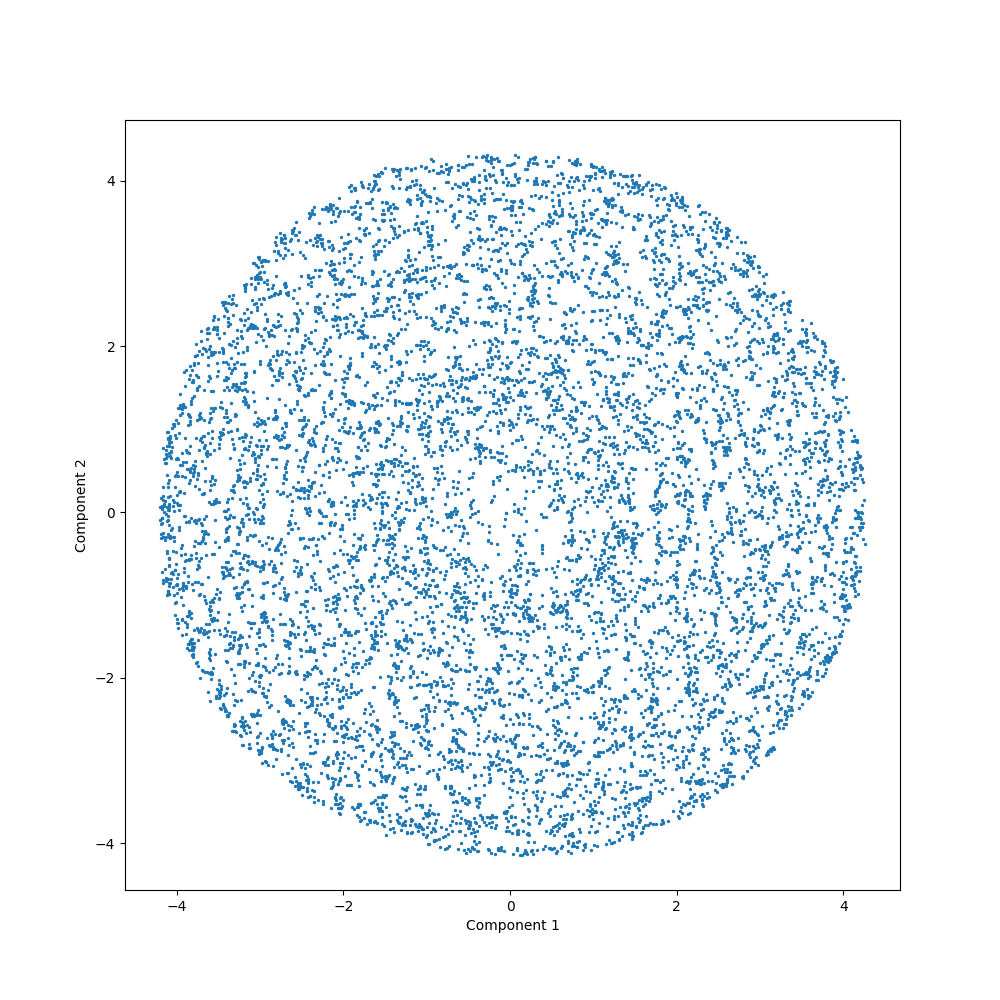}}
   \subfigure[]{\includegraphics[width=0.24\textwidth]{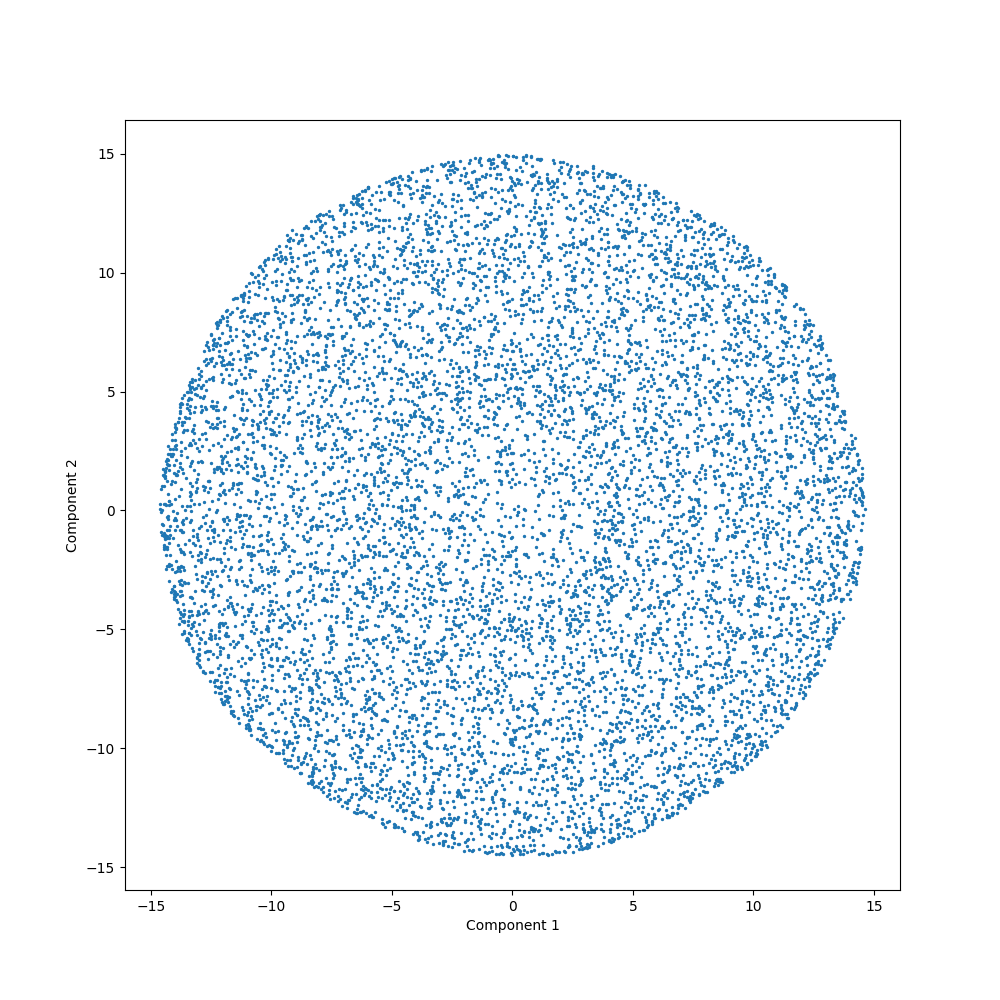}} 
   \subfigure[]{\includegraphics[width=0.24\textwidth]{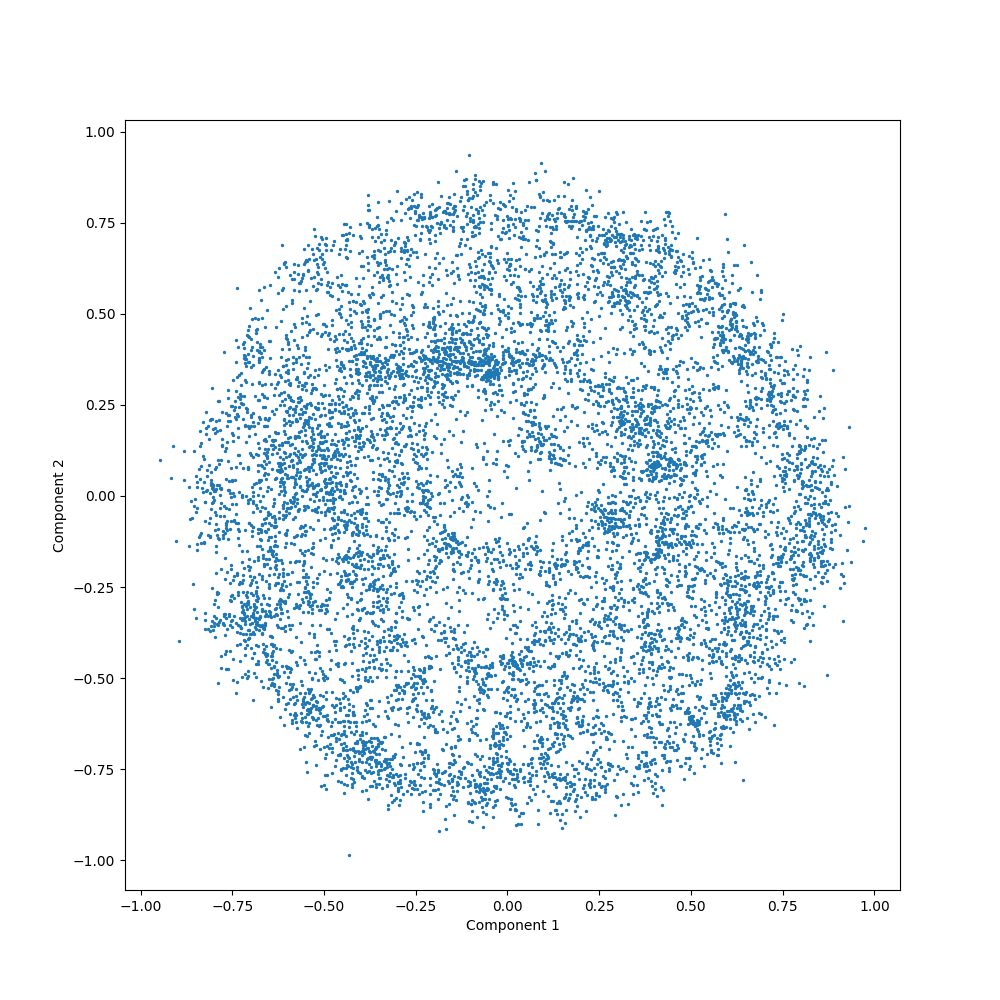}}
   \subfigure[]{\includegraphics[width=0.24\textwidth]{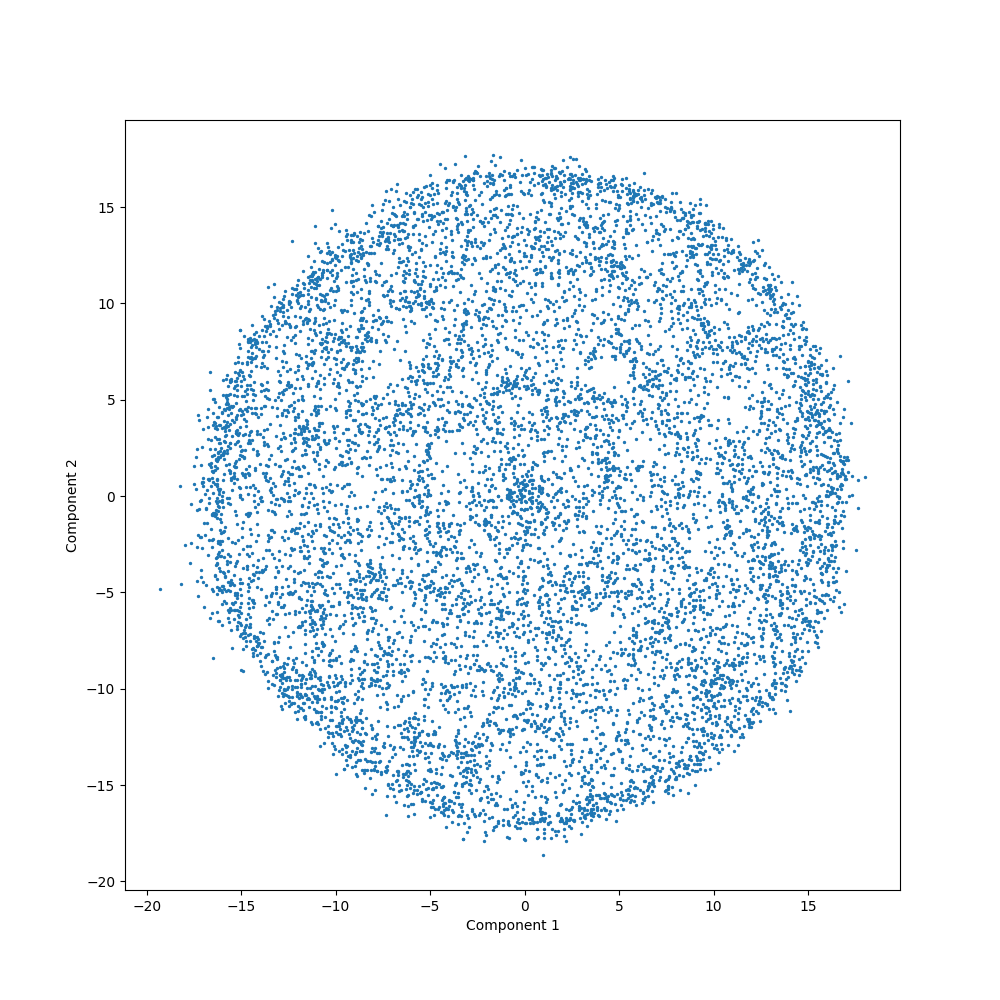}}
    \caption{visualization of a sample of size $10000$ generated on Hemisphere with non-uniform distribution in dimension $2$ by IsUMap without subtracting $\rho_i$ (with $k=30$) with various t-conorms: Algebraic sum (a), Canonical (b), Bounded sum (c), Drastic sum (d).}
    \label{fig:4.8.1}
\end{figure}

 We shall now analyze how IsUMap can deal with important challenges present in the
 data, i.e.\ non-uniform distributions. As we described in Subsection \ref{sec:concreteAlgo}, the normalization factor $\sigma_i$ used to locally distort the initial metric $d$ aims to make distances between nearest neighbors more uniform. It  therefore addresses the issue of non-uniform distributions, where the density of data points varies in the space. 
 To verify this, we construct an example of such a distribution on a
 simple geometric  object, namely a hemisphere, and see how IsUMap performs on
 this example.  We compare the resulting embedding with that of UMAP, which
 also uses similar local distortion of the metric, but uses a different
 approach for combining the local metrics and for the final embedding, and with the Isomap method. \\
Dealing with non-uniform data distributions is an active area of research and several strategies have been developed to address this problem. For instance, methods based on stochastic neighborhood embedding (t-SNE, cf. \cite{vandermaaten08a}), where the embedding is designed in a way that it approximates the weight distribution on the edge-set of the neighborhood graph, introduce weights to account for the density variations in different regions of the data manifold by assigning more importance to the pairwise correlations in denser regions. However, understanding the impact of different weight assignments on the geometry of the final representation can be less straightforward. We therefore examine this effect for our small toy example. \\
Thus, we apply IsUMap to the generated dataset (image (a) in \cref{fig:3.8}) on the upper hemisphere. To introduce a non-uniform density, one can vary the density across different
regions of the hemisphere. Here, we generate  a denser region around the pole
and sparser regions towards the equator. Also, as the hemisphere has a
boundary, we expect to see boundary effects. \\
The comparison between the result of applying IsUMap with that of applying UMAP and Isomap to yield an embedding in dimension $2$ in \cref{fig:3.8} shows that IsUMap results in a projection that looks like a disk with uniformly distributed datapoints in the interior.\\   
Since some choices can be made within the scheme, it is natural to
investigate their effects.  In particular,  for the
uniformization step, we can take different approaches for defining the scaling factor, for combining local structures, we can choose different  t-conorms, and  for the  final embedding in
low-dimensional space,  we can employ  either the probabilistic or the
geometric method. We probe IsUMap and UMAP on the generated dataset (see image
(a) in \cref{fig:3.8}) on the upper hemisphere, varying  the t-conorm  in both methods. Corresponding visualizations are depicted in \cref{fig:4.8}.\\
The images in the second row (images (e)-(h)) show the implementation of UMAP using different t-conorms, namely: Algebraic sum (image (e)), Canonical t-conorm (image (f)), Bounded sum (image (g), and Drastic sum (image(h))). As  previously explained, the results are similar. However, IsUMap in images (a)-(d) produces  different results as it creates a spiral in the projection of the  hemisphere with Algebraic sum (image (e)) 
and bounded sum (image (c)), which also fail to project the boundary of the hemisphere.  In contrast, the Canonical (image (b)) and Drastic sum (image (d)) operations generate a more uniformly distributed projection.\\
The difference in outcomes between Canonical and Drastic sums might be attributed to the subtraction of $\rho_i$ (the distance to the nearest neighbor). Since this dataset's  dimension is reduced only by $1$, applying this subtraction may not be necessary. This is depicted in Figure \cref{fig:4.8.1}.\\ 
Comparing the results of UMAP and IsUMap on the dataset generated on the hemisphere, we notice that IsUMap offers a geometrically more robust visualization, whereas UMAP might be more efficient for clustering tasks. This advantage of UMAP over IsUMap for clustering becomes particularly evident in the visualization of the MNIST dataset, as depicted in \cref{fig:2.8}. However, IsUMap has an advantage over UMAP when it comes to visualization of lower-dimensional manifolds. \\
The parameter $k$, which controls the construction of the graph where each sample point is connected to its $k$ nearest neighbors, requires careful tuning to achieve a satisfactory embedding. As shown in Table \ref{tab:IsUMap_diffNN}, IsUMap is sensitive to the hyperparameter $k$ when the dataset has a geometric structure, such as a Swiss Roll. By increasing the number of nearest neighbors, IsUMap attempts to fold the Swiss Roll and Swiss Roll with a hole, thereby preserving the extrinsic geometry. This sensitivity does not affect the results for the Torus dataset; however, for the MNIST dataset, increasing the number of neighbors improves the clustering performance.

\begin{table}[H]
	\centering
	\tiny
	\begin{tabular}{>{\centering\arraybackslash}m{1cm}|*{6}{>{\centering\arraybackslash}m{1.5cm}}}
		& \textbf{Data set} & \textbf{k=10} & \textbf{k=20}  & \textbf{k=40} & \textbf{k=50} & \textbf{k=100} \\
		& (a) & (b) & (c) & (d)&(e)&(f)  \\
		\hline \\
		\textbf{(1) Swiss Roll} &
		 \includegraphics[width=0.12\textwidth]{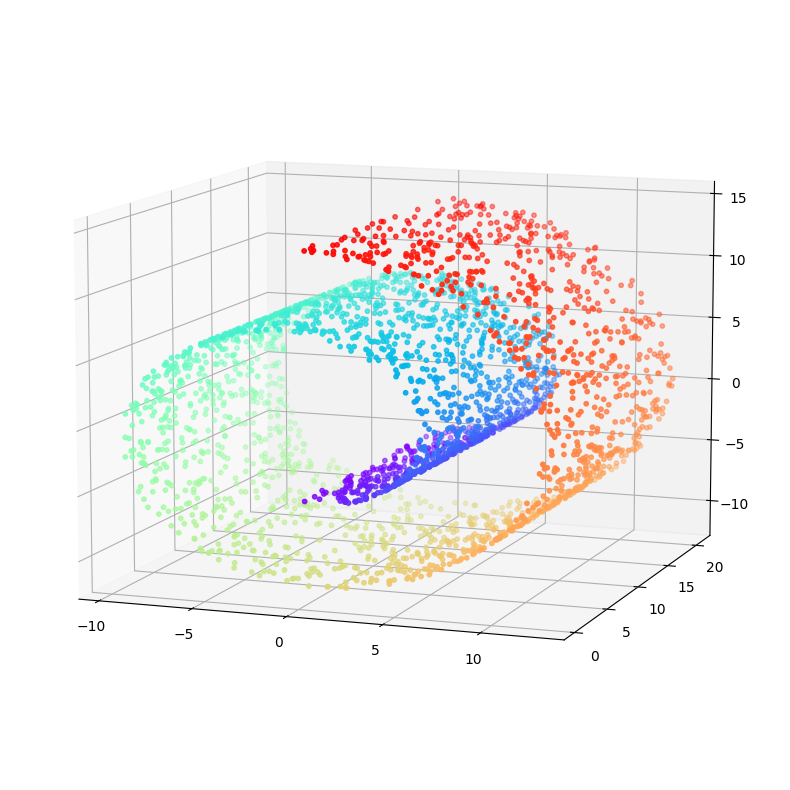} & \includegraphics[width=0.12\textwidth]{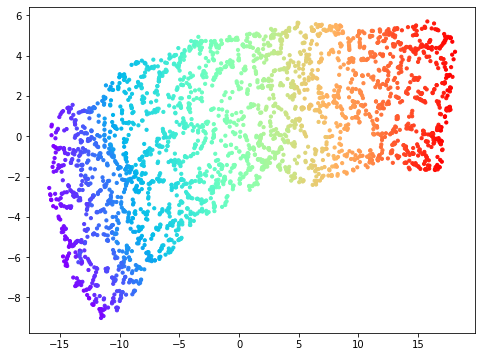} & \includegraphics[width=0.12\textwidth]{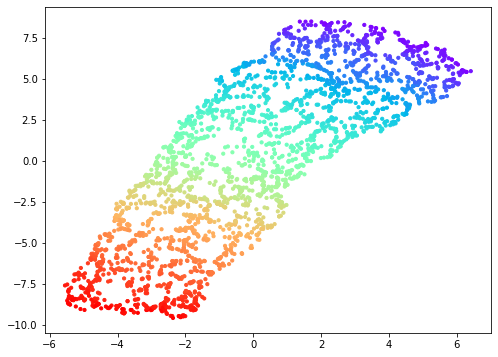} &
		\includegraphics[width=0.12\textwidth]{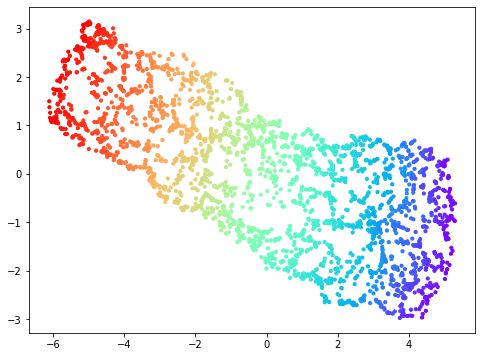} &
			\includegraphics[width=0.12\textwidth]{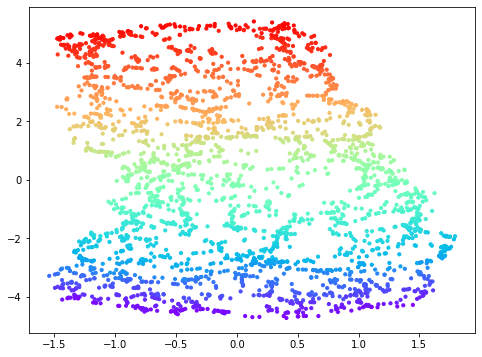} &
		 \includegraphics[width=0.12\textwidth]{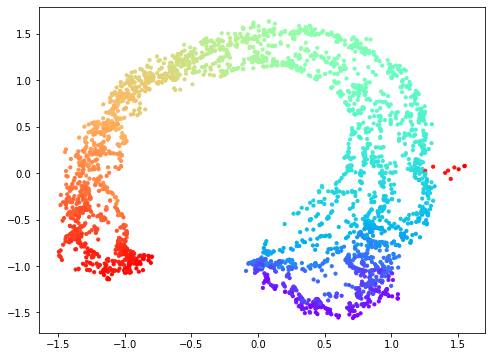} \\
		\textbf{(2) Swiss Roll with a hole} & 		 \includegraphics[width=0.12\textwidth]{image/IsUMAPdiffNN/swissRoll3000.png} & \includegraphics[width=0.12\textwidth]{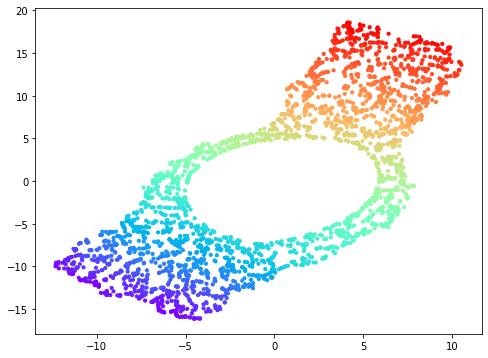} & \includegraphics[width=0.12\textwidth]{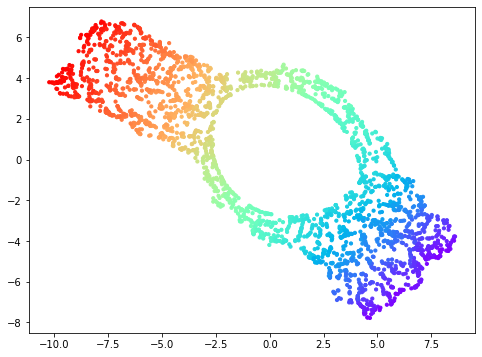} &
		\includegraphics[width=0.12\textwidth]{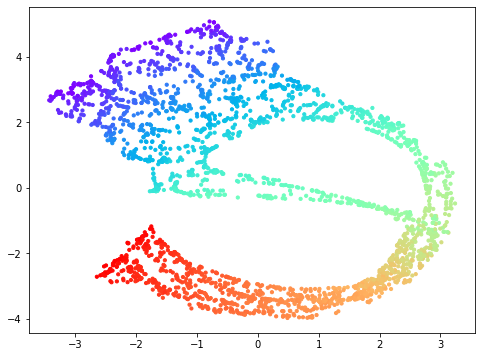} &
		\includegraphics[width=0.12\textwidth]{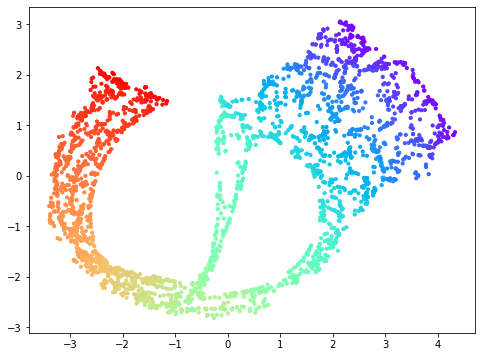} &
		\includegraphics[width=0.12\textwidth]{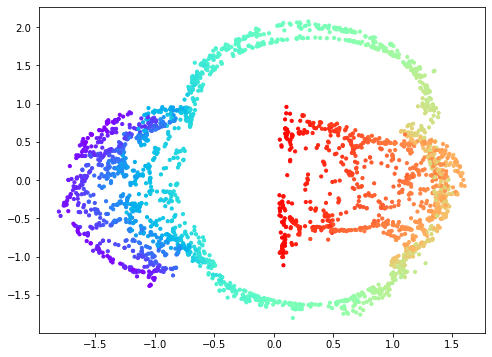} \\
		\textbf{(3) Torus} & 
		 \includegraphics[width=0.12\textwidth]{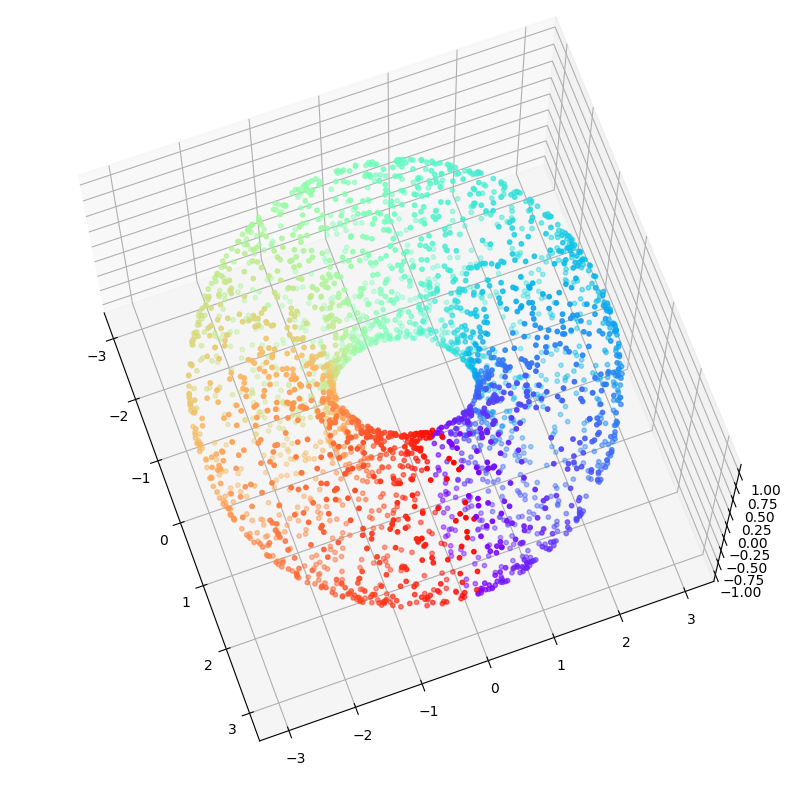} & \includegraphics[width=0.12\textwidth]{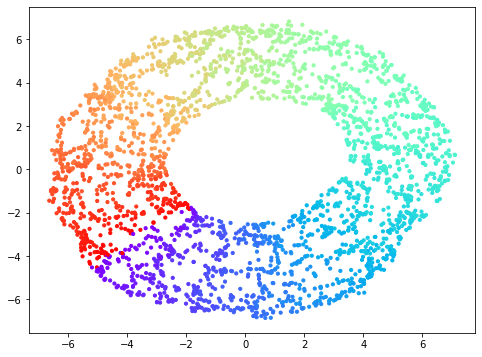} & \includegraphics[width=0.12\textwidth]{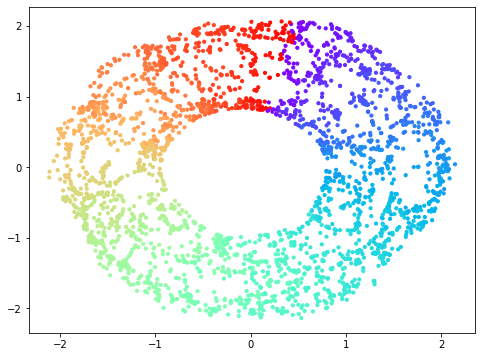} &
         \includegraphics[width=0.12\textwidth]{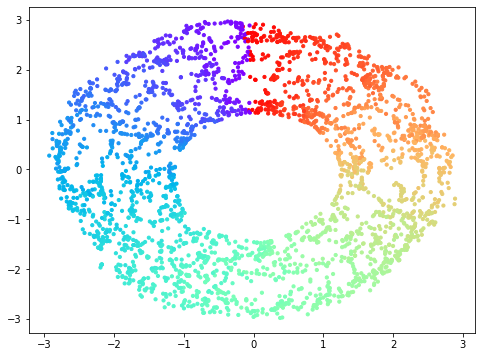} &
         \includegraphics[width=0.12\textwidth]{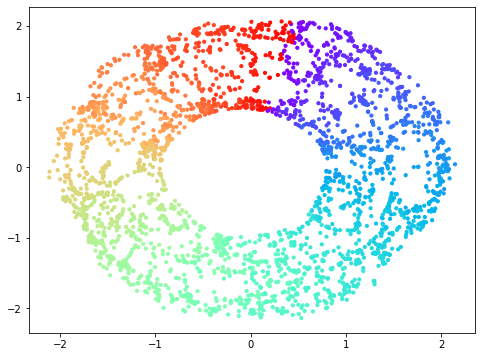} &
        \includegraphics[width=0.12\textwidth]{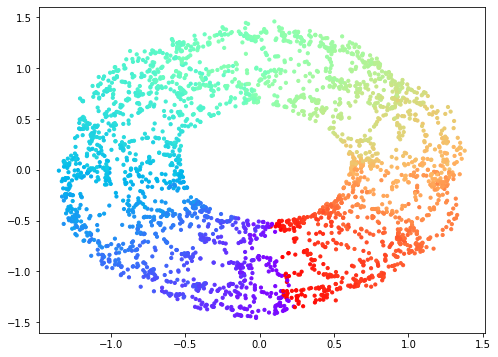} \\
		\textbf{(4) MNIST} & 
				 \includegraphics[width=0.12\textwidth]{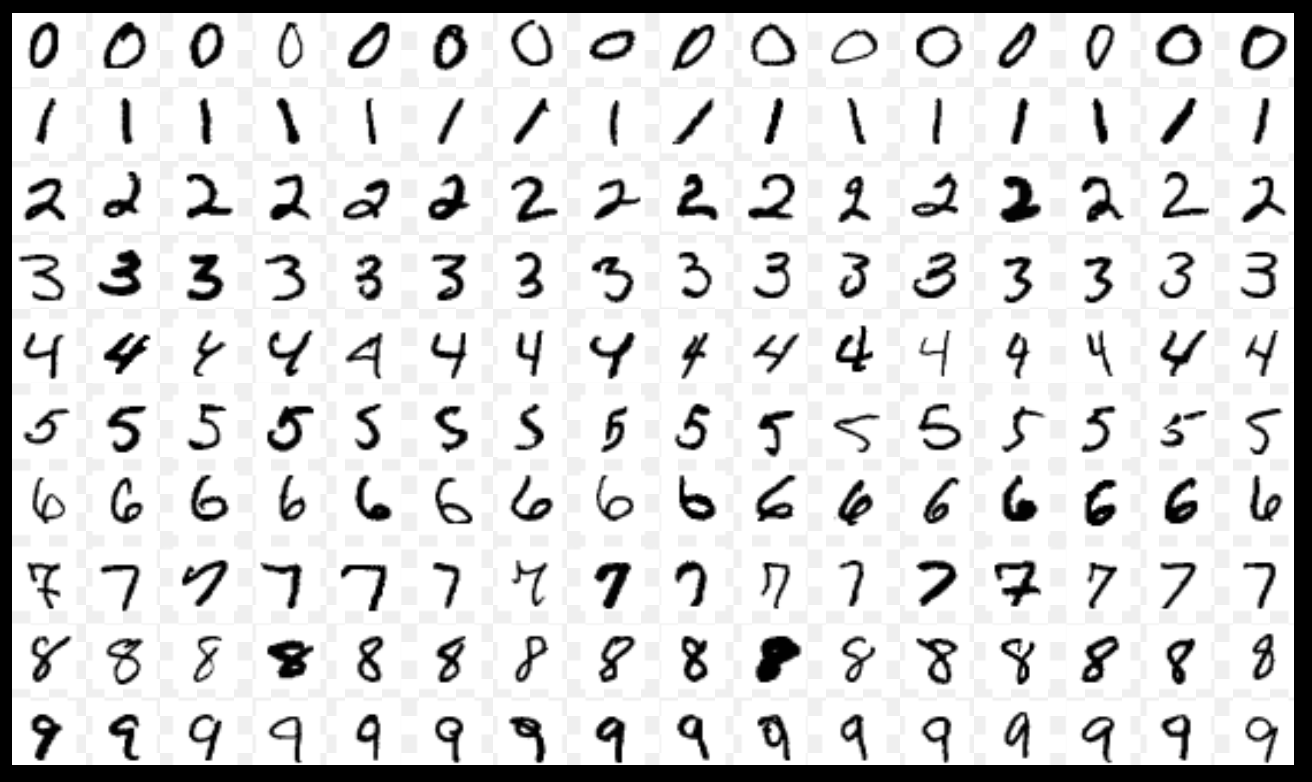} &
				  \includegraphics[width=0.12\textwidth]{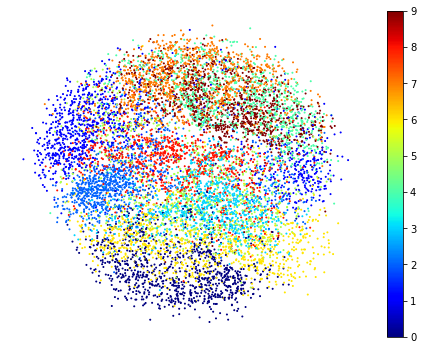} & \includegraphics[width=0.12\textwidth]{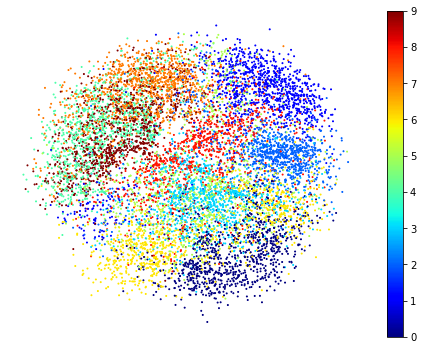} &
		\includegraphics[width=0.12\textwidth]{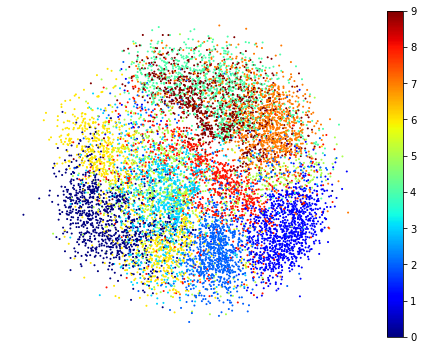} &
		\includegraphics[width=0.12\textwidth]{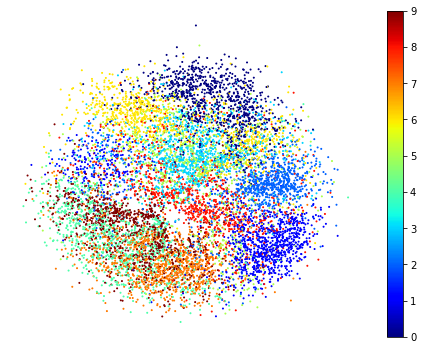} &
		\includegraphics[width=0.12\textwidth]{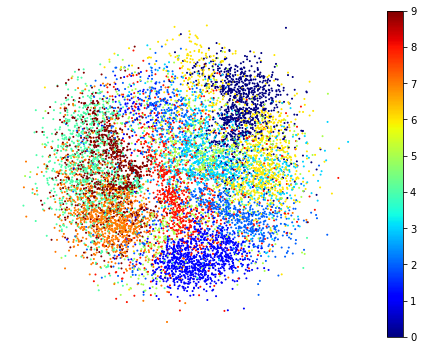} 
	\end{tabular}
	\caption{IsUMap with different $k$.}
	\label{tab:IsUMap_diffNN}
\end{table}
To consider how local distortion changes when altering the geometry of a shape, we demonstrate a Swiss Roll by further rolling up the 2D plane. Table \ref{tab:IsUMap_diffNN_SwissRoll}, shows the results of IsUMap with varying numbers of nearest neighbors for each column. We observe that for up to $30$ neighbors, IsUMap attempts to unroll the Swiss roll, preserving the intrinsic geometry. By increasing the number of nearest neighbors, it rolls the Swiss roll up again to preserve the extrinsic geometry. IsUMap performs well in preserving the clusters.\\
In contrast, Table\ref{tab:UMap_diffNN_SwissRoll}, shows the results of UMAP. With a lower number of neighbors, UMAP tries to unroll the Swiss roll but fails due to the discontinuous nature of the underlying structure, making it ineffective in representing the data in low dimensions. For a higher number of neighbors, UMAP tends to mix points together, thereby rolling the Swiss roll and failing to preserve clusters and underlying structures. Additionally, we observe a hole in the UMAP results, indicating a change in the topology of the dataset.

\begin{table}[H]
	\centering
	\tiny
	\begin{tabular}{>{\centering\arraybackslash}m{0.5cm}|*{7}{>{\centering\arraybackslash}m{1cm}}}
		& \textbf{Data set} & \textbf{k=10} & \textbf{k=20}  & \textbf{k=30}  & \textbf{k=50}& \textbf{k=100}&\textbf{k=200}\\
		& (a) & (b) & (c) & (d)&(e)&(f)&(g) \\
		\hline \\
		\textbf{(1)  $\cos(4t)$}&
		\includegraphics[width=0.10\textwidth]{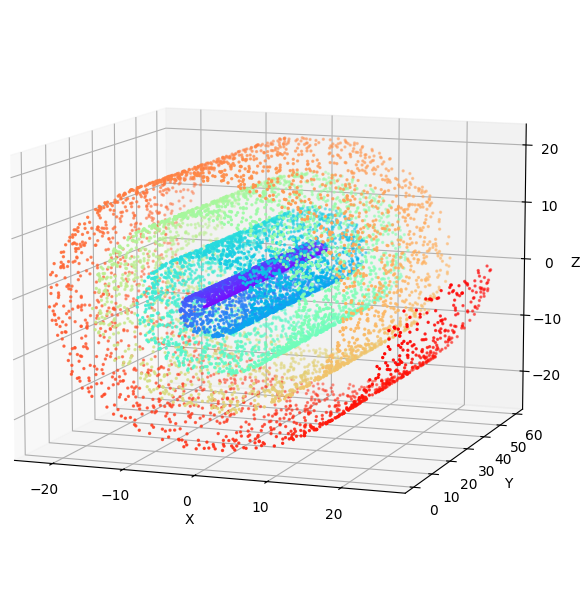} & \includegraphics[width=0.10\textwidth]{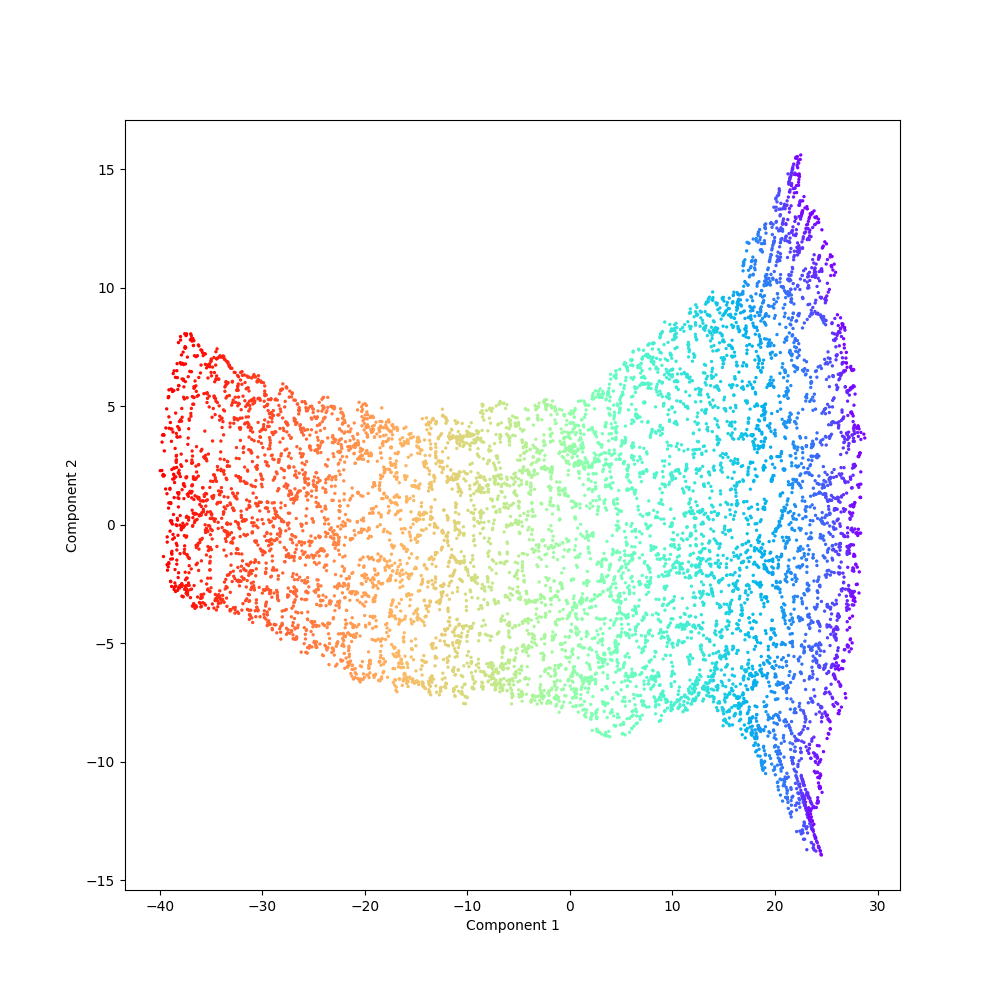} & \includegraphics[width=0.10\textwidth]{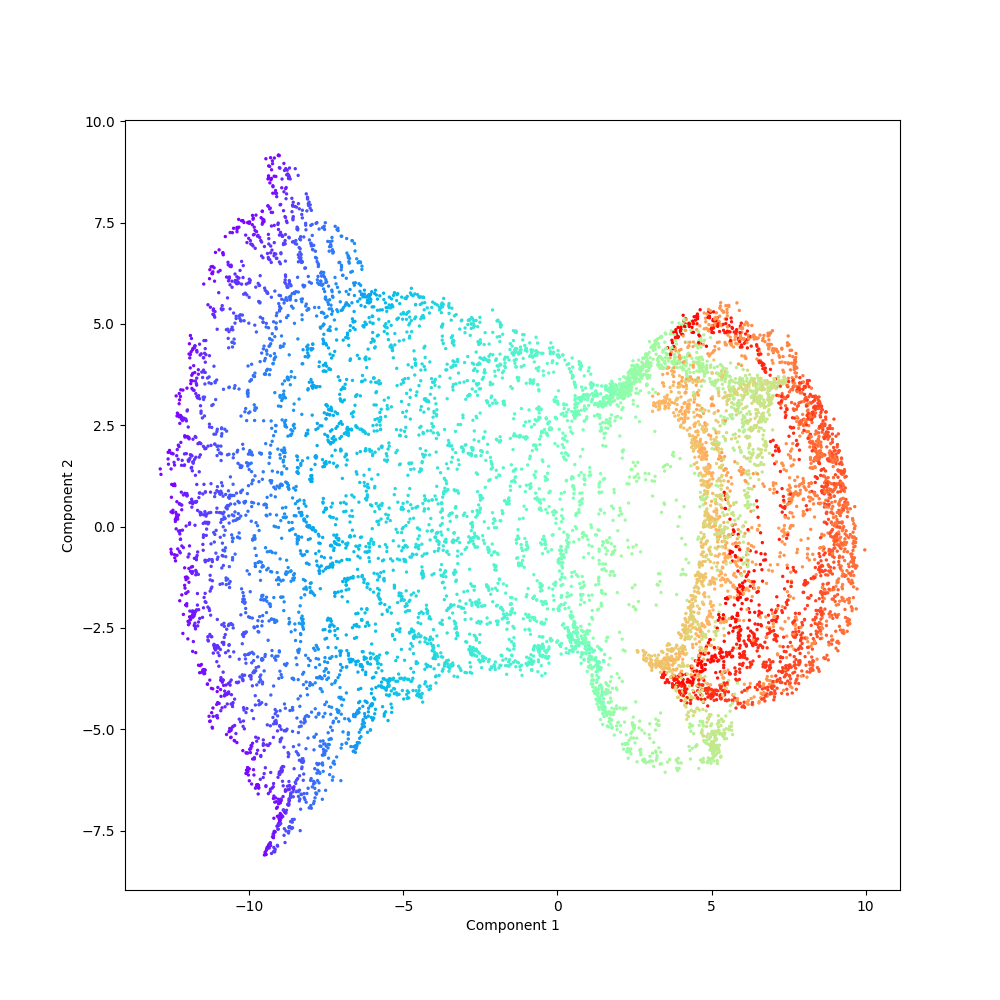} &
		\includegraphics[width=0.10\textwidth]{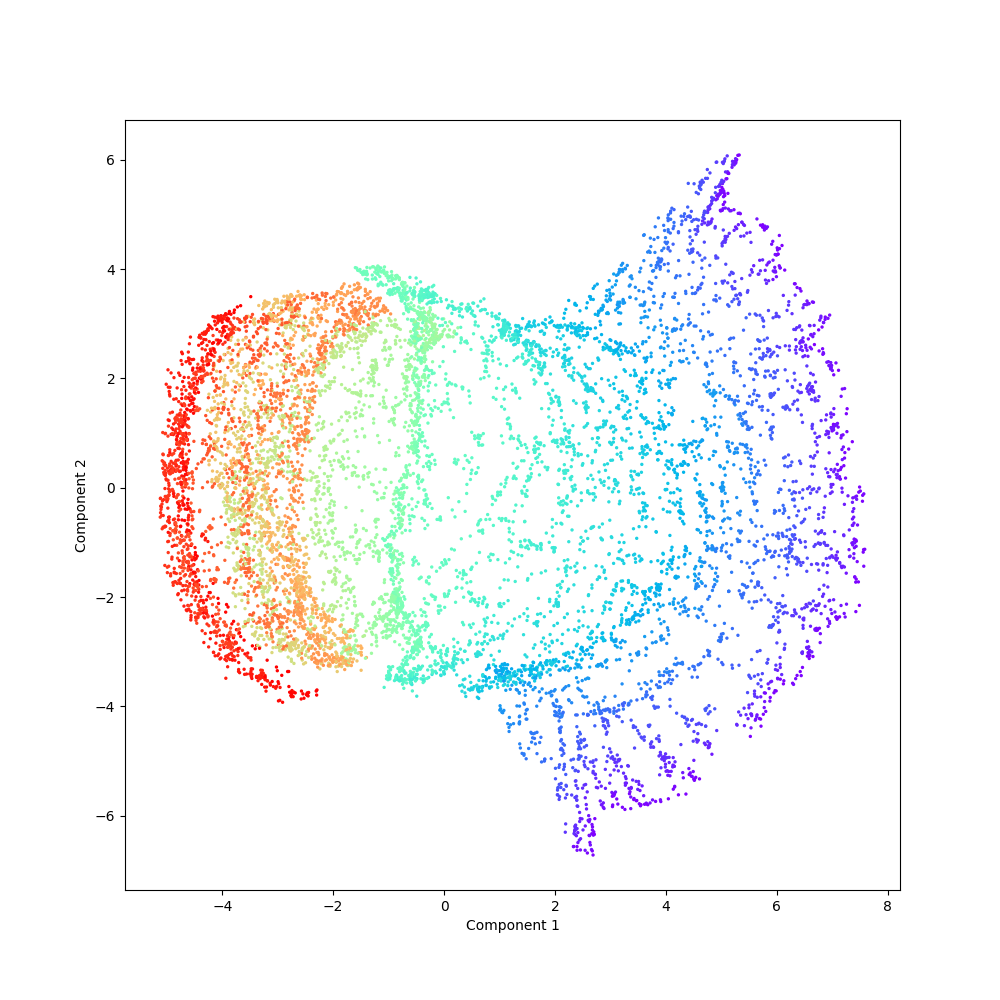} &
		\includegraphics[width=0.10\textwidth]{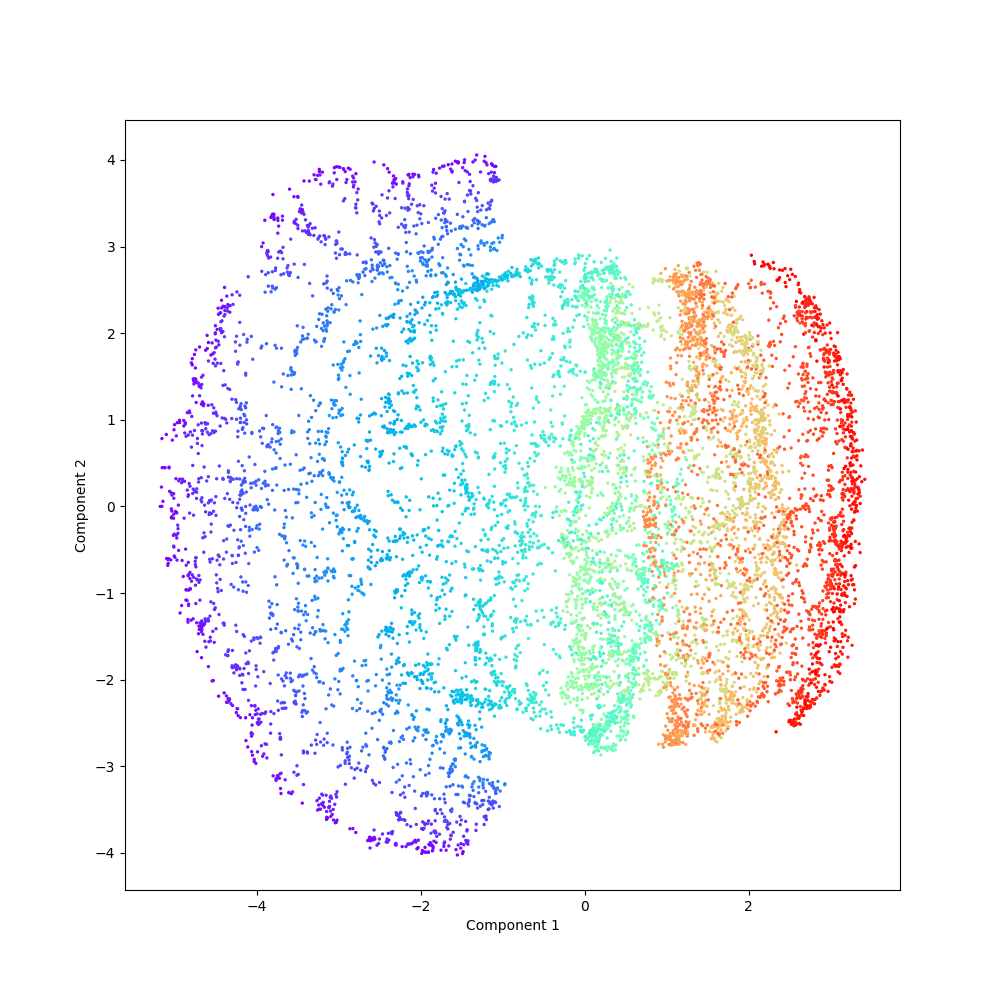} &
		\includegraphics[width=0.10\textwidth]{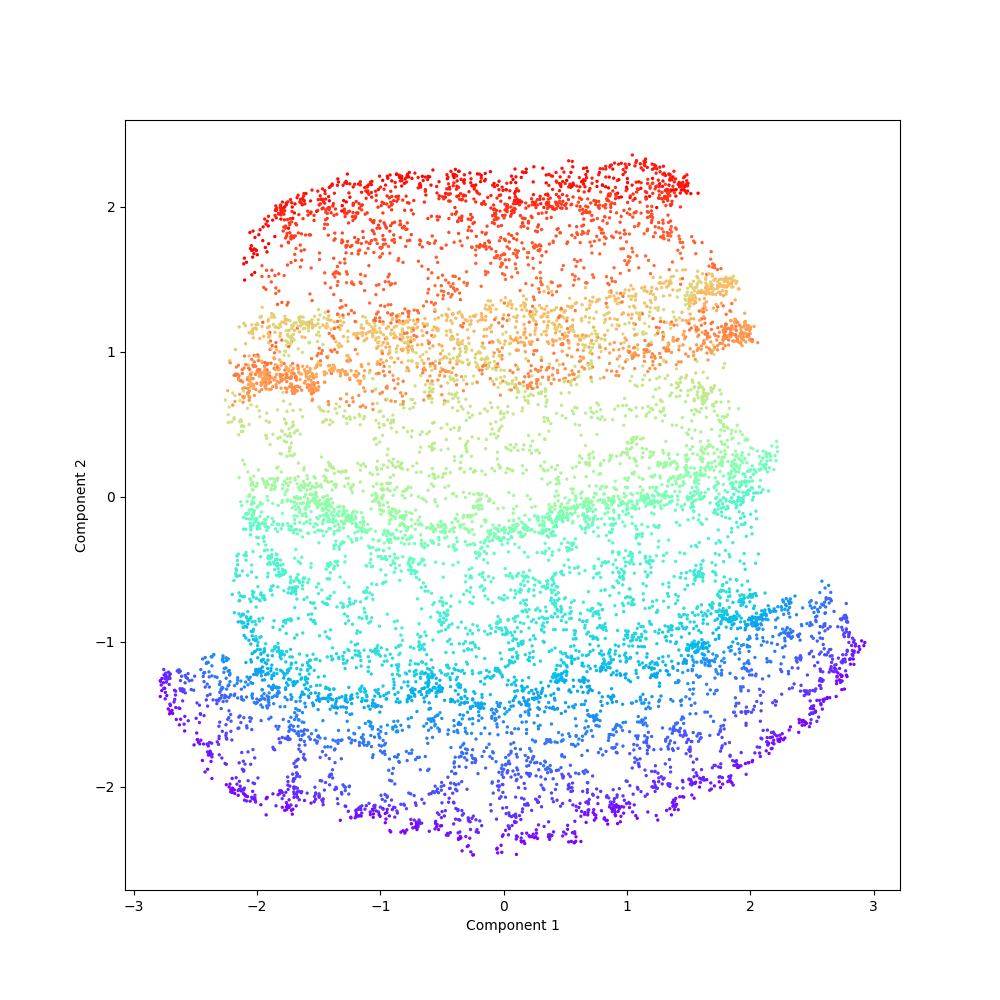}&
		\includegraphics[width=0.10\textwidth]{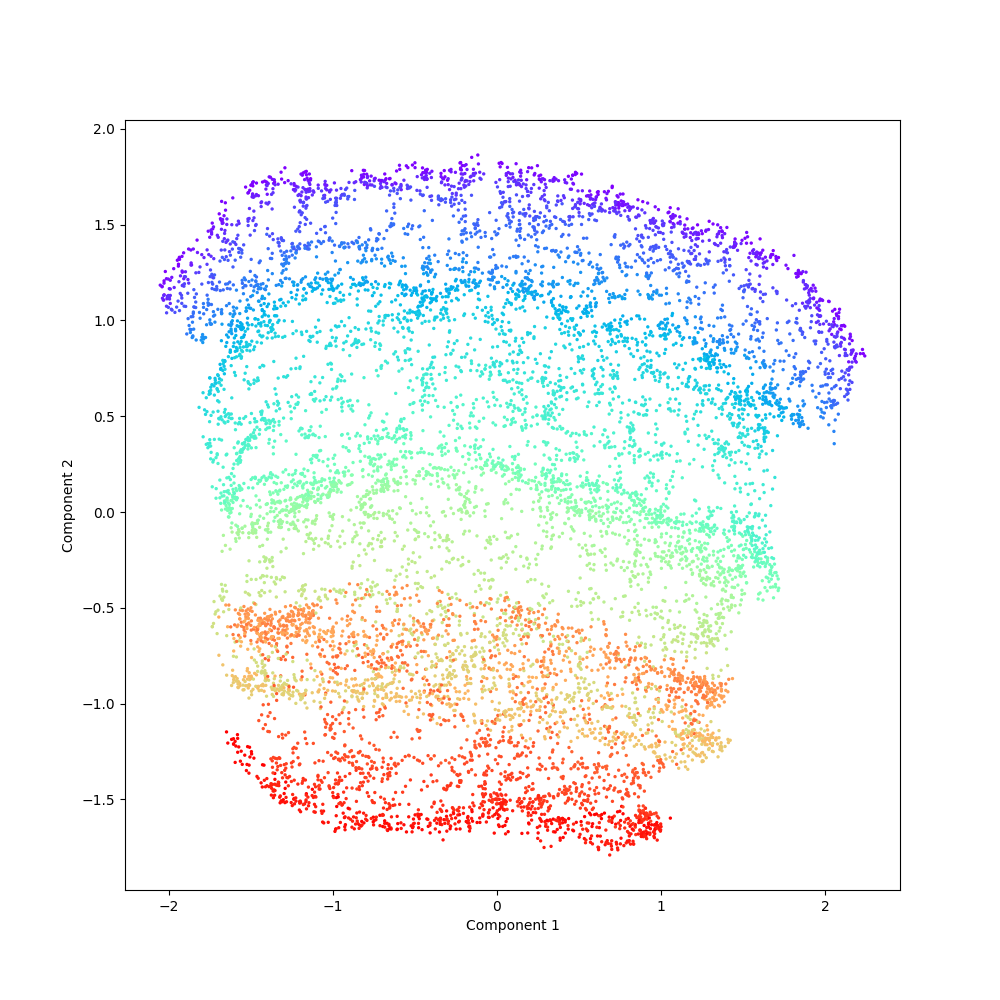} \\
		\textbf{(2)  $\cos(5t)$} & 		 \includegraphics[width=0.10\textwidth]{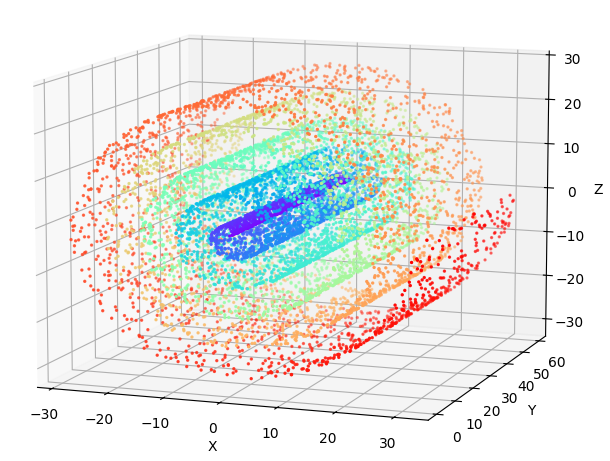} &
		\includegraphics[width=0.10\textwidth]{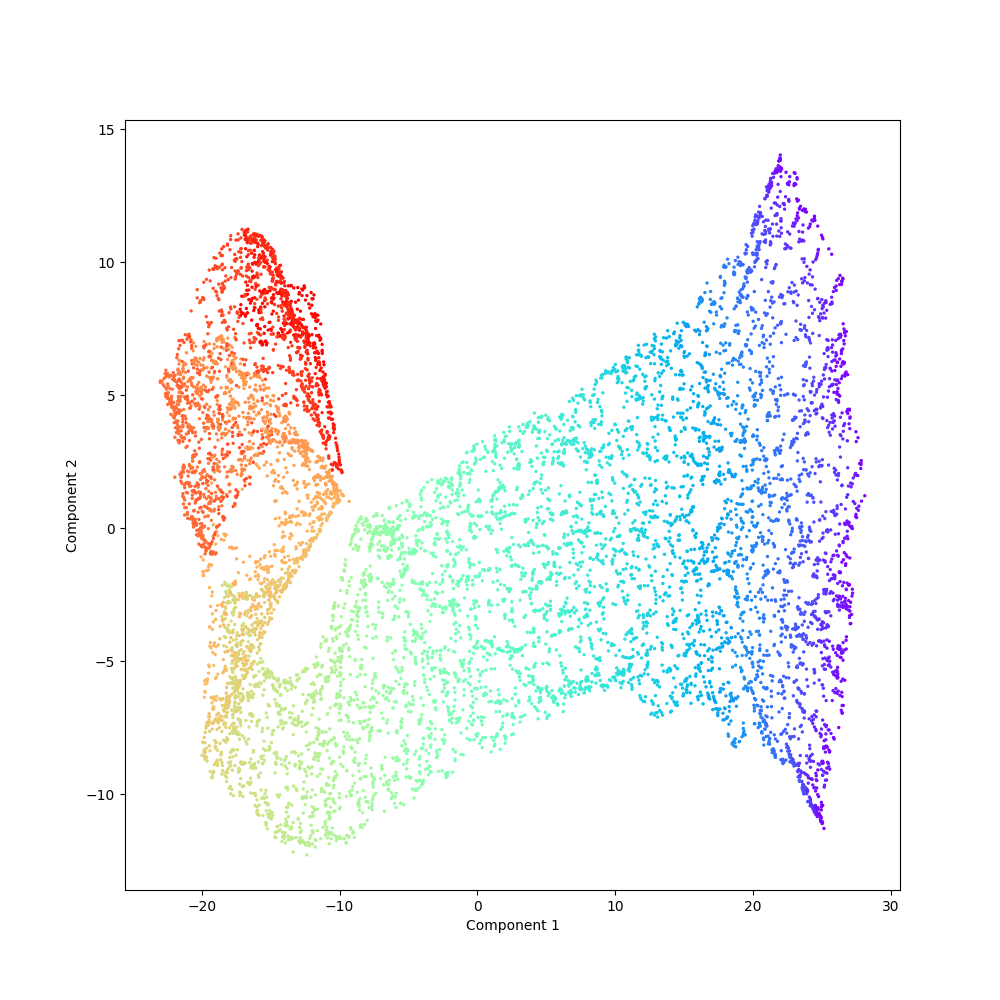}&
		\includegraphics[width=0.10\textwidth]{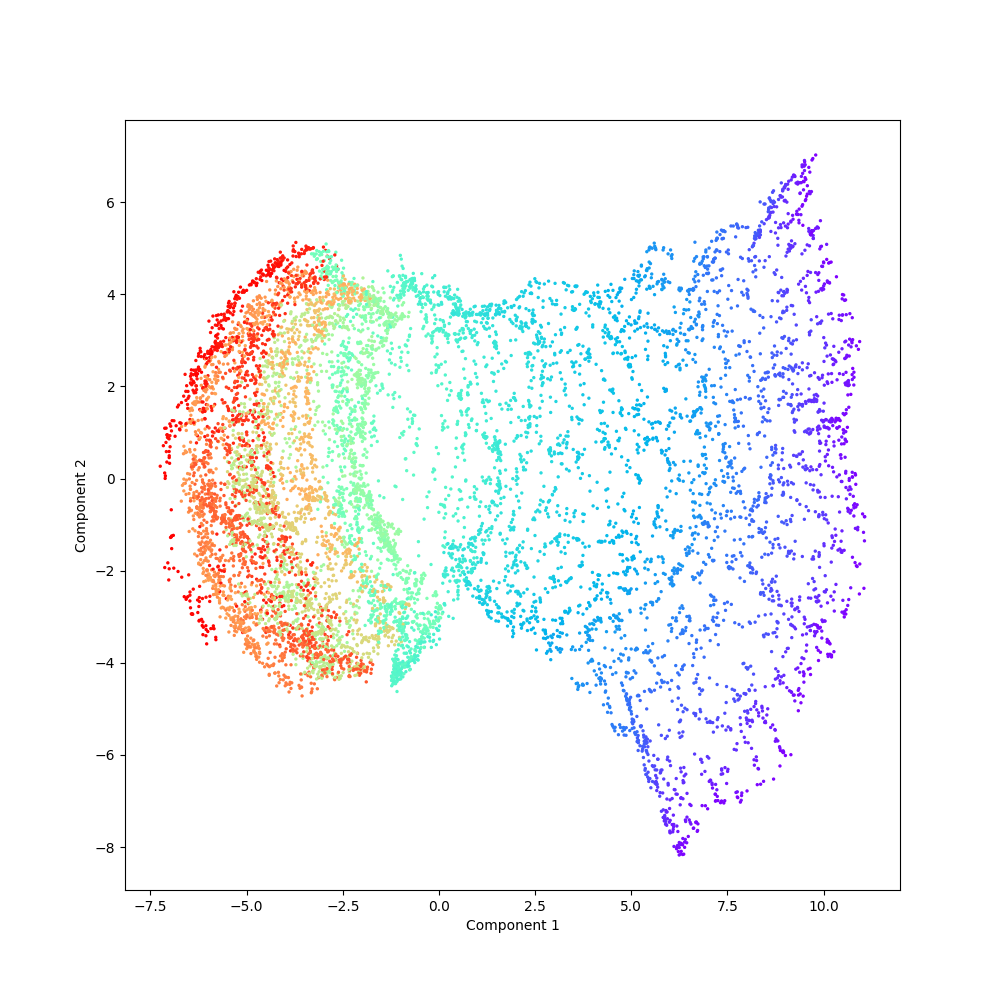} & \includegraphics[width=0.10\textwidth]{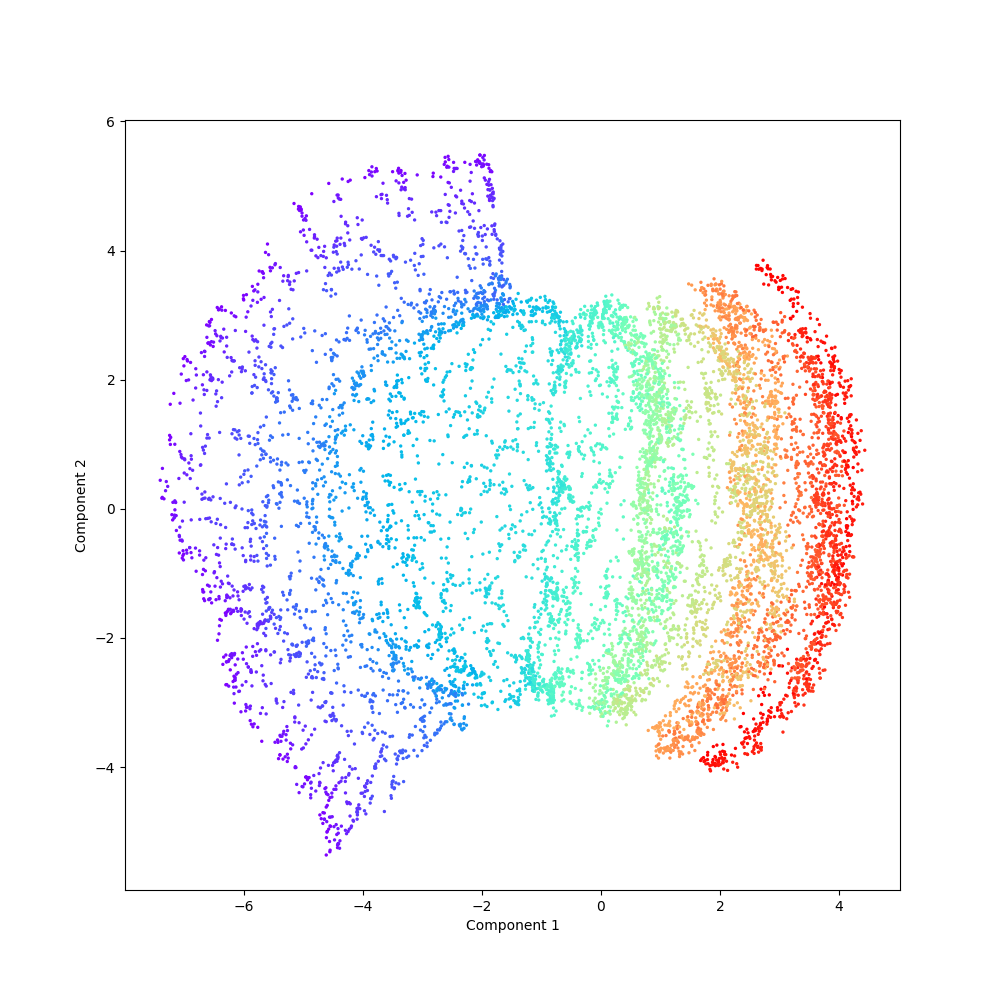} &
		\includegraphics[width=0.10\textwidth]{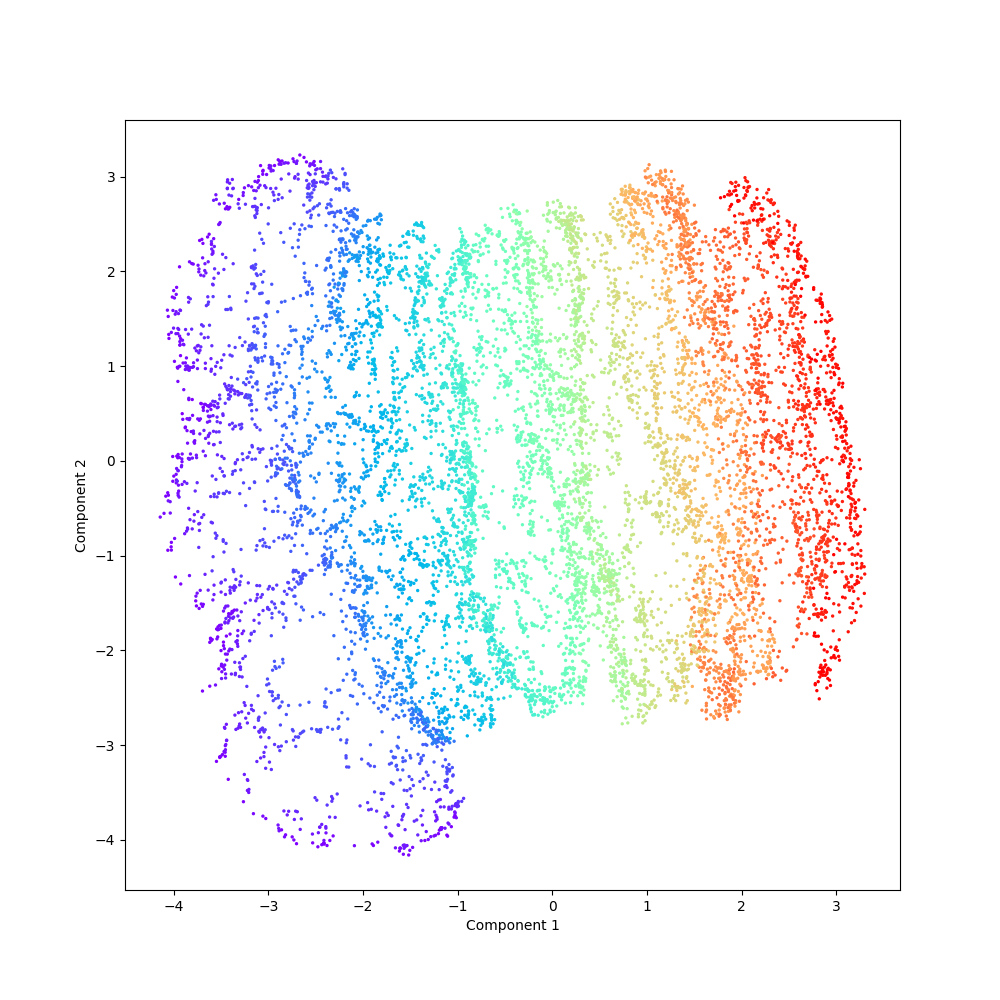} &
		\includegraphics[width=0.10\textwidth]{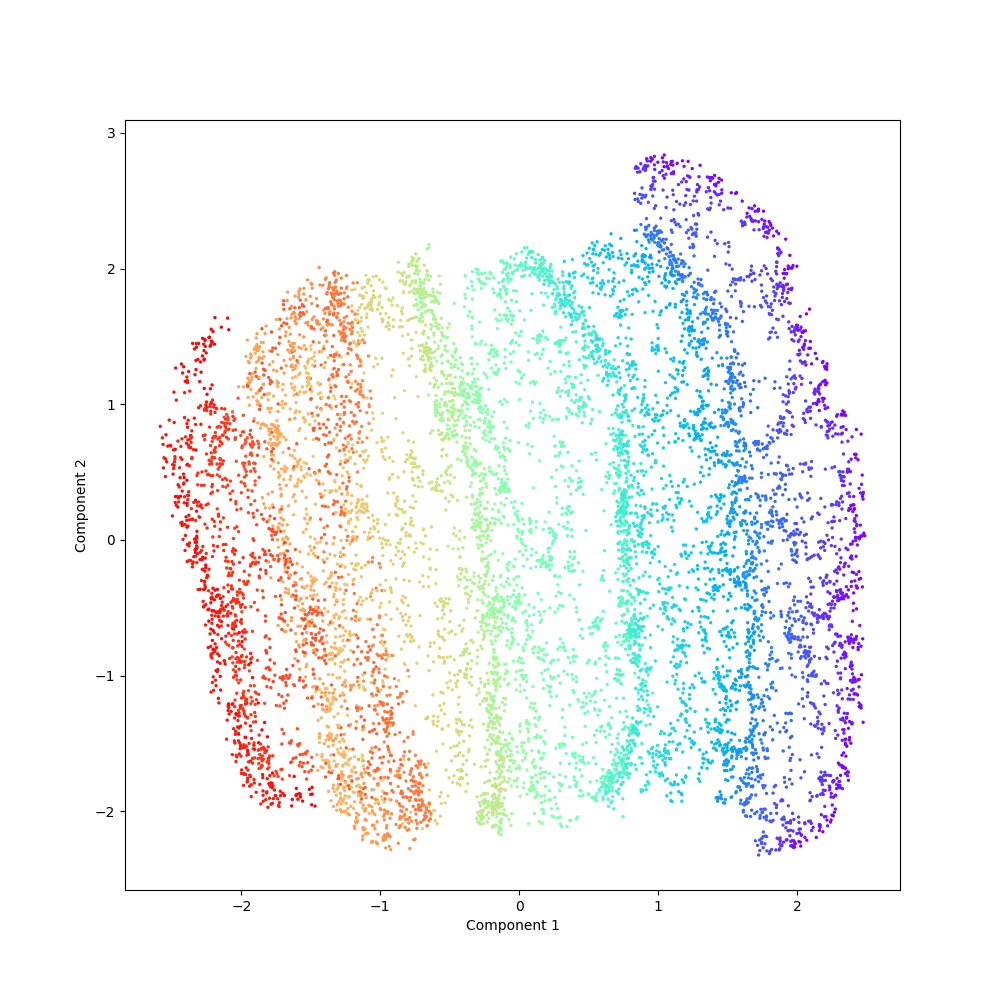}&
		\includegraphics[width=0.10\textwidth]{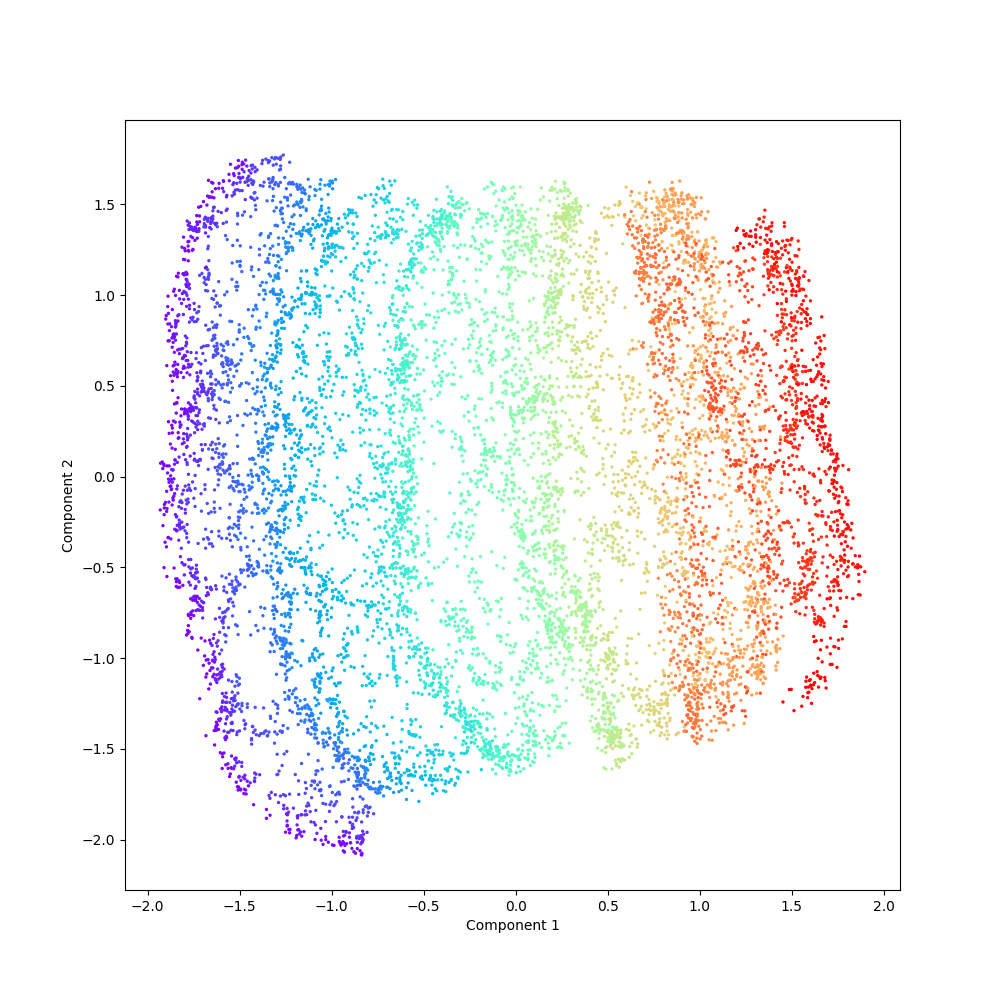}\\
		\textbf{(3) $\cos(6t)$}& 
		\includegraphics[width=0.10\textwidth]{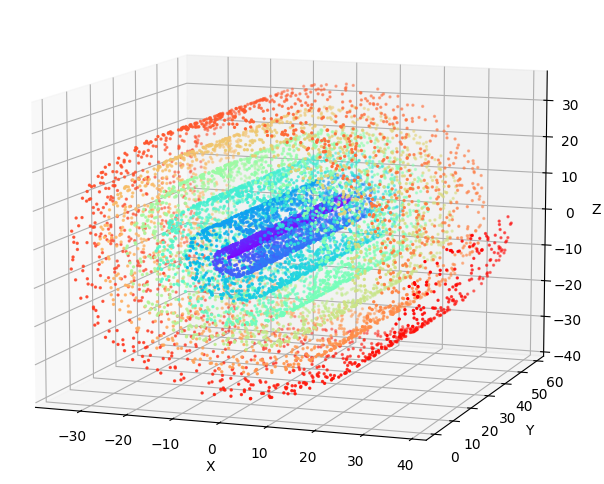} &
		\includegraphics[width=0.10\textwidth]{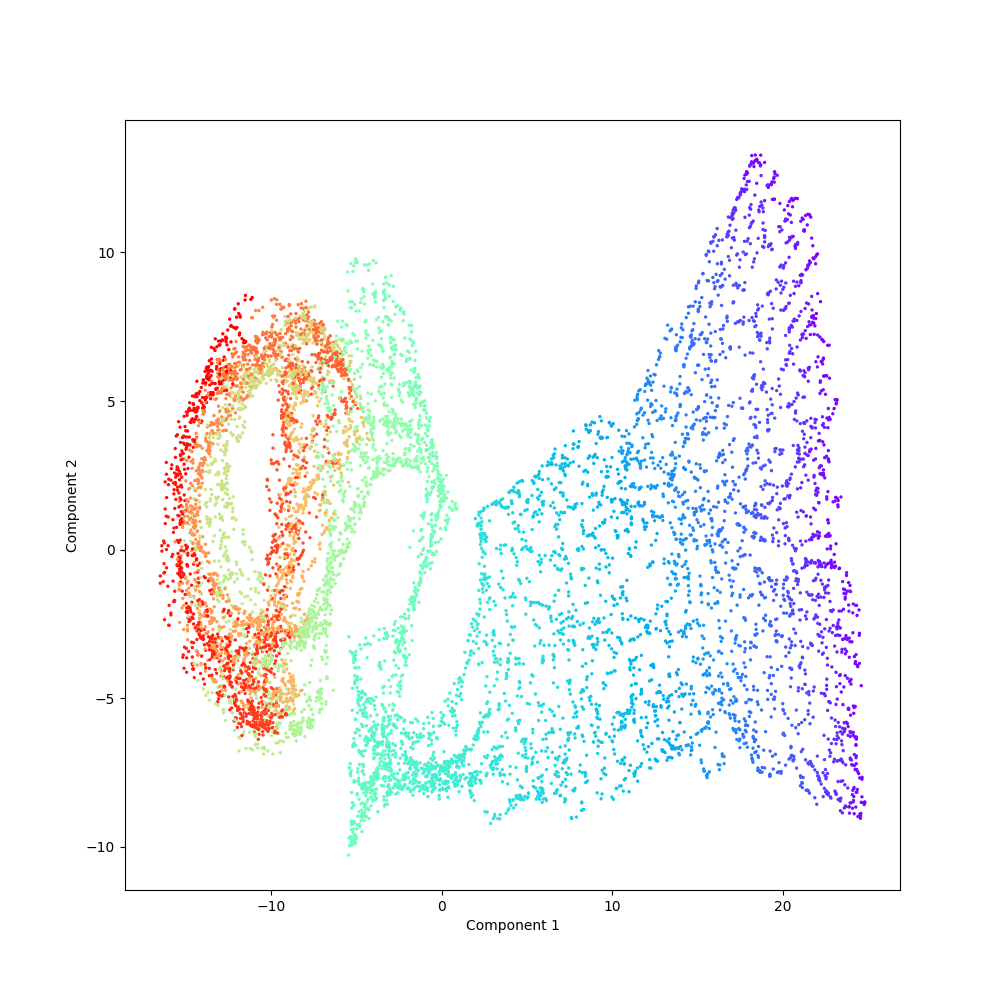}&
		\includegraphics[width=0.10\textwidth]{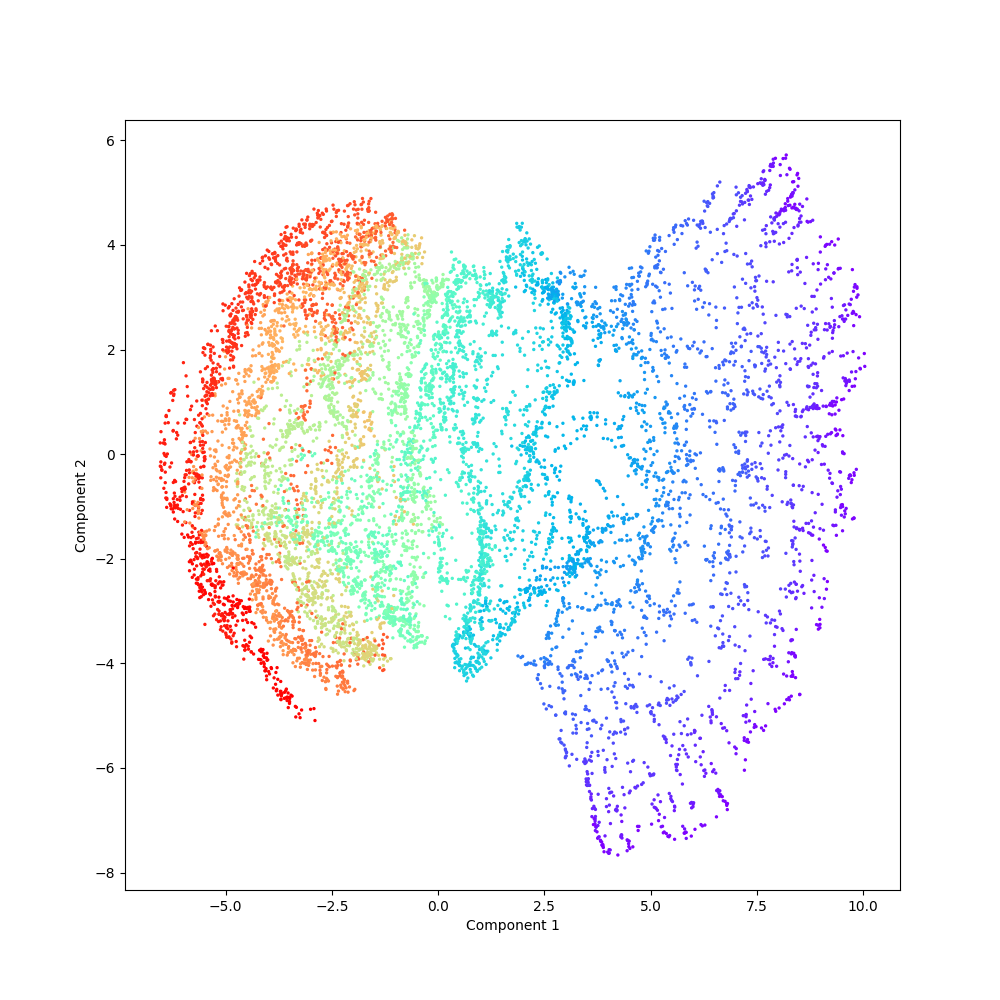} & \includegraphics[width=0.10\textwidth]{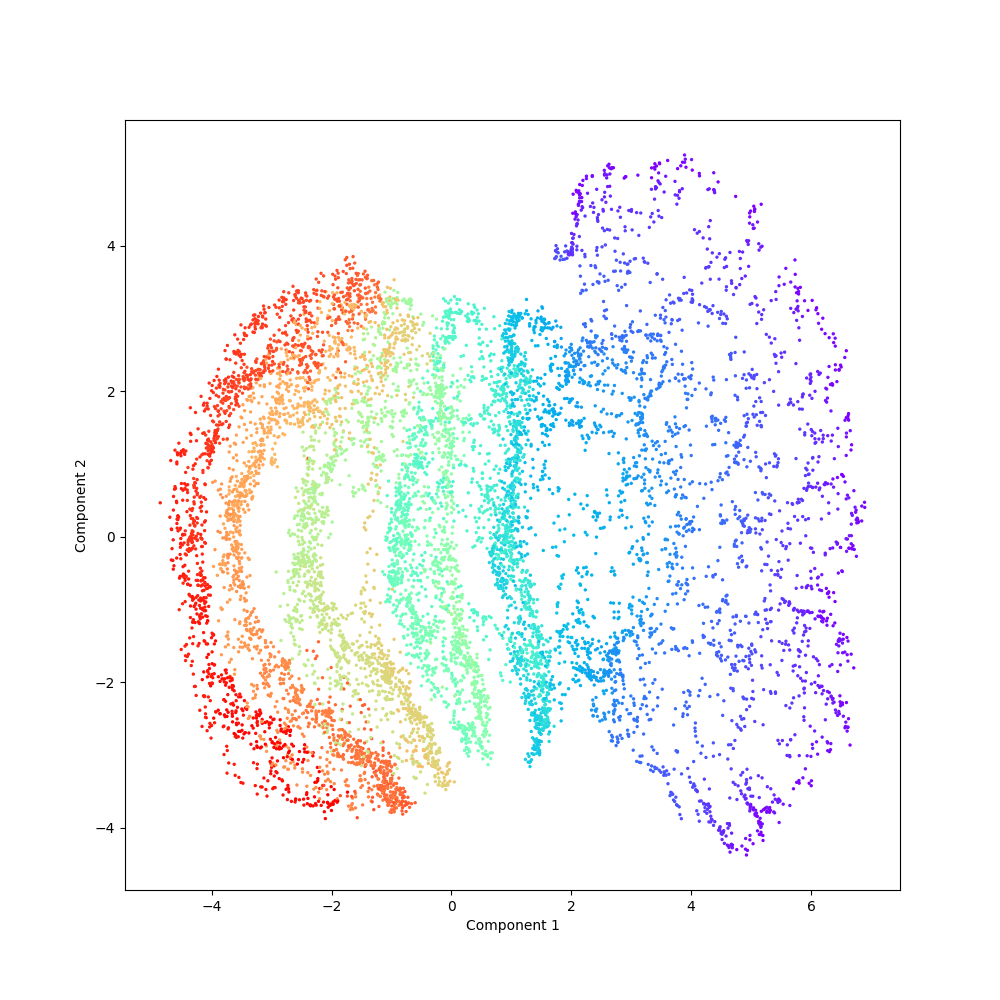} &
		\includegraphics[width=0.10\textwidth]{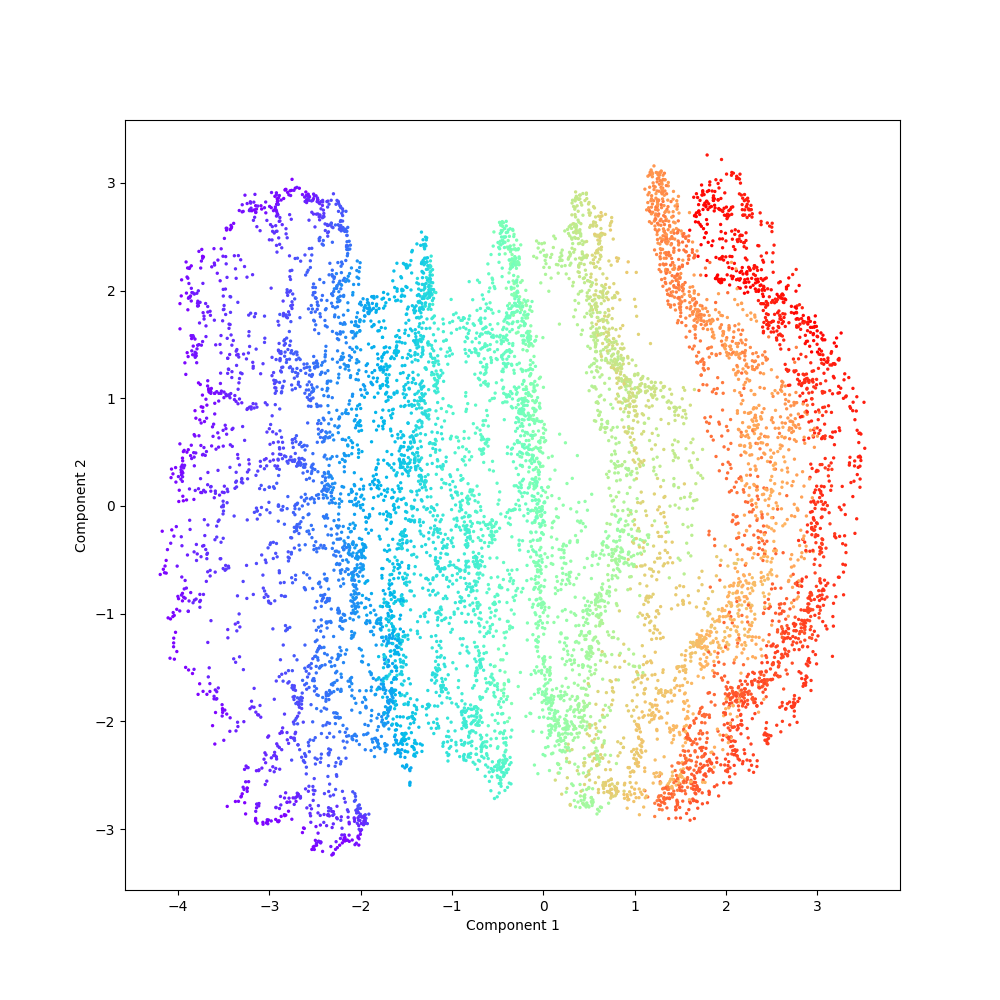} &
		\includegraphics[width=0.10\textwidth]{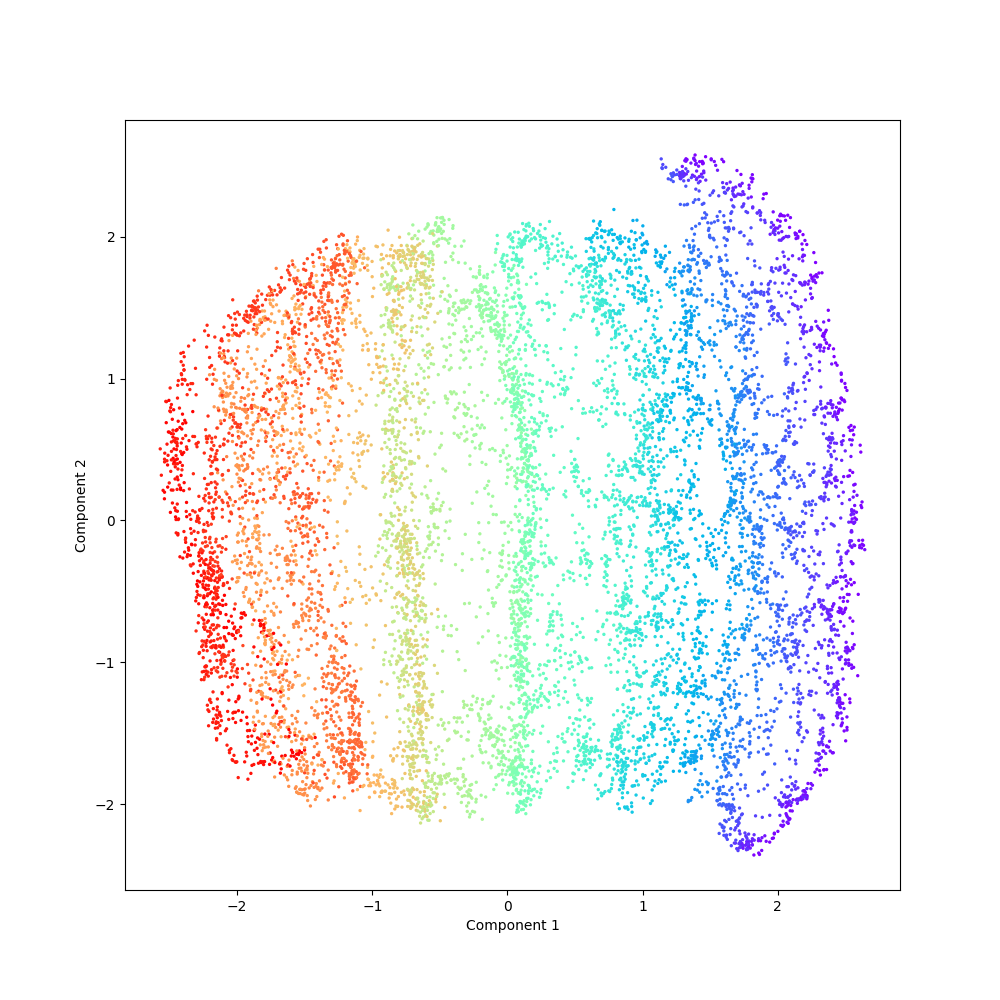}&\includegraphics[width=0.10\textwidth]{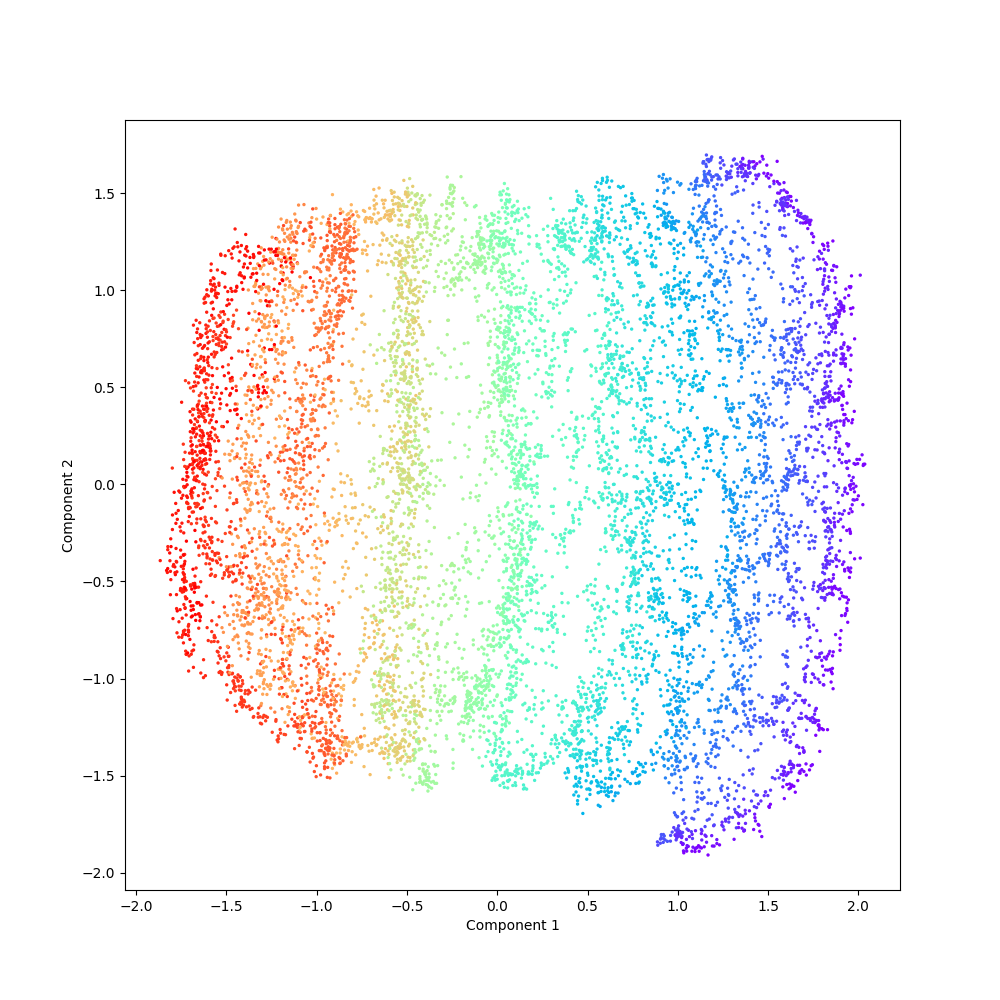}
	\end{tabular}
	\caption{Swiss Roll IsUMap embedding with different $k$.}
	\label{tab:IsUMap_diffNN_SwissRoll}
\end{table}

\begin{table}[H]
	\centering
	\tiny
	\begin{tabular}{>{\centering\arraybackslash}m{0.5cm}|*{7}{>{\centering\arraybackslash}m{1cm}}}
		& \textbf{Data set} & \textbf{k=10} & \textbf{k=20}  & \textbf{k=30}  & \textbf{k=50}& \textbf{k=100}&\textbf{k=200}\\
		& (a) & (b) & (c) & (d)&(e)&(f)&(g) \\
		\hline \\
		\textbf{(1)  $\cos(4t)$} &
		\includegraphics[width=0.10\textwidth]{image/further_rollig_swissRoll/original_data_8.png} & \includegraphics[width=0.10\textwidth]{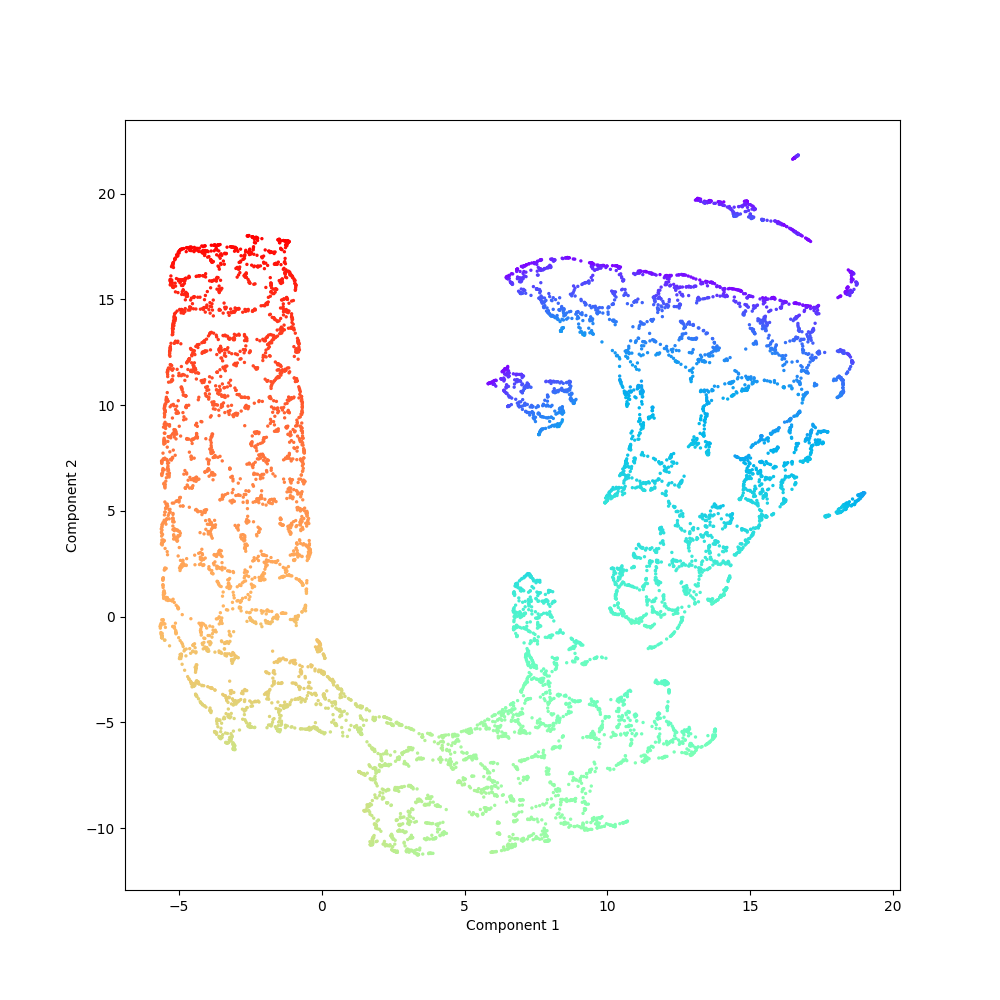} & \includegraphics[width=0.10\textwidth]{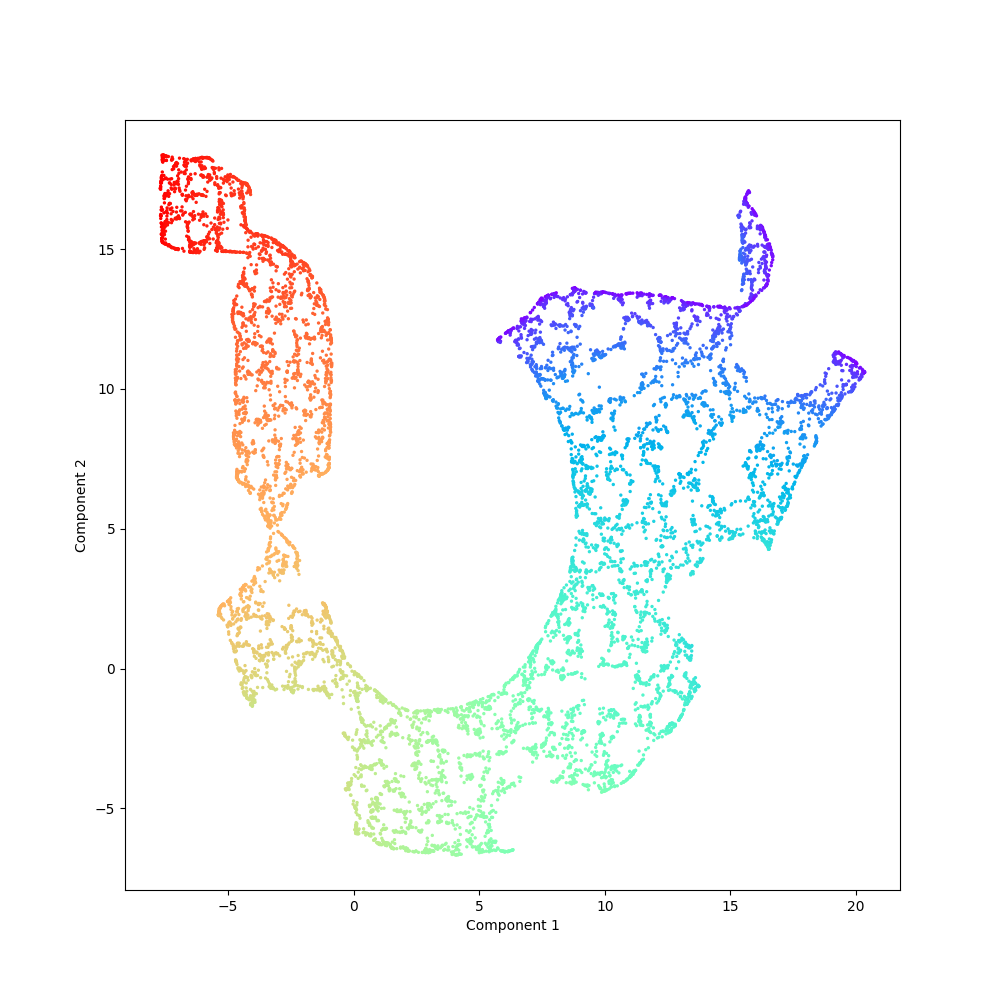} &
		\includegraphics[width=0.10\textwidth]{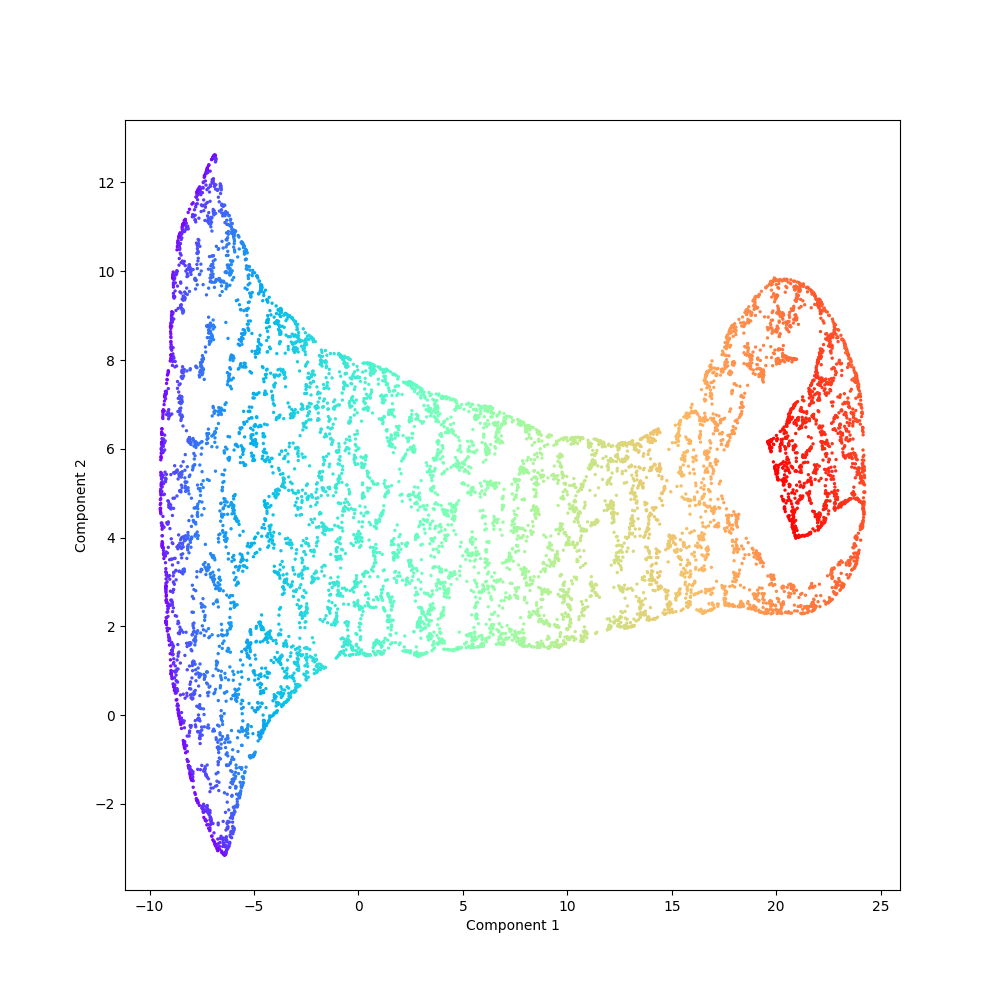} &
		\includegraphics[width=0.10\textwidth]{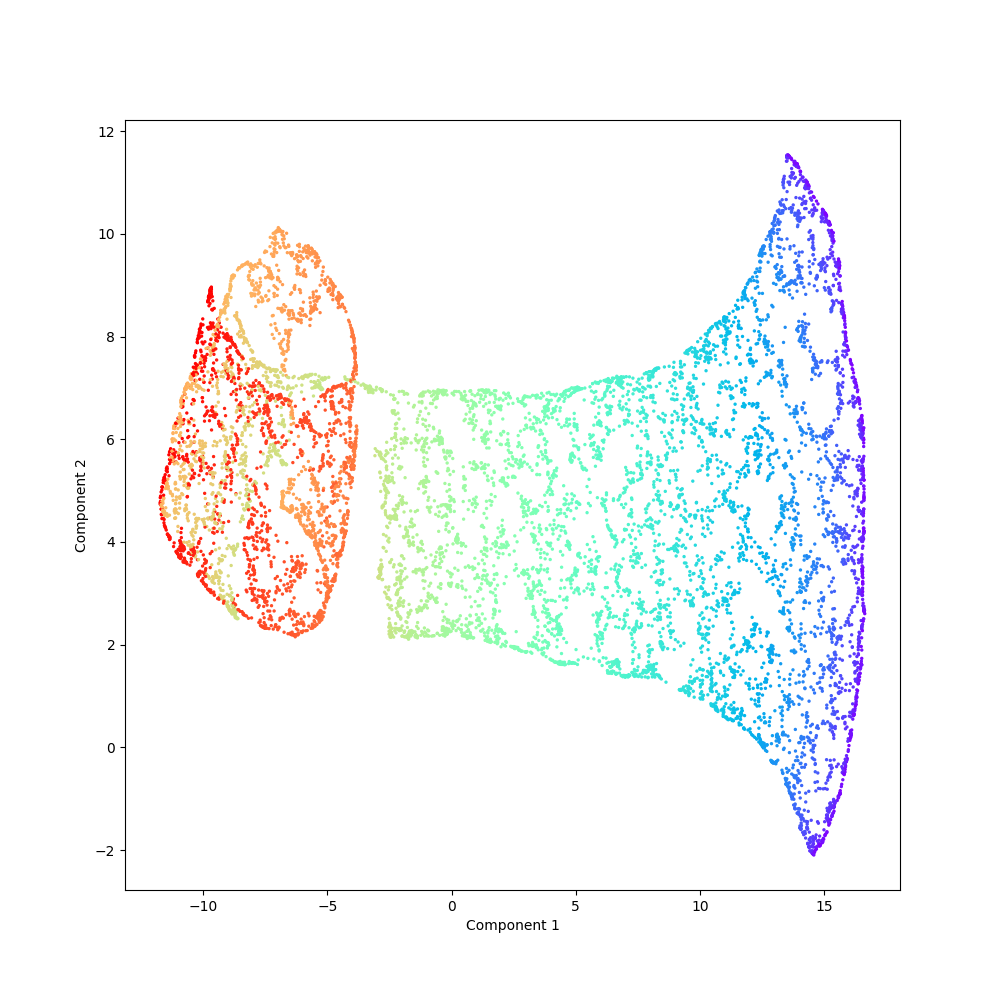} &
		\includegraphics[width=0.10\textwidth]{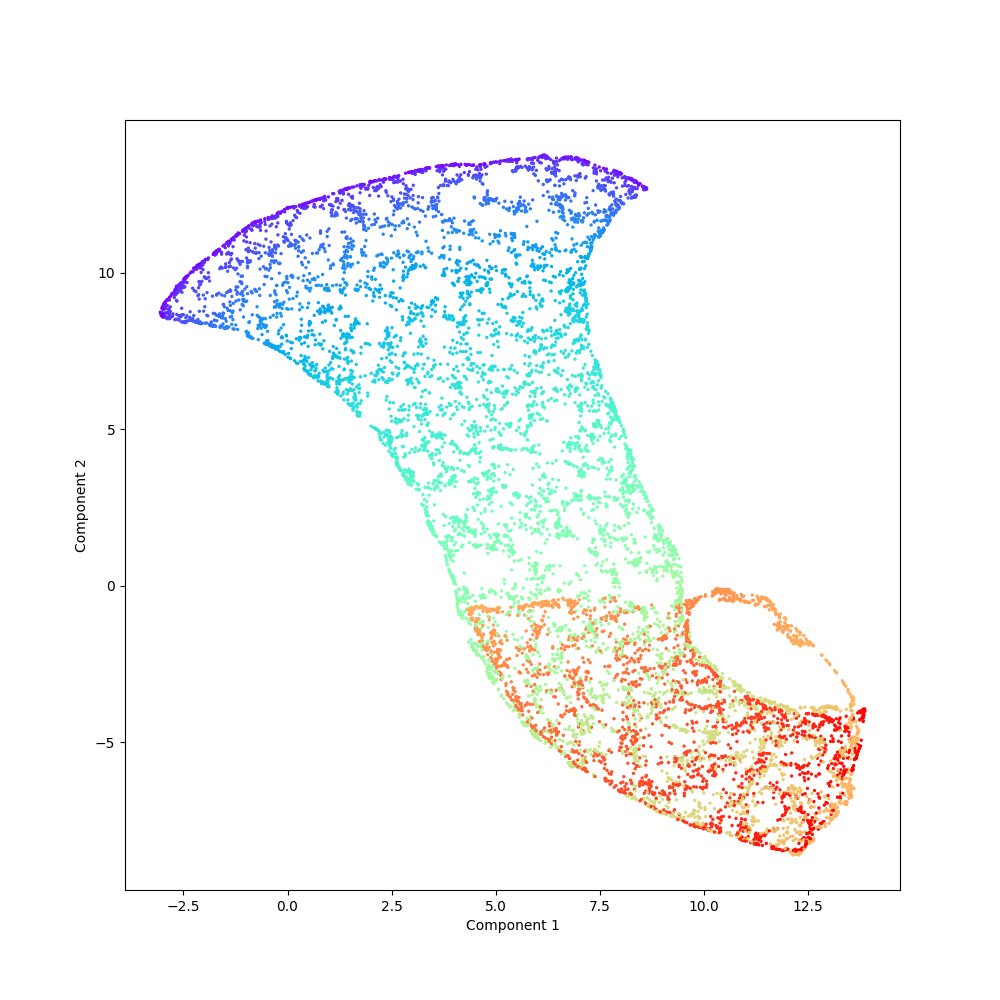}&
		\includegraphics[width=0.10\textwidth]{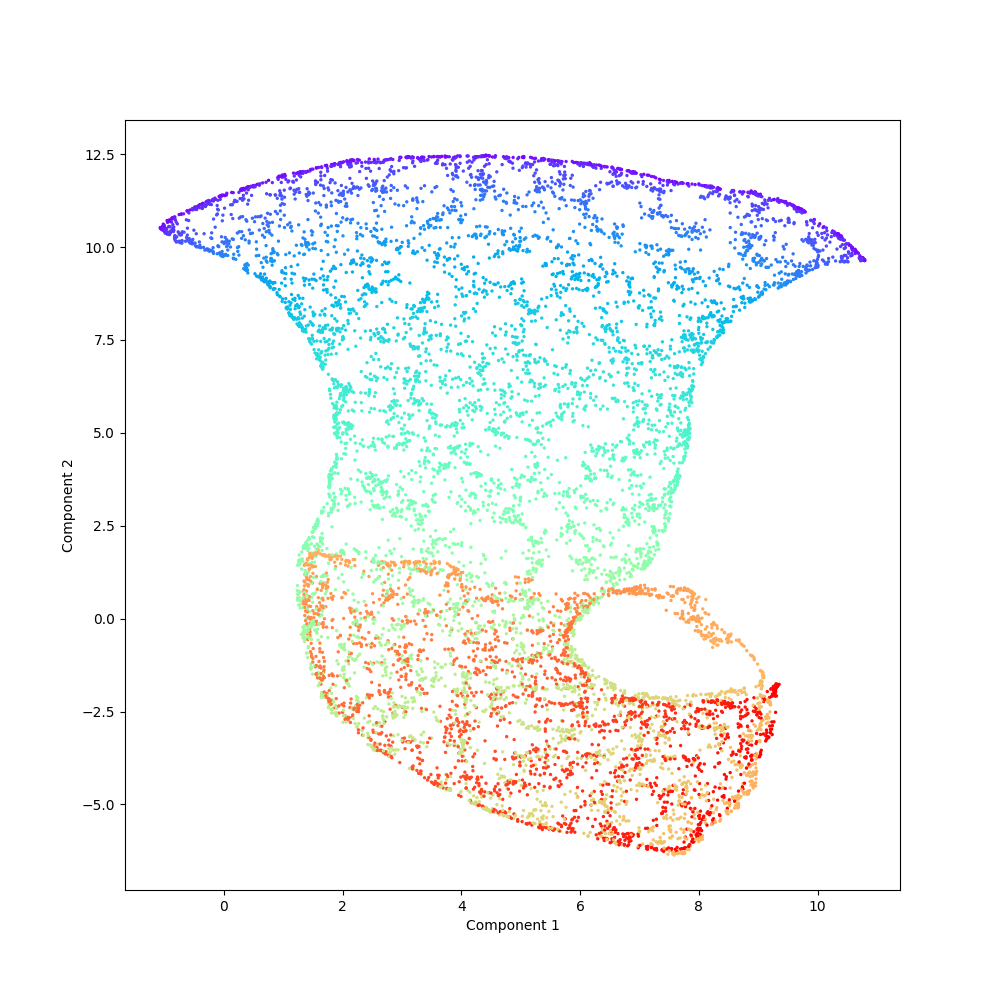} \\
		\textbf{(2)  $\cos(5t)$ } & 		 \includegraphics[width=0.10\textwidth]{image/further_rollig_swissRoll/original_data_10.png} &
		\includegraphics[width=0.10\textwidth]{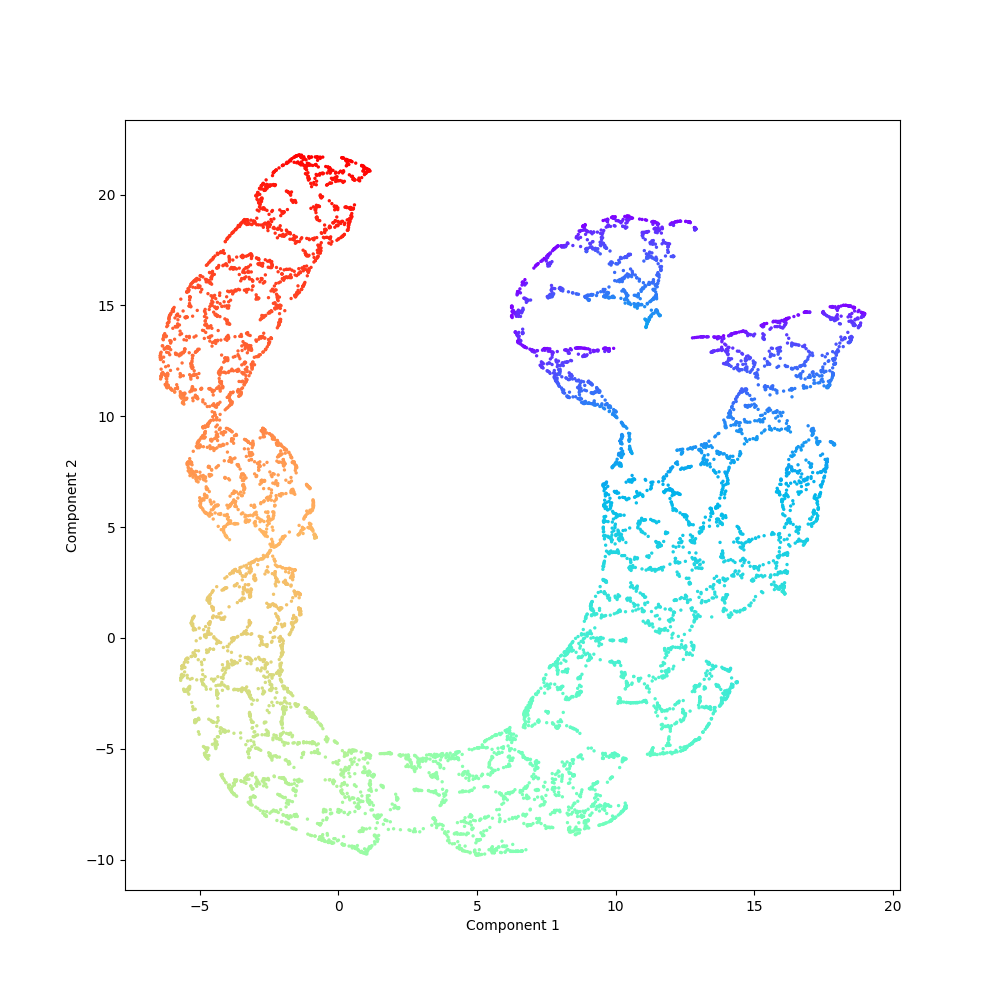}&
		\includegraphics[width=0.10\textwidth]{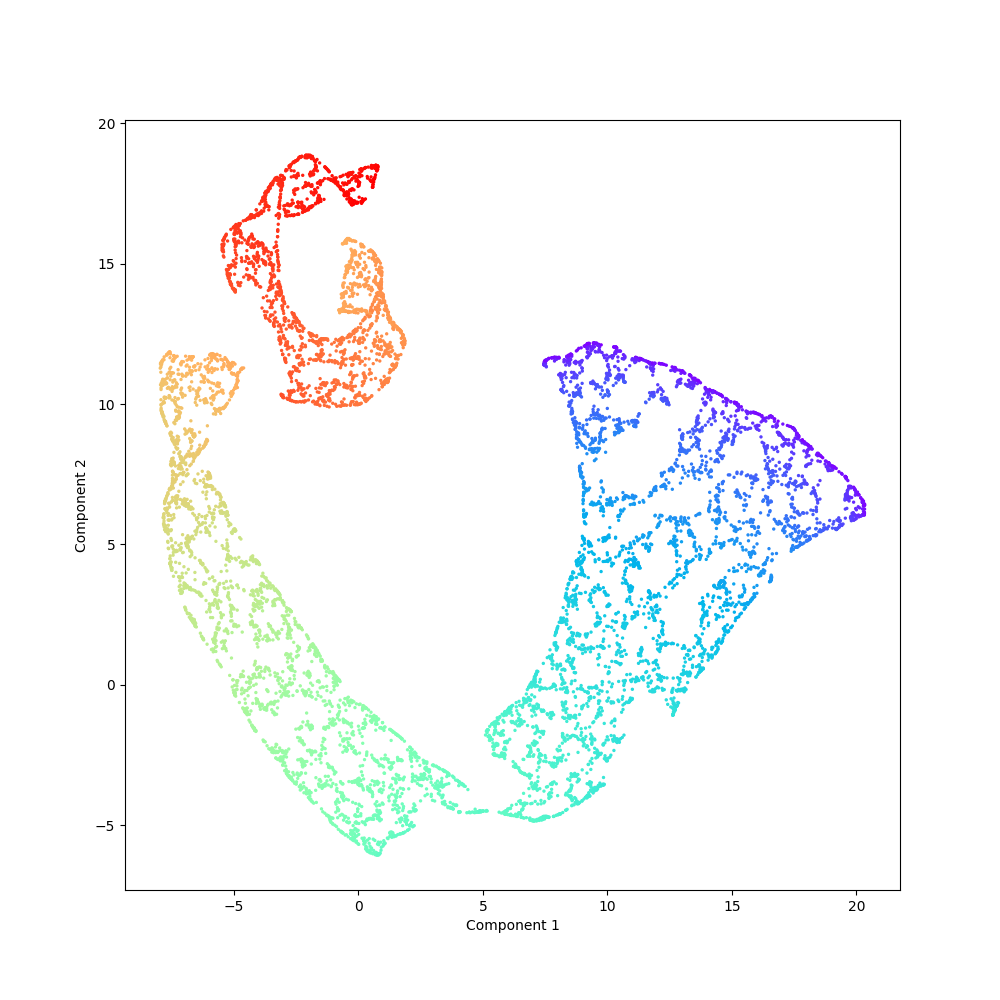} & \includegraphics[width=0.10\textwidth]{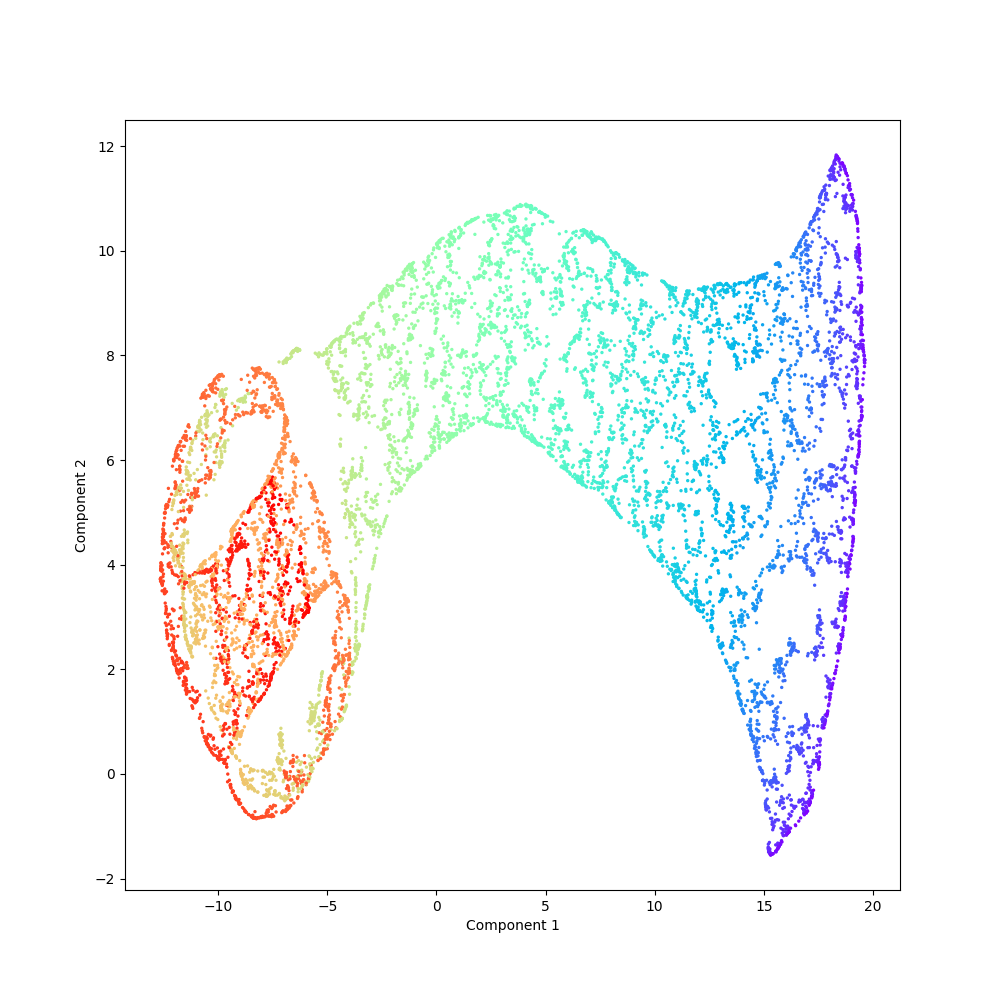} &
		\includegraphics[width=0.10\textwidth]{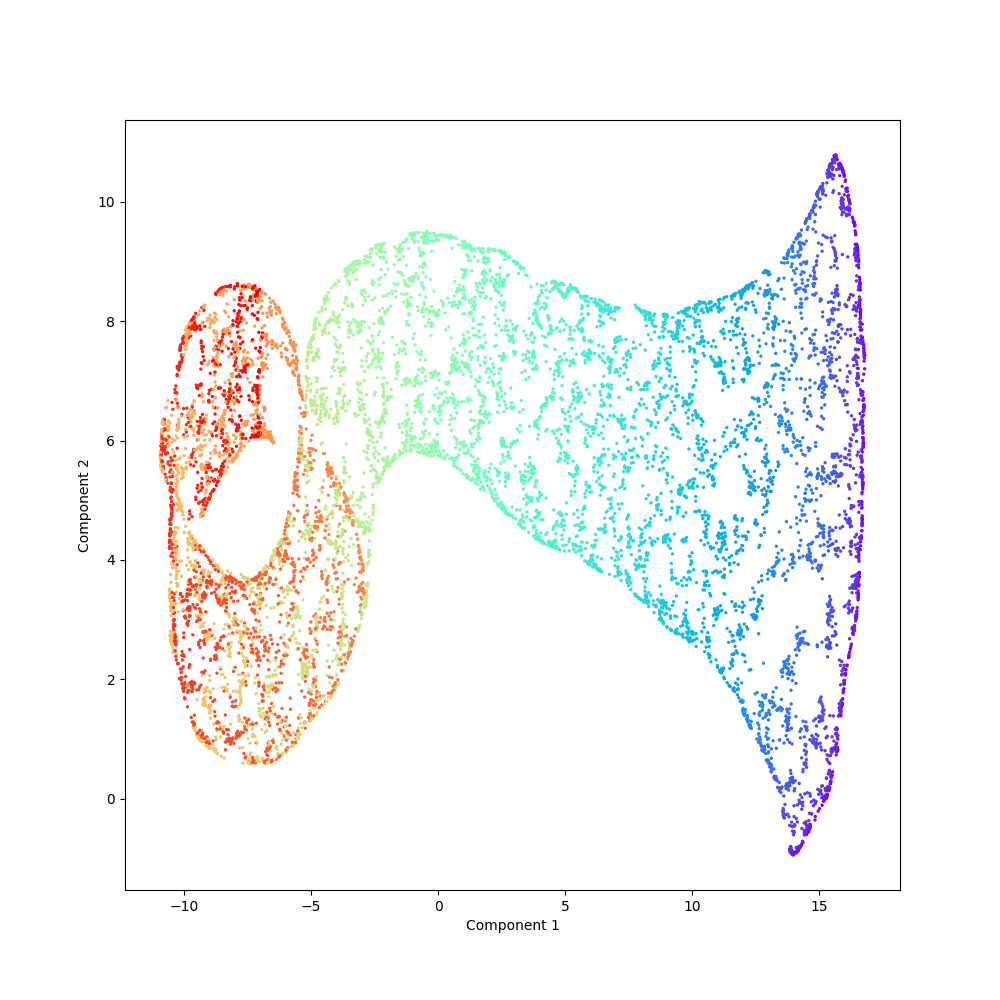} &
		\includegraphics[width=0.10\textwidth]{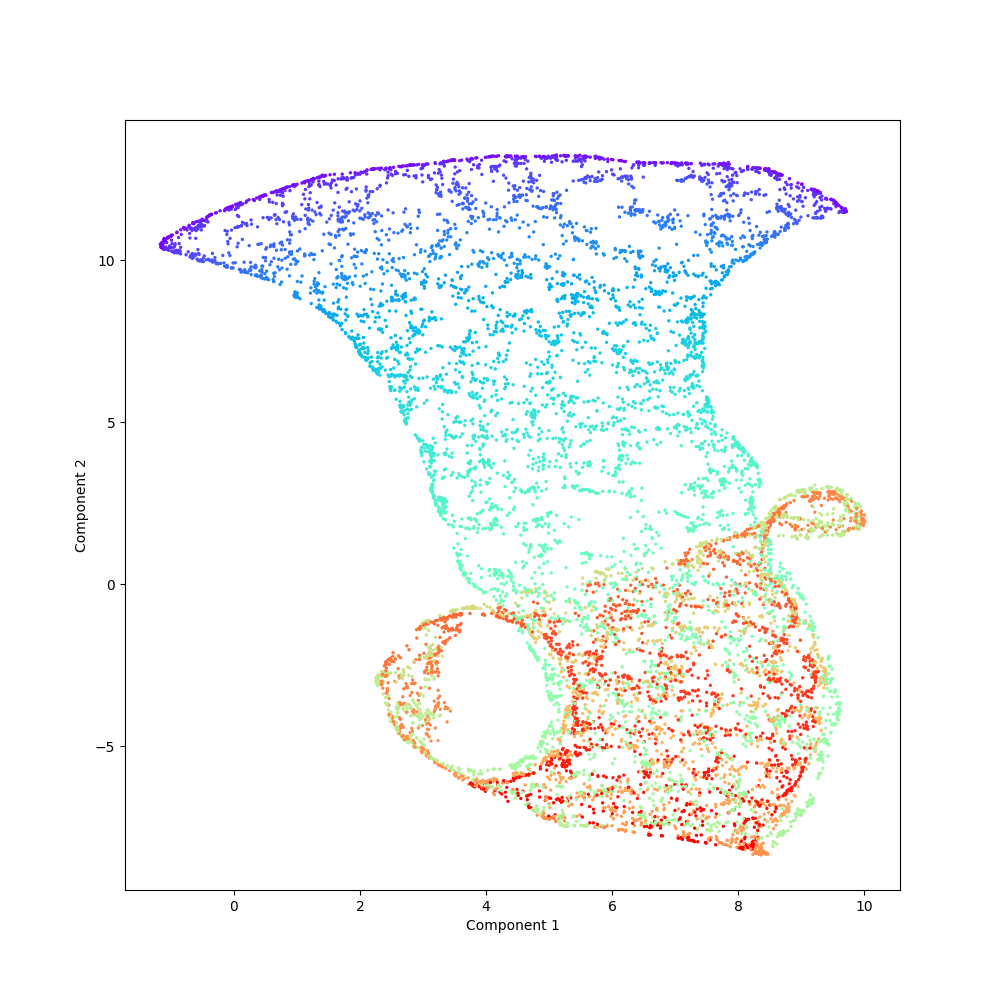}&
		\includegraphics[width=0.10\textwidth]{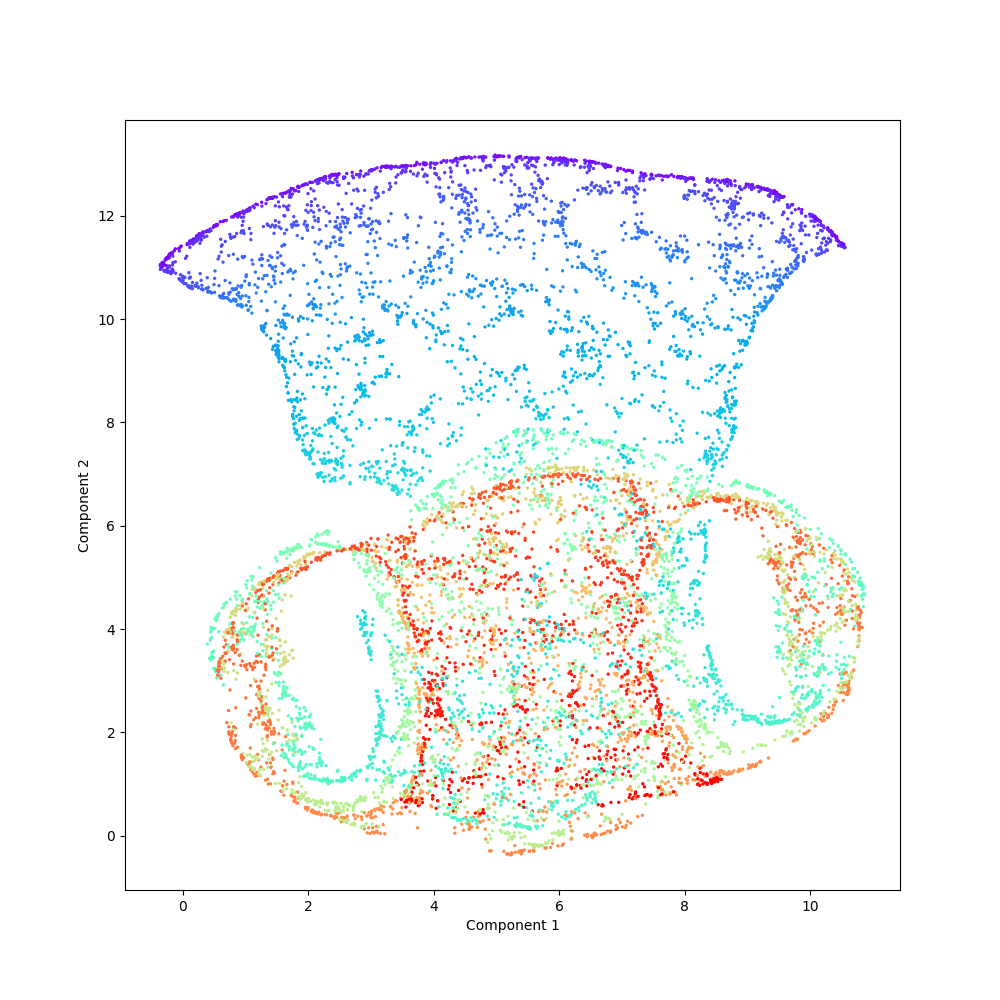}\\
		\textbf{(3)  $	\cos(6t)$}& 
		\includegraphics[width=0.10\textwidth]{image/further_rollig_swissRoll/original_data_12.png} &
		\includegraphics[width=0.10\textwidth]{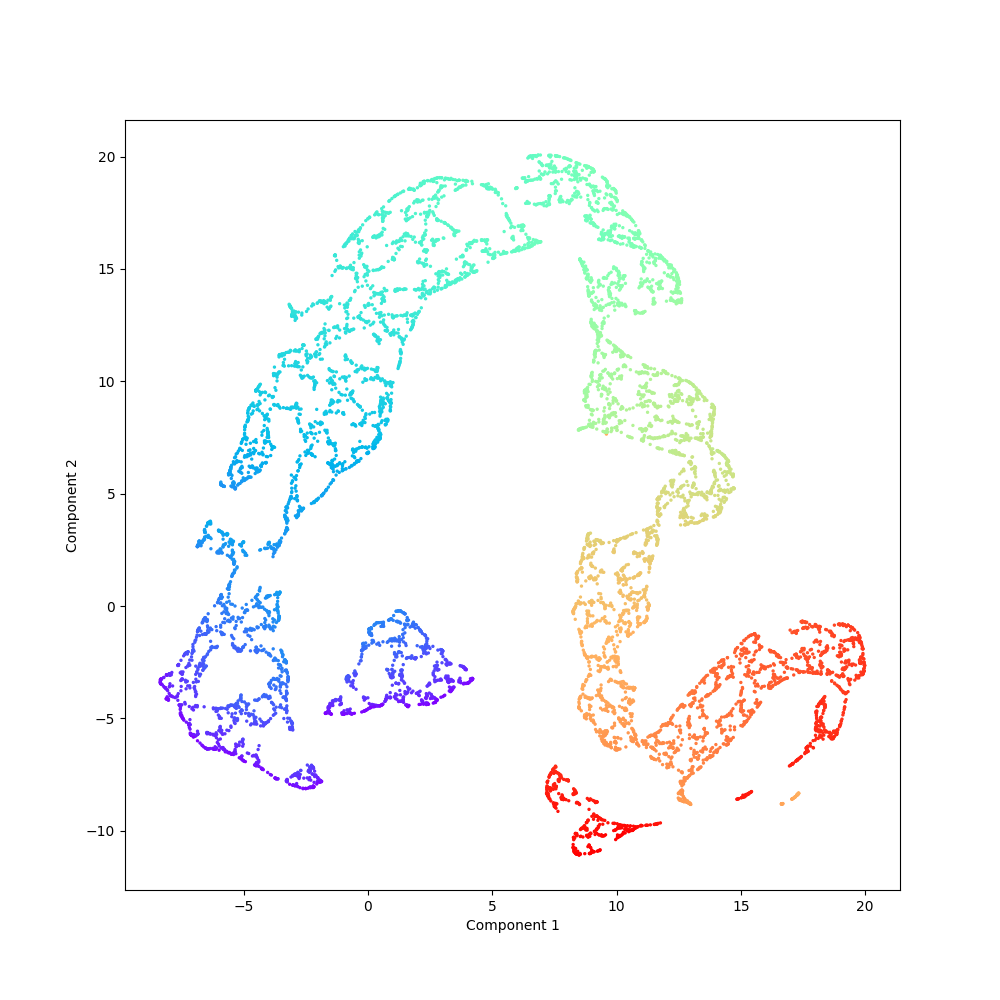}&
		\includegraphics[width=0.10\textwidth]{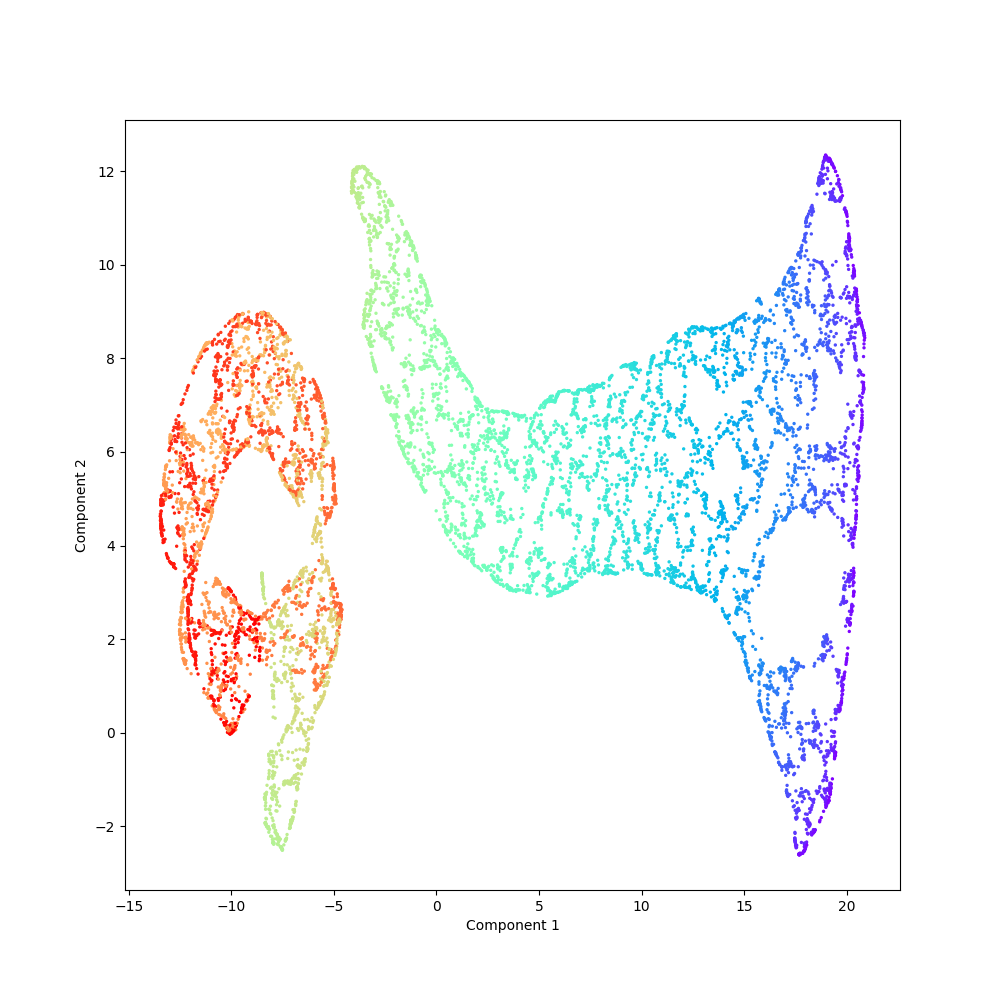} & \includegraphics[width=0.10\textwidth]{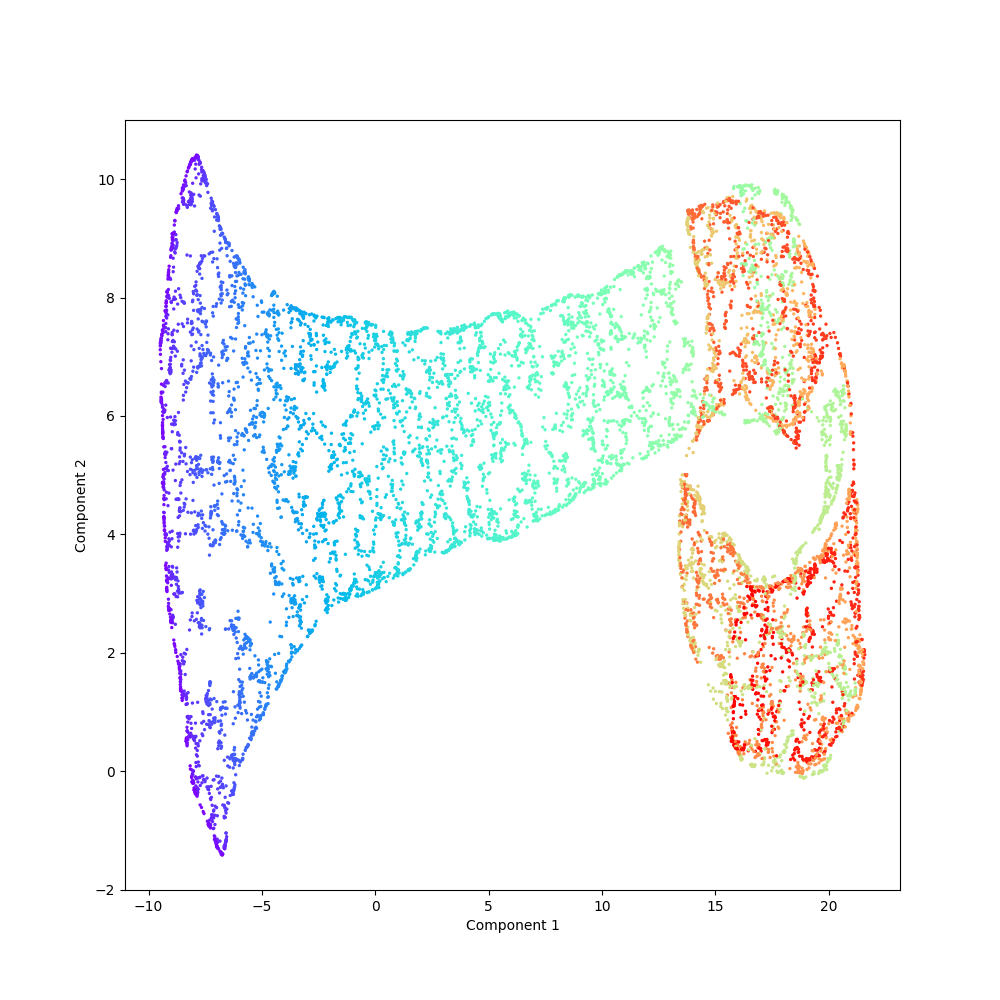} &
		\includegraphics[width=0.10\textwidth]{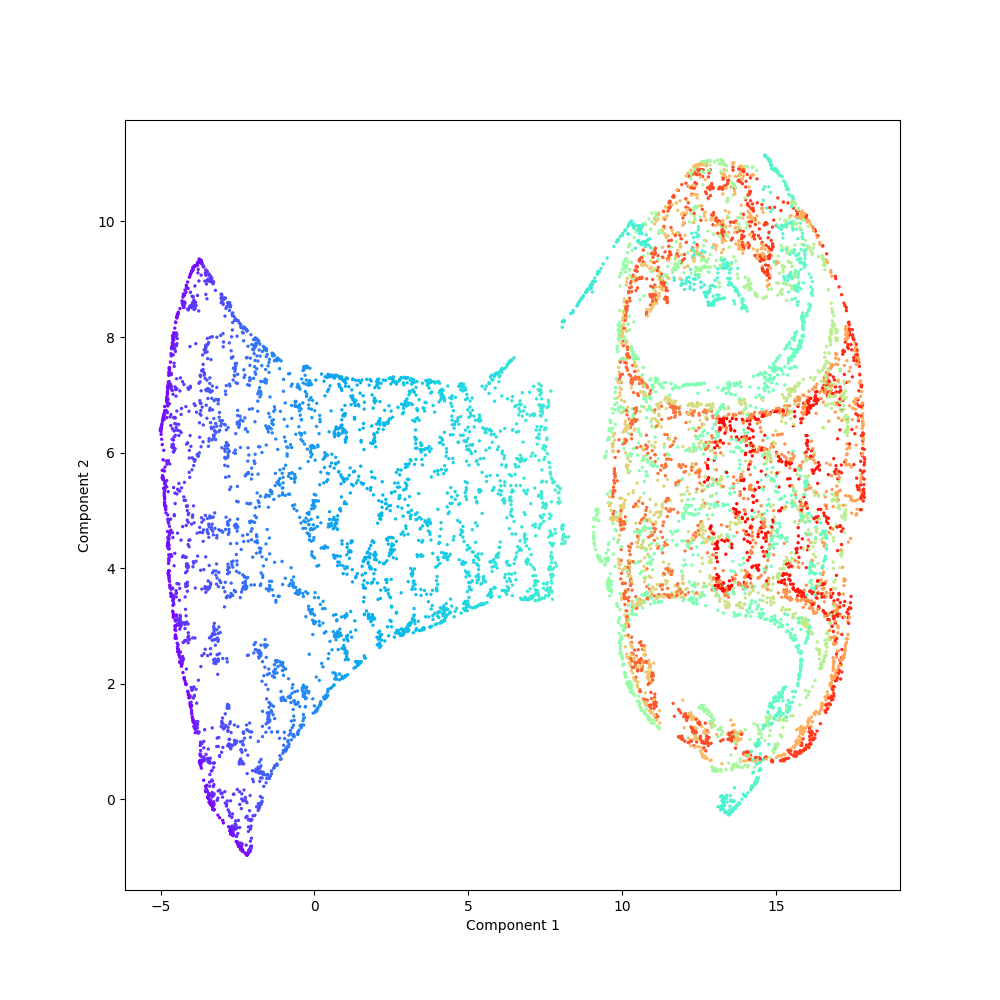} &
		\includegraphics[width=0.10\textwidth]{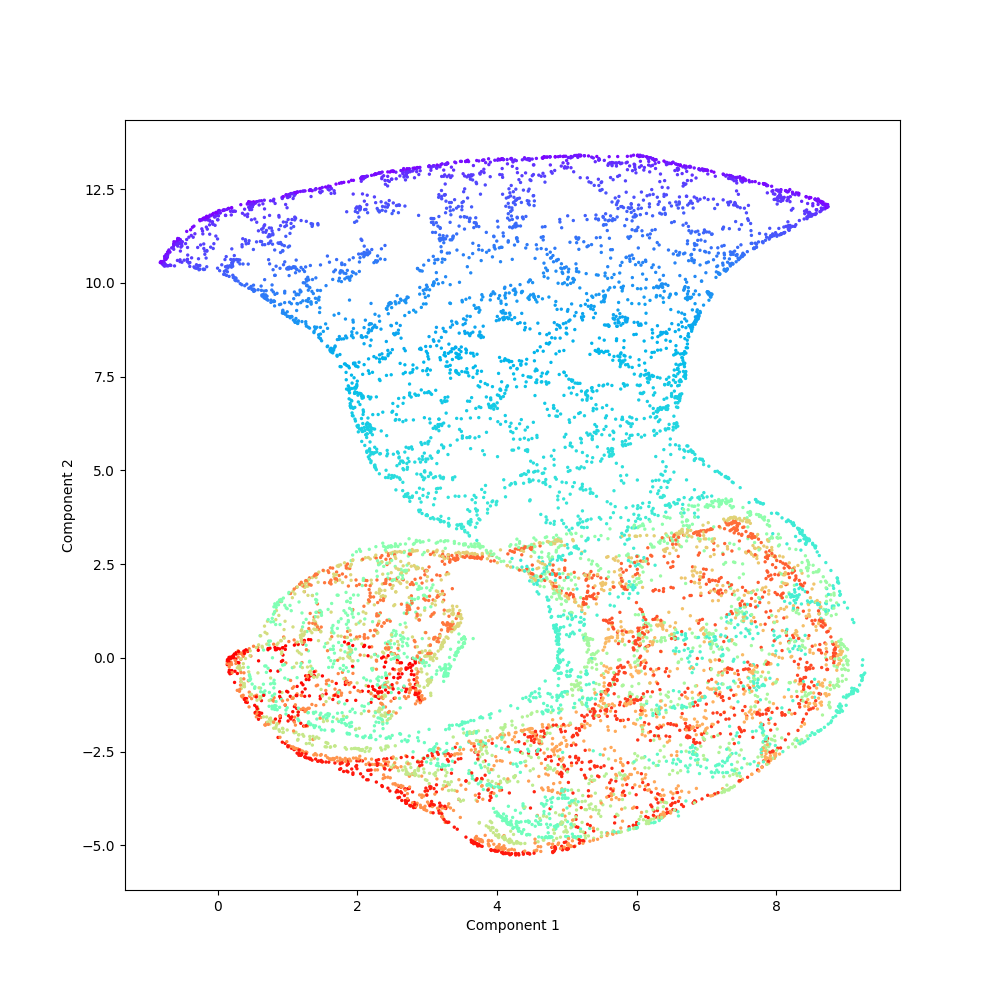}&\includegraphics[width=0.12\textwidth]{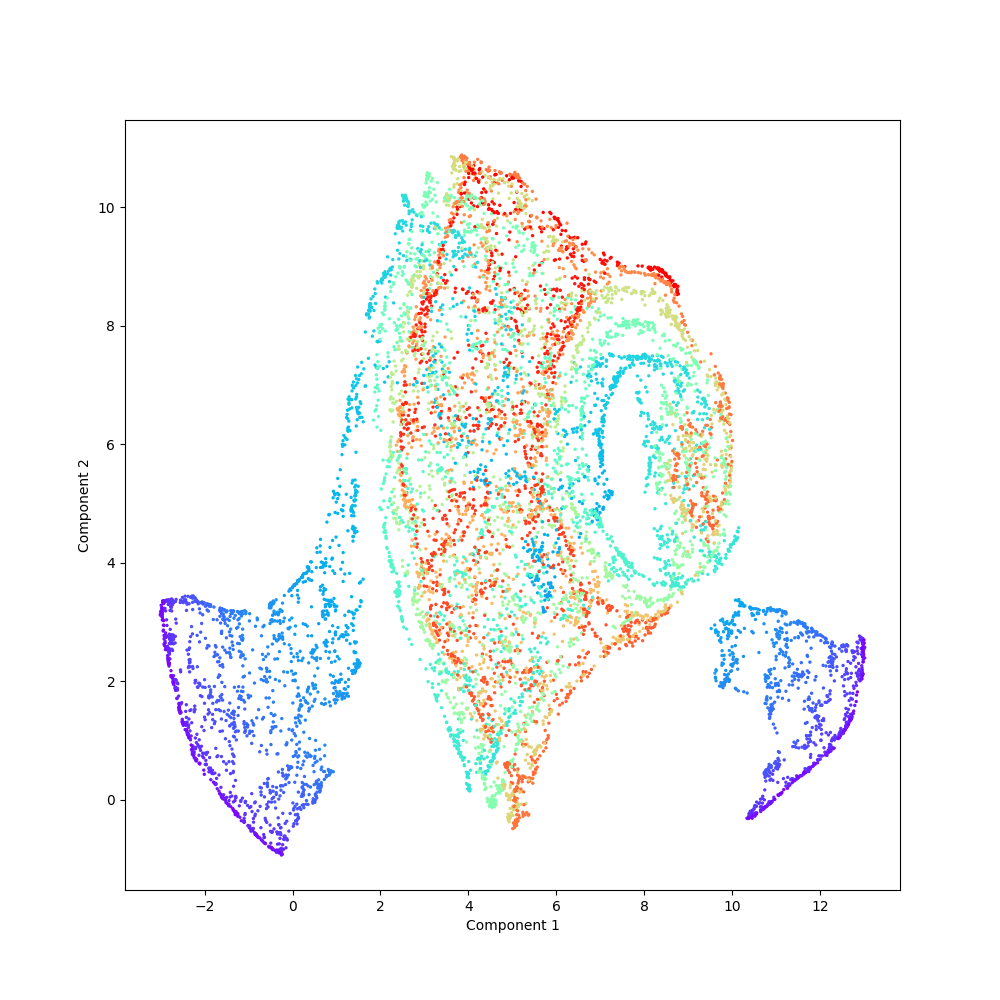}
	\end{tabular}
	\caption{Swiss Roll UMAP embedding with different $k$.}
	\label{tab:UMap_diffNN_SwissRoll}
\end{table}

\subsection{Trefoil-knotted protein chains}
\begin{figure}[H]
	\graphicspath{{image/ProteinData/}}
	\centering
	\subfigure[]{\includegraphics[width=0.24\textwidth]{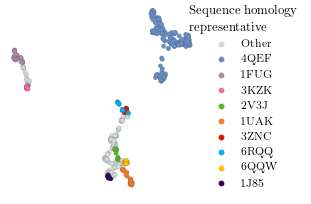}}
	\subfigure[]{\includegraphics[width=0.24\textwidth]{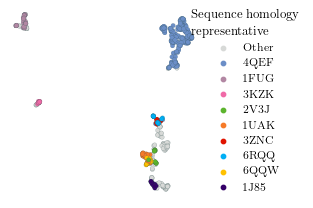}}
	
	\subfigure[]{\includegraphics[width=0.24\textwidth]{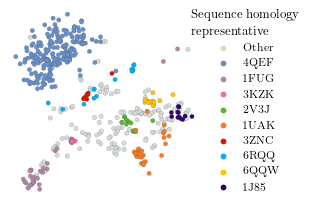}}
	\subfigure[]{\includegraphics[width=0.24\textwidth]{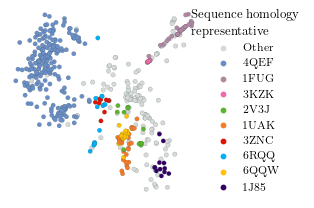}}
	\caption{ Clustering corresponding to sequence homology classes. First row: UMAP embedding of the space of trefoil-knotted proteins equipped with two distinct distances, k=15 : (a)Wasserstein distance, (b) landscape distance.
		Second row: IsUMap embedding, k=15 (c)Wasserstein distance, (d) landscape distance.}
	\label{fig:4.9}
\end{figure}

\begin{figure}[H]
	\graphicspath{{image/ProteinData/}}
	\centering
	\subfigure[]{\includegraphics[width=0.24\textwidth]{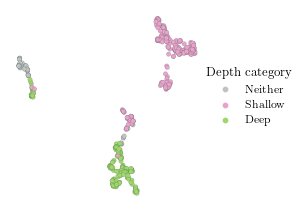}}
	\subfigure[]{\includegraphics[width=0.24\textwidth]{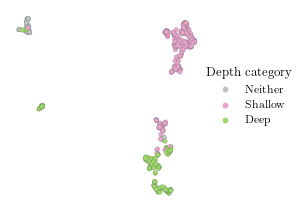}}
	
	\subfigure[]{\includegraphics[width=0.24\textwidth]{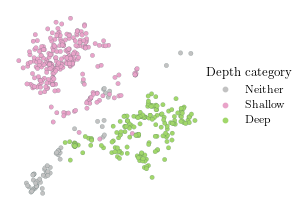}}
	\subfigure[]{\includegraphics[width=0.24\textwidth]{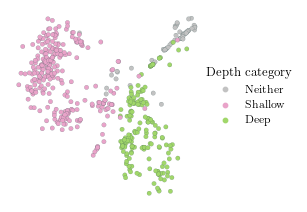}}
	\caption{Clustering corresponding to depth category. First row: UMAP embedding of the space of trefoil-knotted proteins equipped with two distinct distance on persistence diagrams, k=15 : (a)Wasserstein distance, (b) landscape distance.Second row: IsUMap embedding, k=15 (c)Wasserstein distance, (d) landscape distance.}
	\label{fig:4.10}
      \end{figure}

In the following, we analyze trefoil-knotted protein chains by employing Wasserstein and $L_1$ distances on their persistence diagrams and persistence landscapes, respectively, in place of the Euclidean distance to measure the topological similarity of different proteins within our dataset. Specifically, we compare the visualization results obtained using UMAP and IsUMap, building on the work of \cite{Benjamin23}, where a mathematical pipeline to analyze the geometric features and topological dissimilarity of protein entanglement is developed incorporating persistent homology and Isomap. \\
The dataset consists of proteins with backbones forming open-ended positive trefoil knots, labeled once by their knot depth (shallow, deep, or neither) and next by structural homology classes based on sequence similarity.\\
We first explore the dataset with labels specifying the structural homology classes, after applying IsUMap and UMAP. The resulting visualizations are illustrated in \cref{fig:4.9}, showing UMAP in the first row and IsUMap in the second row.\\
 Our findings indicate that while UMAP effectively clusters the dataset using both Wasserstein and $L_1$ distances, it does not distinctly reveal the underlying structures of the protein space in a distinct manner. In contrast, IsUMap not only highlights the tripod shape characteristic of trefoil-knotted proteins but also ensures a uniform distribution of the dataset. 
This uniform distribution prevents data points from being mapped on top of each other, which is seen clearly in UMAP representation and also to some extent in Isomap representation used in \cite{Benjamin23}. Therefore, IsUMap provides a clearer representation of the dataset’s underlying structure while effectively preserving clusters. 
In \cref{fig:4.10}, we examine the dataset of trefoil-knotted proteins labeled based on the depth category. The first row shows the UMAP visualization, and the second row showcases the IsUMap visualization. Both methods maintain cluster integrity using the two different distance metrics (i.e., Wasserstein and $L_1$ metrics on persistence diagram and landscape, respectively). However, IsUMap consistently reveals the tripod shape characteristic of the dataset while  maintaining the corresponding clusters depicted in Figures \cref{fig:4.9} and \cref{fig:4.10}, emphasizing its effectiveness in capturing the underlying structural essence of trefoil-knotted proteins, which was not explicitly detailed in the analyses in \cite{Benjamin23}.

      \subsection{Trajectory inference from single-cell RNA data}
We next consider trajectory inference and pseudo-time tasks using RNA velocity \cite{la18}, a method measuring the time derivative of gene expression. This approach is applied to infer the maturation trajectories of neural progenitor cells into various neural cell types, emphasizing continuity in underlying structures. Leveraging the human forebrain dataset from \cite{la18}, which explores transcriptional dynamics during brain development, we utilize IsUMap and UMAP to visualize these dynamics within this complex dataset. IsUMap enhances this representation in low dimension by ensuring a more uniform dataset distribution and identifying geodesic paths within the data, thereby preserving continuous underlying structures. 

Table \ref{tab:Trajectories} compares UMAP (First row) and IsUMap (second row) results in trajectory inference, as highlighted in \cite{chari23}. This comparison demonstrates limitations of UMAP in representing continuous relationships and its shortcoming in establishing a stage for the trajectory inference as a downstream task, primarily due to its reliance on a negative sampling step rather than local distortions \cite{damrich21}.\\
 In contrast, applying RNA velocity to representations resulted by applying IsUMap show an improved continuity in underlying structures and a more accurate trajectory inference, as depicted in Table  \ref{tab:Trajectories}. Additionally, varying the number of neighbors does not alter the trajectory inference and the continuous structures.

\begin{table}[H]
	\centering
	\tiny
	\begin{tabular}{>{\centering\arraybackslash}m{1cm}|*{2}{>{\centering\arraybackslash}m{3cm}}}
		& \textbf{k=17} &  \textbf{k=50} \\
		& (a) & (b)\\
		\hline \\
		\textbf{(1) UMAP} &
		\includegraphics[width=0.20\textwidth]{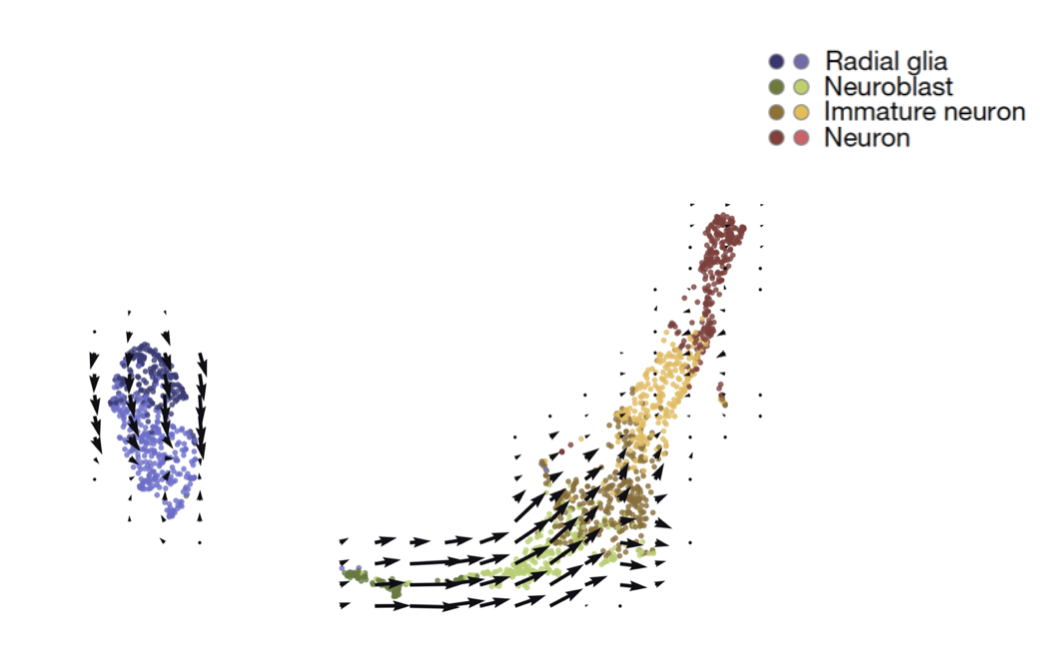} &
		\includegraphics[width=0.20\textwidth]{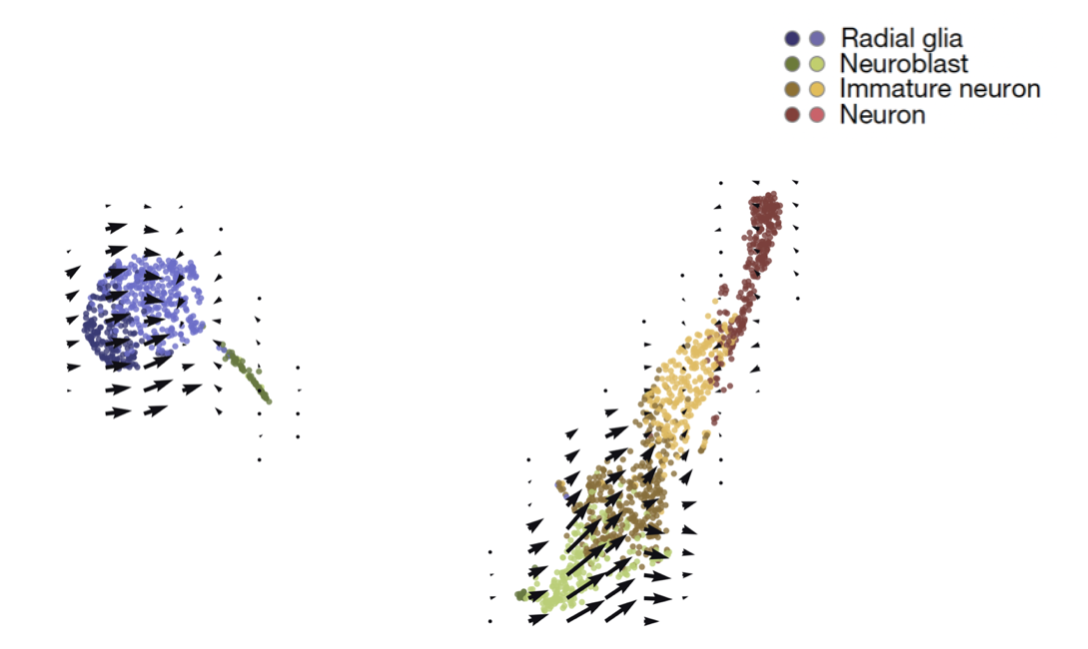}\\
		\textbf{(2) IsUMap} & 		
		\includegraphics[width=0.20\textwidth]{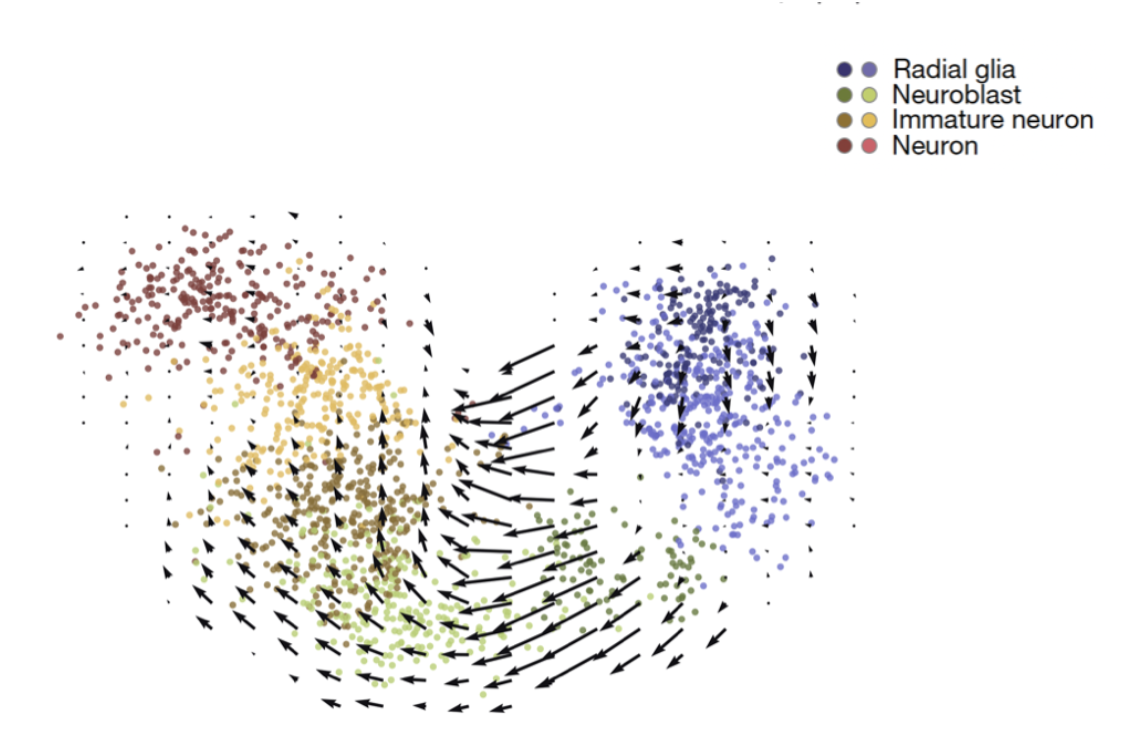} &
		\includegraphics[width=0.20\textwidth]{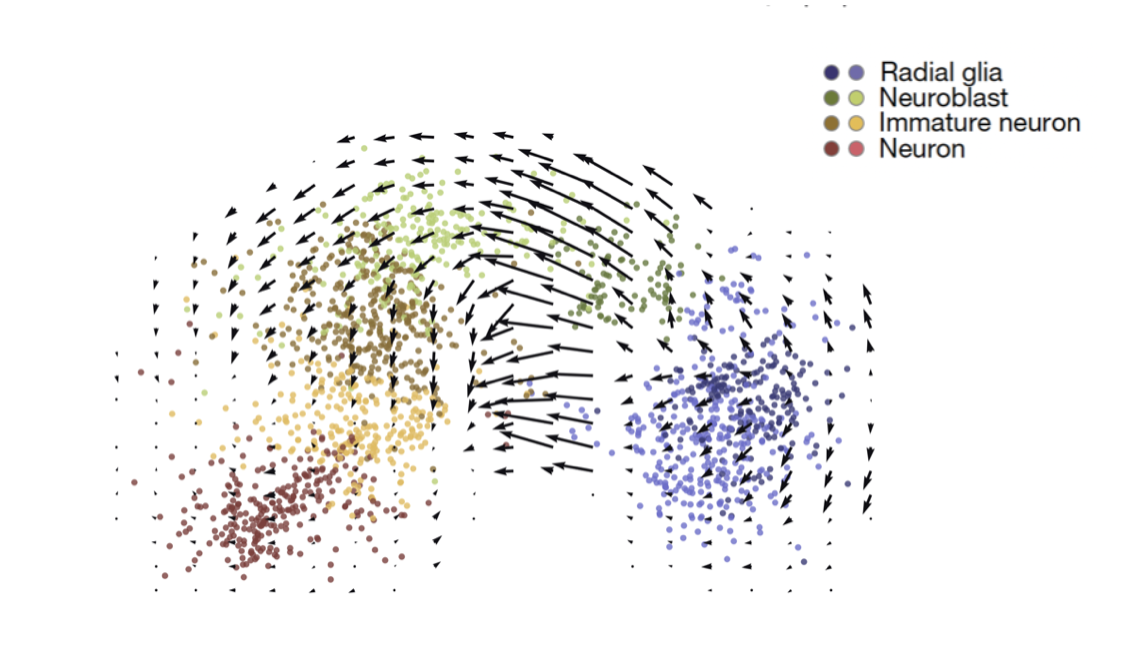}\\
	\end{tabular}
	\caption{Trajectory inference and continuous relationships. First row: Velocyto RNA velocity embeddings for UMAP with $k=17$ and $k=50$. Second row: Velocyto RNA velocity embeddings for IsUMap with $k=17$ and $k=50$.}
	\label{tab:Trajectories}
\end{table}

 \newpage 

\bibliographystyle{plainnat}    
\bibliography{bib} 
\end{document}